\documentclass[twocolumn,twoside]{IEEEtran}

\usepackage{etex}
\reserveinserts{28}
\usepackage{setspace,cite}

\usepackage[dvips]{graphicx}
\usepackage{multirow,multicol}
\usepackage[table]{xcolor}
\usepackage{ctable}

\usepackage[tbtags]{amsmath}
\usepackage{amsbsy}
\usepackage{amssymb}
\usepackage{amsfonts}

\usepackage{graphicx}
\usepackage{pstricks}
\usepackage{pst-node}
\usepackage{pstricks-add}
\usepackage{url}
\usepackage{algorithmic}
\usepackage{algorithm}
\usepackage{balance}
\usepackage{rotating}
\usepackage{multirow}
\usepackage{caption}
\usepackage{subcaption}
\usepackage{fancyvrb}
\usepackage{latexsym}
\usepackage{verbatim}

\newtheorem{theorem}{Theorem}
\newtheorem{proposition}{Proposition}
\newtheorem{lemma}{Lemma}

\newtheorem{remark}{Remark}

\newtheorem{example}{Example}

\newenvironment{densitemize}
{\begin{list}               
    {$\bullet$ \hfill}{
        \setlength{\leftmargin}{\parindent}
        \setlength{\parsep}{0.04\baselineskip}
        \setlength{\itemsep}{0.5\parsep}
        \setlength{\labelwidth}{\leftmargin}
        \setlength{\labelsep}{0em}}
    }
{\end{list}}

\providecommand{\eref}[1]{\eqref{eq:#1}}  
\providecommand{\cref}[1]{Chapter~\ref{#1}}
\providecommand{\sref}[1]{Section~\ref{sec:#1}}
\providecommand{\fref}[1]{Figure~\ref{fig:#1}}
\providecommand{\tref}[1]{Table~\ref{#1}}
\providecommand{\thref}[1]{Theorem~\ref{#1}}

\providecommand{\R}{\ensuremath{\mathbb{R}}}

\providecommand{\E}{\ensuremath{\mathbb{E}}}

\providecommand{\abs}[1]{\lvert#1\rvert}
\providecommand{\norm}[1]{\lVert#1\rVert}

\providecommand{\set}[1]{\left\{#1\right\}}

\providecommand{\bydef}{\overset{\text{def}}{=}}

\renewcommand{\vec}[1]{\ensuremath{\boldsymbol{#1}}}
\providecommand{\mat}[1]{\ensuremath{\boldsymbol{#1}}}

\providecommand{\calL}{\mathcal{L}}

\providecommand{\calO}{\mathcal{O}}

\providecommand{\calX}{\mathcal{X}}
\providecommand{\calY}{\mathcal{Y}}

\providecommand{\mI}{\mat{I}}

\providecommand{\vp}{\vec{p}}

\providecommand{\vs}{\vec{s}}

\providecommand{\vu}{\vec{u}}
\providecommand{\vv}{\vec{v}}

\providecommand{\vx}{\vec{x}}
\providecommand{\vy}{\vec{y}}

\providecommand{\mLambda}{\mat{\Lambda}}

\providecommand{\veta}{\vec{\eta}}

\providecommand{\vlambda}{\vec{\lambda}}

\providecommand{\Ytilde}{\widetilde{Y}}

\providecommand{\vone}{\vec{1}}

\providecommand{\Var}{\mathrm{Var}}

\newcommand{\defequal}{\mathop{\overset{\mbox{\tiny{def}}}{=}}}
\newcommand{\subjectto}{\mathop{\mathrm{subject\, to}}}
\newcommand{\argmin}[1]{\mathop{\underset{#1}{\mathrm{arg\,min}}}}

\newcommand{\minimize}[1]{\mathop{\underset{#1}{\mathrm{minimize}}}}

\newcommand{\diag}[1]{\mathop{\mathrm{diag}\left\{#1\right\}}}
\newcommand{\MSE}{\mathrm{MSE}}

\newgray{Gray08}{0.8}
\newgray{Gray07}{0.7}
\newgray{Gray06}{0.6}
\newgray{Gray05}{0.5}

\graphicspath{{pix7/}}

\begin{document}

\title{Monte Carlo Non-Local Means: Random\\ Sampling for Large-Scale Image Filtering}
\author{Stanley~H.~Chan,~\IEEEmembership{Member,~IEEE,}
        Todd~Zickler,~\IEEEmembership{Member,~IEEE}, \\ and Yue~M.~Lu,~\IEEEmembership{Senior Member,~IEEE}
\thanks{The authors are with the School of Engineering and Applied Sciences, Harvard University, Cambridge, MA 02138, USA. E-mails: \texttt{\{schan,zickler,yuelu\}@seas.harvard.edu}.}
\thanks{This work was supported in part by the Croucher Foundation Post-doctoral Research Fellowship (2012-2013), and in part by the U.S. National Science Foundation under Grant CCF-1319140. Preliminary material in this paper was presented at the 38th IEEE International Conference on Acoustics, Speech and Signal Processing (ICASSP), Vancouver, May 2013.}
\thanks{This paper follows the concept of reproducible research. All the results and examples presented in the paper are reproducible using the code and images available online at \texttt{http://lu.seas.harvard.edu/}.}}

\markboth{}{Chan \MakeLowercase{\textit{et al.}}: Monte-Carlo Non-Local Means}

\maketitle

\begin{abstract}
We propose a randomized version of the non-local means (NLM) algorithm for large-scale image filtering. The new algorithm, called Monte Carlo non-local means (MCNLM), speeds up the classical NLM by  computing a small subset of image patch distances, which are randomly selected according to a designed sampling pattern. We make two contributions. First, we analyze the performance of the MCNLM algorithm and show that, for large images or large external image databases, the random outcomes of MCNLM are tightly concentrated around the deterministic full NLM result. In particular, our error probability bounds show that, at any given sampling ratio, the probability for MCNLM to have a large deviation from the original NLM solution decays exponentially as the size of the image or database grows. Second, we derive explicit formulas for optimal sampling patterns that minimize the error probability bound by exploiting partial knowledge of the pairwise similarity weights. Numerical experiments show that MCNLM is competitive with other state-of-the-art fast NLM algorithms for single-image denoising. When applied to denoising images using an external database containing ten billion patches, MCNLM returns a randomized solution that is within 0.2 dB of the full NLM solution while reducing the runtime by three orders of magnitude.
\end{abstract}



\section{Introduction}
\label{section:introduciton}

\subsection{Background and Motivation}

In recent years, the image processing community has witnessed a wave of research aimed at developing new image denoising algorithms that exploit similarities between non-local patches in natural images. Most of these can be traced back to the non-local means (NLM) denoising algorithm of Buades \emph{et al.} \cite{Buades_Coll_2005_Journal,Buades_Coll_2005_AVSS} proposed in 2005. Although it is no longer the state-of-the-art method (see, \emph{e.g.}, \cite{Dabov_Foi_Katkovnik_2007,Talebi_Milanfar_2013} for some more recent leading algorithms), NLM remains one of the most influential algorithms in the current denoising literature.

Given a noisy image, the NLM algorithm uses two sets of image patches for denoising. The first is a set of noisy patches $\calY = \{\vy_1,\ldots,\vy_m\}$, where $\vy_i \in \R^d$ is a $d$-dimensional (\emph{i.e.}, $d$-pixel) patch centered at the $i$th pixel of the noisy image. The second set, $\calX = \{\vx_1,\ldots,\vx_n\}$, contains patches that are obtained from some reference images. Conceptually, NLM simply replaces each $i$th noisy pixel with a weighted average of pixels in the reference set. Specifically, the filtered value at the $i$th pixel (for $1 \le i \le m$) is given by
\begin{equation}
z = \frac{\sum_{j=1}^n w_{i,j} x_{j}}{\sum_{j=1}^n w_{i,j}},
\label{eq:fhat}
\end{equation}
where $x_j$ denotes the value of the center pixel of the $j$th reference patch $\vx_j \in \calX$, and the weights $\set{w_{i,j}}$ measure the similarities between the patches $\vy_i$ and $\vx_j$. A standard choice for the weights is
\begin{equation}
w_{i,j} = e^{-\norm{\vy_i - \vx_j}_{\mLambda}^2/(2h_r^2)},
\end{equation}
where $h_r$ is a scalar parameter determined by the noise level, and $\norm{\cdot}_{\mLambda}$ is the weighted $\ell_2$-norm with a diagonal weight matrix $\mLambda$, \emph{i.e.}, $\|\vy_i-\vx_j\|_{\mLambda}^2 \bydef (\vy_i-\vx_j)^T\mLambda(\vy_i-\vx_j)$.

In most implementations of NLM (see, \emph{e.g.}, \cite{Protter_Elad_Takeda_2009,Mairal_Bach_Ponce_2009,Chaudhury_Singer_2013,Milanfar_2013a,Dong_Zhang_Shi_2013,VanDeVille_Kocher_2009,Kervrann_Boulanger_2006}), the denoising process is based on a single image: the reference patches $\calX$ are the same as the noisy patches $\calY$. We refer to this setting, when $\calX = \calY$, as \emph{internal denoising}. This is in contrast to the setting in which the set of reference patches $\calX$ come from external image databases \cite{Zontak_Irani_2011,Levin_Nadler_2011,Levin_Nadler_Durand_2012}, which we refer to as \emph{external denoising}. For example, $15,000$ images (corresponding to a reference set of $n \approx 10^{10}$ patches) were used in \cite{Levin_Nadler_2011,Levin_Nadler_Durand_2012}. One theoretical argument for using large-scale external denoising was provided in \cite{Levin_Nadler_2011}: It is shown that, in the limit of large reference sets (\emph{i.e.}, when $n \rightarrow \infty$), external NLM converges to the minimum mean squared error estimator of the underlying clean images.

Despite its strong performance, NLM has a limitation of high computational complexity. It is easy to see that computing all the weights $\{w_{i,j}\}$ requires $\calO(mnd)$ arithmetic operations, where $m, n, d$ are, respectively, the number of pixels in the noisy image, the number of reference patches used, and the patch dimension. Additionally, about $\calO(mn)$ operations are needed to carry out the summations and multiplications in \eref{fhat} for all pixels in the image. In the case of internal denoising, these numbers are nontrivial since current digital photographs can easily contain tens of millions of pixels (\emph{i.e.}, $m = n \sim 10^7$ or greater). For external denoising with large reference sets (\emph{e.g.}, $n \sim 10^{10}$), the complexity is even more of an issue, making it very challenging to fully utilize the vast number of images that are readily available online and potentially useful as external databases.

\subsection{Related Work}

The high complexity of NLM is a well-known challenge. Previous methods to speed up NLM can be roughly classified in the following categories:

\emph{1. Reducing the reference set $\calX$.} If searching through a large set $\calX$ is computationally intensive, one natural solution is to pre-select a subset of $\calX$ and perform computation only on this subset \cite{Mahmoudi_Sapiro_2005,Coupe_Yger_2006,Brox_Kleinschmidt_2008}. For example, for internal denoising, a spatial weight $w^s_{i,j}$ is often included so that
\begin{equation}
w_{i,j} = w^s_{i,j} \;\cdot\; \underset{w^r_{i,j}}{\underbrace{e^{-\|\vy_i - \vx_j\|_{\mLambda}^2/(2h_r^2)}}}.
\label{eq:wswr}
\end{equation}
A common choice of the spatial weight is
\begin{equation}
w^s_{i,j} = \exp\{-d_{i,j}^2/(2h_s^2)\} \cdot \mathbb{I}\{ d'_{i,j} \le \rho\},
\label{eq:spatial_weight}
\end{equation}
where $d_{i,j}$ and $d'_{i,j}$ are, respectively, the Euclidean distance and the $\ell_\infty$ distance between the spatial locations of the $i$th and $j$th pixels; $\mathbb{I}$ is the indicator function; and $\rho$ is the width of the spatial search window. By tuning $h_s$ and $\rho$, one can adjust the size of $\calX$ according to the heuristic that nearby patches are more likely to be similar.

\emph{2. Reducing dimension $d$.} The patch dimension $d$ can be reduced by several methods. First, SVD projection \cite{Tasdizen_2008,Orchard_Ebrahimi_2008,VanDeVille_Kocher_2009,VanDeVille_Kocher_2010} can be used to project the $d$-dimensional patches onto a lower dimensional space spanned by the principal components computed from $\calX$. Second, the integral image method \cite{Darbon_Cunha_2008,Wang_Guo_2006,Karnati_Uliyar_2009} can be used to further speed up the computation of $\|\vy_i-\vx_j\|_{\mLambda}^2$.  Third, by assuming a Gaussian model on the patch data, a probabilistic early termination scheme \cite{Vignesh_Oh_2010} can be used to stop computing the squared patch difference before going through all the pixels in the patches.

\emph{3. Optimizing data structures.} The third class of methods embed the patches in $\calX$ and $\calY$ in some form of optimized data structures. Some examples include the fast bilateral grid \cite{Paris_Durand_2009}, the fast Gaussian transform \cite{Yang_Duraiswami_2003}, the Gaussian KD tree \cite{Adams_Baek_Davis_2010,Adams_Gelfand_2009}, the adaptive manifold method \cite{Gastal_Oliveira_2012}, and the edge patch dictionary \cite{Bhujle_Chaudhuri_2014}. The data structures used in these algorithms can significantly reduce the computational complexity of the NLM algorithm. However, building these data structures often requires a lengthy pre-processing stage, or require a large amount of memory, thereby placing limits on one's ability to use large reference patch sets $\cal X$. For example, building a Gaussian KD tree requires the storage of $\calO(nd)$ double precision numbers (see, \emph{e.g.}, \cite{Adams_Gelfand_2009,Arietta_Lawrence_2011}.)


\subsection{Contributions}
In this paper, we propose a randomized algorithm to reduce the computational complexity of NLM for both internal and external denoising. We call the method \emph{Monte Carlo Non-Local Means} (MCNLM), and the basic idea is illustrated in \fref{MCNLM} for the case of internal denoising. For each pixel $i$ in the noisy image, we randomly select a set of $k$ reference pixels according to some sampling pattern and compute a $k$-subset of the weights $\{w_{i,j}\}_{j=1}^n$ to form an approximated solution to \eref{fhat}. The computational complexity of MCNLM is $\calO(mkd)$, which can be significantly lower than the original complexity $\calO(mnd)$ when only $k \ll n$ weights are computed. Furthermore, since there is no need to re-organize the data, the memory requirement of MCNLM is $\calO(m+n)$. Therefore, MCNLM is scalable to large reference patch sets $\calX$, as we will demonstrate in \sref{experiment}.

\begin{figure}[t]
\centering
\begin{pspicture}(-1.8,-1.5)(6,1.5)
\rput(-0.8,0){\includegraphics[width=0.3\linewidth]{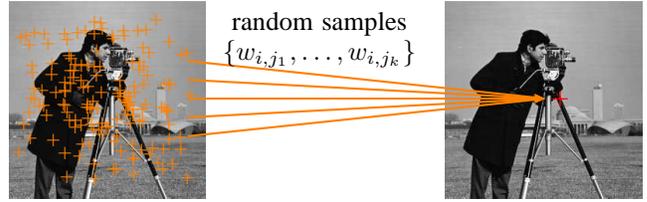}}
\rput(5,0){\includegraphics[width=0.3\linewidth]{cameraman.eps}}
\psRandom[randomPoints=100,dotsize=4pt,dotstyle=+,linecolor=orange,fillcolor=orange](-2,-1.1)(0.3,1.1){}
\psRandom[randomPoints=50,dotsize=4pt,dotstyle=+,linecolor=orange,fillcolor=orange](-1.4,-0.5)(-0.3,0.5){}
\psline[showpoints=false,linecolor=orange]{->}(0.25,0.5)(5,0)
\psline[showpoints=false,linecolor=orange]{->}(0.25,0.25)(5,0)
\psline[showpoints=false,linecolor=orange]{->}(0.25,0)(5,0)
\psline[showpoints=false,linecolor=orange]{->}(0.25,-0.25)(5,0)
\psline[showpoints=false,linecolor=orange]{->}(0.25,-0.5)(5,0)
\psdots[fillcolor=orange,linecolor=red,dotsize=5pt,dotstyle=+](5.2,0)
\rput(2,1){random samples}
\rput(2,0.6){$\{w_{i,j_1},\ldots,w_{i,j_k}\}$}
\end{pspicture}
\caption{Illustration of the proposed MCNLM algorithm for internal denoising: We randomly select, according to a given sampling pattern, a set of $k$ weights $\{w_{i,j_1},\ldots,w_{i,j_k}\}$, and use these to compute an approximation of the full NLM result in \eref{fhat}. The output of MCNLM is random. However, as the size of the problem (\emph{i.e.}, $n$) gets larger, these random estimates become tightly concentrated around the true result.}
\label{fig:MCNLM}
\end{figure}

The two main contributions of this paper are as follows.

\emph{1. Performance guarantee}. MCNLM is a randomized algorithm. It would not be a useful one if its random outcomes fluctuated widely in different executions on the same input data. In \sref{performance_analysis}, we address this concern by showing that, as the size of the reference set $\calX$ increases, the randomized MCNLM solutions become tightly concentrated around the original NLM solution. In particular, we show in Theorem~\ref{thm:general_main} (and Proposition~\ref{proposition:uniform sampling}) that, for \emph{any} given sampling pattern, the probability of having a large deviation from the original NLM solution drops exponentially as the size of $\calX$ grows.

\emph{2. Optimal sampling patterns}. We derive optimal sampling patterns to minimize the approximation error probabilities established in our performance analysis. We show that seeking the optimal sampling pattern is equivalent to solving a variant of the classical water-filling problem, for which a closed-form expression can be found (see Theorem~\ref{thm:optimal p_j}). We also present two practical sampling pattern designs that exploit partial knowledge of the pairwise similarity weights.

The rest of the paper is organized as follows. After presenting the MCNLM algorithm and discuss its basic properties in \sref{proposed_method}, we analyze the performance in \sref{performance_analysis} and derive the optimal sampling patterns in \sref{optimal_sampling}. Experimental results are given in \sref{experiment}, and concluding remarks are given in \sref{conclusion}.



\section{Monte Carlo Non-local Means}
\label{sec:proposed_method}

\textbf{Notation:} Throughout the paper, we use $m$ to denote the number of pixels in the noisy image, and $n$ the number patches in the reference set $\calX$. We use upper-case letters, such as $X,Y,Z$, to represent random variables, and lower-case letters, such as $x, y, z$, to represent deterministic variables. Vectors are represented by bold letters, and $\vec{1}$ denotes a constant vector of which all entries are one. Finally, for notational simplicity in presenting our theoretical analysis, we assume that all pixel intensity values have been normalized to the range $[0, 1]$.

\subsection{The Sampling Process}

As discussed in Section \ref{section:introduciton}, computing all the weights $\{w_{i,j}\}_{1 \le i \le m, 1 \le j \le n}$ is computationally prohibitive when $m$ and $n$ are large. To reduce the complexity, the basic idea of MCNLM is to randomly select a subset of $k$ representatives of $\{w_{i,j}\}$ (referred to as samples) to approximate the sums in the numerator and denominator in \eref{fhat}. The sampling process in the proposed algorithm is applied to each of the $m$ pixels in the noisy image \textit{independently}. Since the sampling step and subsequent computations have the same form for each pixel, we shall drop the pixel index $i$ in $\set{w_{i,j}}$, writing the weights as $\set{w_j}_{1 \le j \le n}$ for notational simplicity.

The sampling process of MCNLM is determined by a sequence of \emph{independent} random variables $\{I_j\}_{j=1}^n$ that take the value $0$ or $1$ with the following probabilities
\begin{equation}\label{eq:Bernoulli_rv}
\Pr[I_j = 1] = p_j \quad\quad \mbox{and} \quad\quad \Pr[I_j = 0] = 1- p_j.
\end{equation}
The $j$th weight $w_{j}$ is sampled if and only if $I_j = 1$. In what follows, we assume that $0 < p_j \le 1$, and refer to the vector of all these probabilities $\vp \defequal [p_1,\ldots,p_n]^T$ as the \emph{sampling pattern} of the algorithm.

The ratio between the number of samples taken and the number of reference patches in $\calX$ is a random variable
\begin{equation}
S_n = \frac{1}{n}\sum_{j=1}^n I_j,
\label{eq:Sn}
\end{equation}
of which the expected value is
\begin{equation}
\E[S_n] = \frac{1}{n}\sum_{j=1}^n \E[I_j] = \frac{1}{n}\sum_{j=1}^n p_j \bydef \xi.
\label{eq:xi}
\end{equation}
We refer to $S_n$ and $\xi$ as the \emph{empirical sampling ratio} and the \emph{average sampling ratio}, respectively. $\xi$ is an important parameter of the MCNLM algorithm. The original (or ``full'') NLM corresponds to the setting when $\xi = 1$: In this case, $\vp = \vec{1} \bydef [1, \ldots, 1]^T$, so that all the samples are selected with probability one.

\subsection{The MCNLM Algorithm}
\label{section:proposed method MCNLM algorithm}

Given a set of random samples from $\calX$, we approximate the numerator and denominator in \eref{fhat} by two random variables
\begin{equation}
A(\vp) \defequal \frac{1}{n}\sum_{j=1}^n \frac{x_jw_{j}}{p_j} I_j \quad \mbox{and} \quad
B(\vp) \defequal \frac{1}{n}\sum_{j=1}^n \frac{w_{j}}{p_j} I_j,
\label{eq:A and B}
\end{equation}
where the argument $\vp$ emphasizes the fact that the distributions of $A$ and $B$ are determined by the sampling pattern $\vp$.

It is easy to compute the expected values of $A(\vp)$ and $B(\vp)$ as
\begin{align}
\mu_A &\defequal \E[A(\vp)] = \frac{1}{n}\sum_{j=1}^n x_j w_j, \label{eq:muA}\\
\mu_B &\defequal \E[B(\vp)] = \frac{1}{n}\sum_{j=1}^n w_j. \label{eq:muB}
\end{align}
Thus, up to a common multiplicative constant $1/n$, the two random variables $A(\vp)$ and $B(\vp)$ are \textit{unbiased} estimates of the true numerator and denominator, respectively.

The full NLM result $z$ in \eref{fhat} is then approximated by
\begin{equation}
Z(\vp) \defequal \frac{A(\vp)}{B(\vp)} = \frac{\sum_{j=1}^n \frac{x_jw_{j}}{p_j} I_j}{\sum_{j=1}^n \frac{w_{j}}{p_j} I_j}.
\label{eq:Zhat}
\end{equation}
In general, $\E[Z(\vp)] = \E\left[\frac{A(\vp)}{B(\vp)}\right] \neq \frac{\E[A(\vp)]}{\E[B(\vp)]} = z$, and thus $Z(\vp)$ is a \emph{biased} estimate of $z$. However, we will show in \sref{performance_analysis} that the probability of having a large deviation in $|Z(\vp)-z|$ drops exponentially as $n \rightarrow \infty$. Thus, for a large $n$, the MCNLM solution \eref{Zhat} can still form a very accurate approximation of the original NLM solution \eref{fhat}.

\begin{algorithm}[t]
\caption{Monte Carlo Non-local Means (MCNLM)}
\begin{algorithmic}[1]
\STATE For each noisy pixel $i = 1,\ldots,m$, do the followings.
\STATE Input: Noisy patch $\vy_i \in \calY$, database $\calX = \{\vx_1,\ldots,\vx_n\}$ and sampling pattern $\vp = [p_1,\ldots,p_n]^T$ such that $0 < p_j \le 1$, and $\sum_{j=1}^n p_j = n\xi$.
\STATE Output: A randomized estimate $Z(\vp)$.
\FOR{$j = 1,\ldots,n$}
    \STATE Generate a random variable $I_j \sim \mbox{Bernoulli}(p_j)$.
    \STATE If $I_j = 1$, then compute the weight $w_{j}$.
\ENDFOR
\STATE Compute $A(\vp) = \frac{1}{n} \sum_{j=1}^n \frac{w_{j}x_j}{p_j}I_j$.
\STATE Compute $B(\vp) = \frac{1}{n} \sum_{j=1}^n \frac{w_{j}}{p_j}I_j$.
\STATE Output $Z(\vp) =  A(\vp)/B(\vp)$.
\end{algorithmic}
\label{alg:MCNLM}
\end{algorithm}

Algorithm \ref{alg:MCNLM} shows the pseudo-code of MCNLM for internal denoising. We note that, except for the Bernoulli sampling process, all other steps are identical to the original NLM. Therefore, MCNLM can be thought of as adding a complementary sampling process on top of the original NLM. The marginal cost of implementation is thus minimal.

\begin{figure*}[!]
\centering
\begin{tabular}{ccc}
\includegraphics[width=0.3\linewidth]{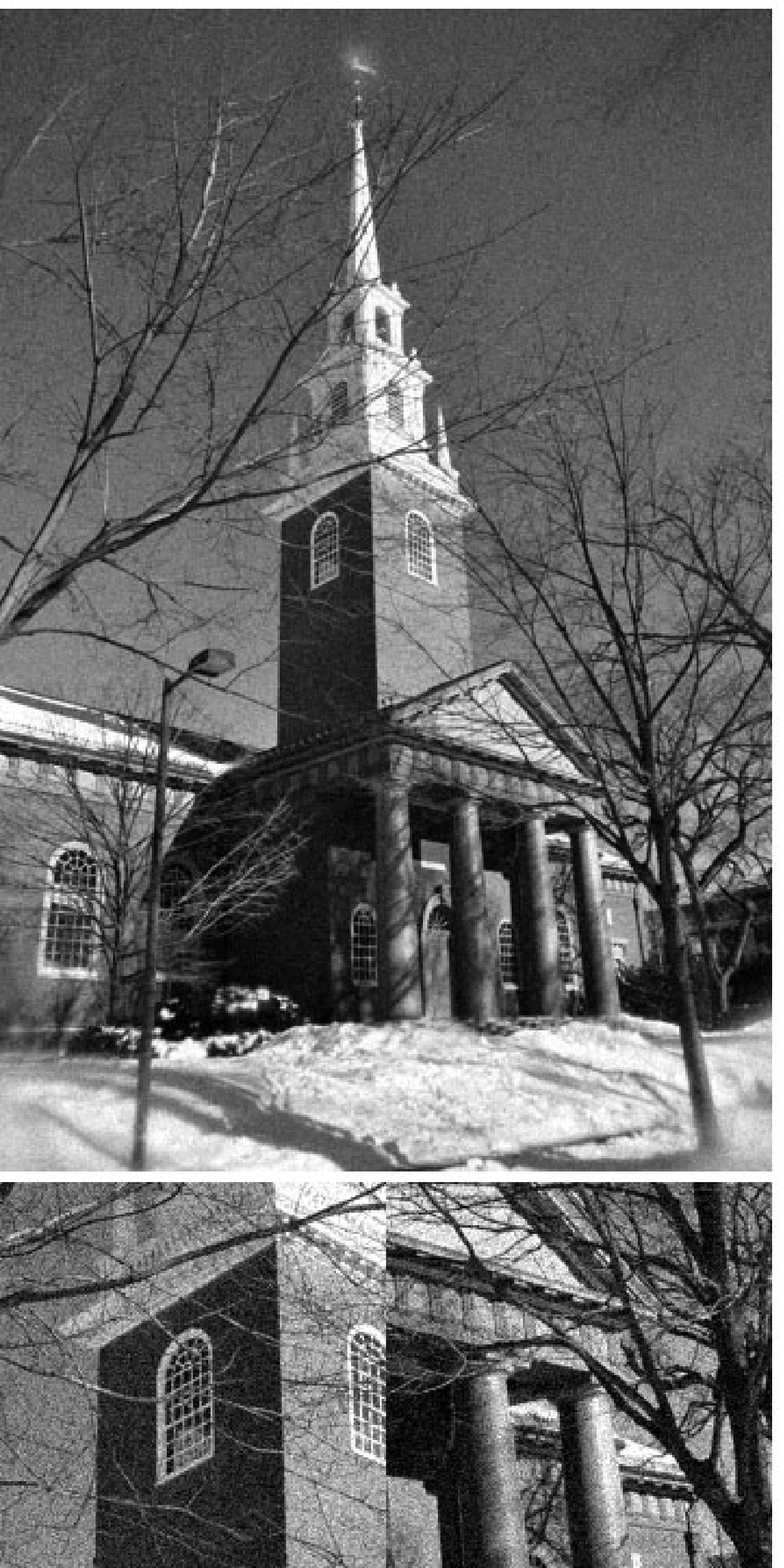}&
\includegraphics[width=0.3\linewidth]{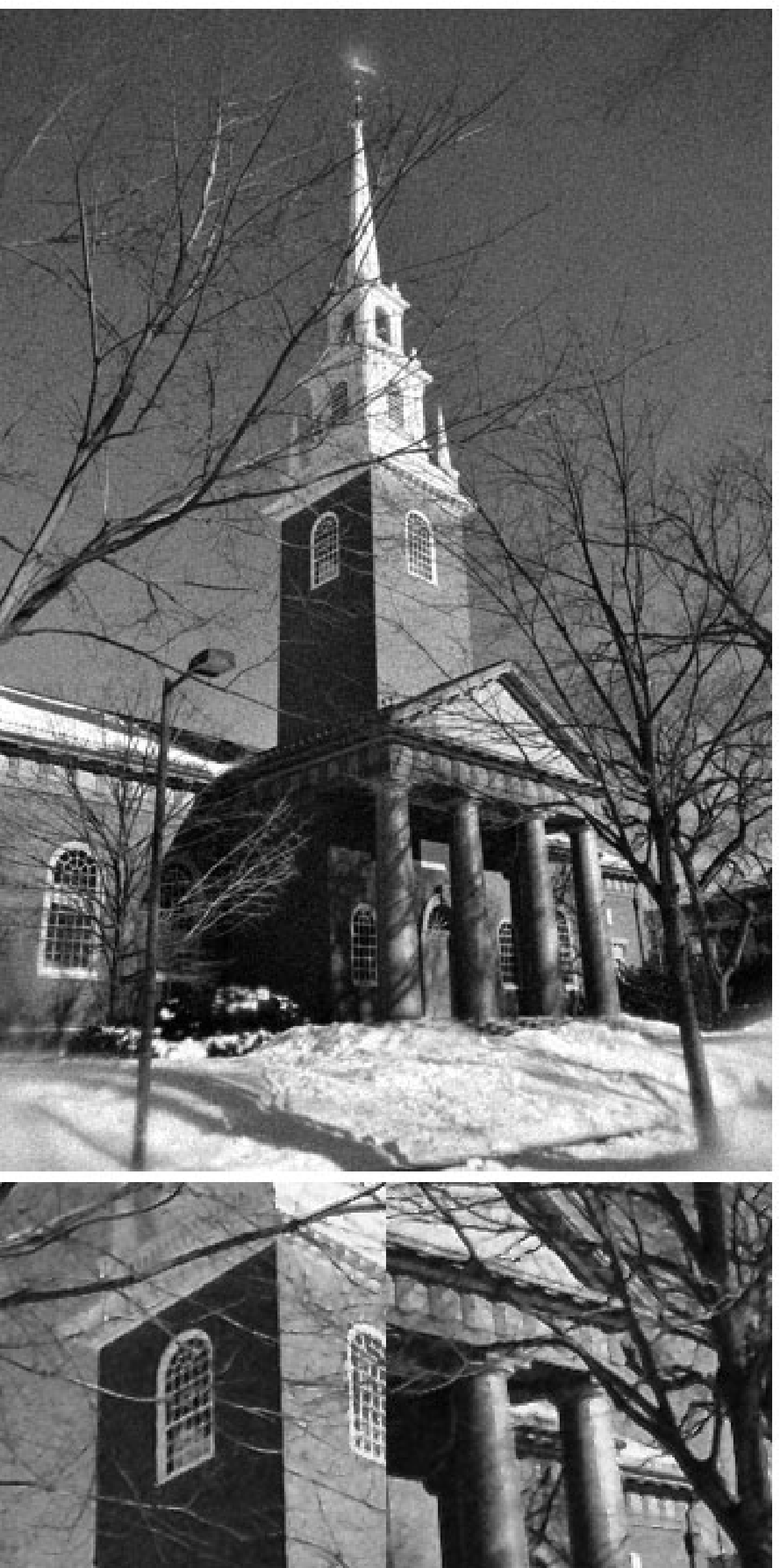}&
\includegraphics[width=0.3\linewidth]{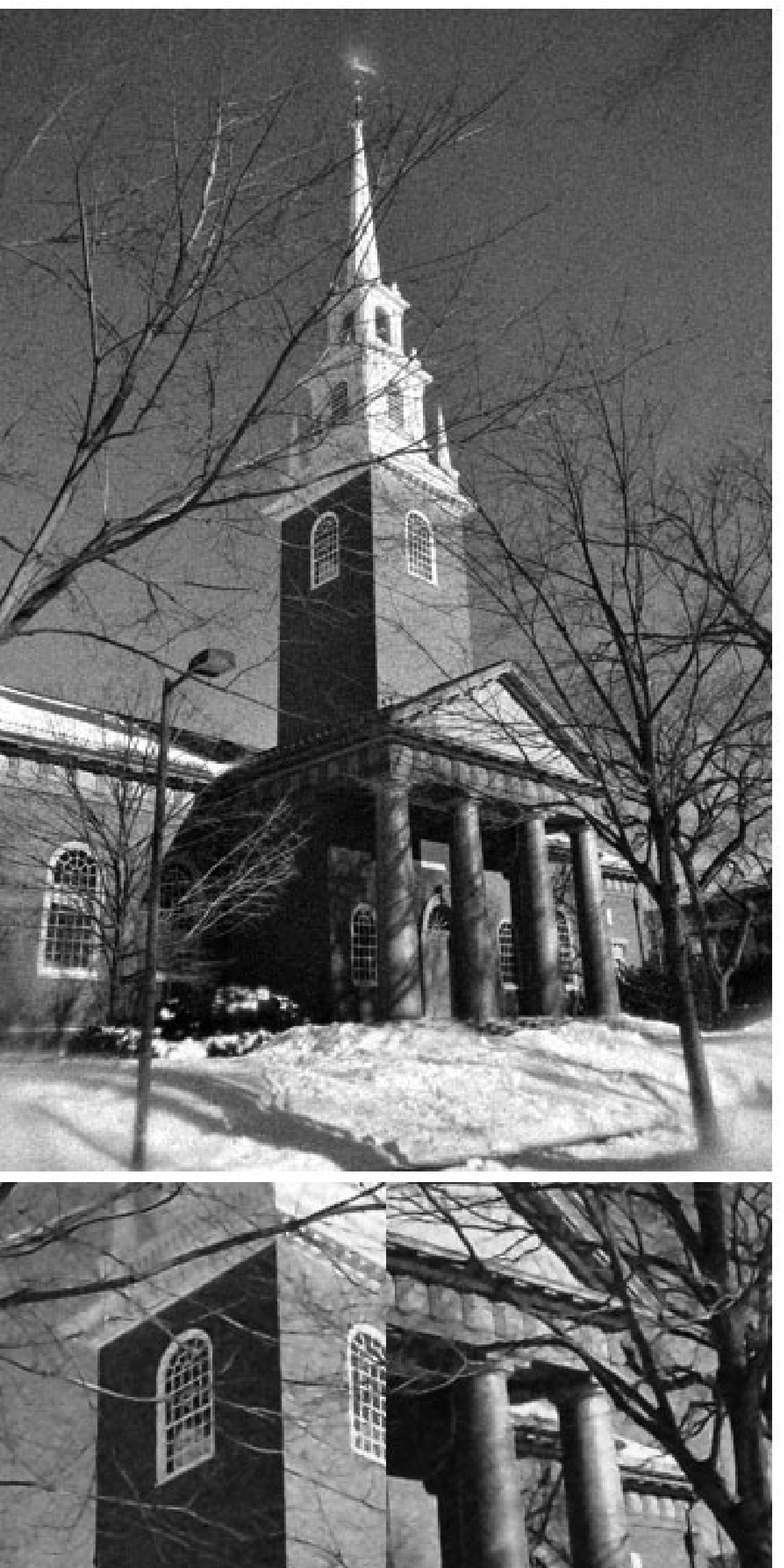}\\
noisy (24.60 dB) & $\xi = 0.005$ (27.58 dB) & $\xi = 0.1$ (28.90 dB)
\end{tabular}
\caption{Denoising an image of size $1072 \times 712$ by MCNLM with uniform sampling. (a) The original image is corrupted with i.i.d. Gaussian noise with $\sigma = 15/255$. (b) and (c) Denoised images with sampling ratio $\xi = 0.005$ and $\xi = 0.1$, respectively. Shown in parenthesis are the PSNR values (in dB) averaged over 100 trials.}
\label{fig:harvard}
\end{figure*}

\begin{example}
To empirically demonstrate the usefulness of the simple sampling mechanism of MCNLM, we apply the algorithm to a $1072 \times 712$ image shown in \fref{harvard}(a). Here, we use $\calX = \calY$, with $m = n \approx 7.6 \times 10^5$. In this experiment, we let the noise be i.i.d. Gaussian with zero mean and standard deviation $\sigma = 15/255$. The patch size is $5 \times 5$. In computing the similarity weights in \eref{wswr} and \eref{spatial_weight}, we set the parameters as follows: $h_r = 15/255$, $h_s = \infty$, $\rho = \infty$ (\emph{i.e.}, no spatial windowing) and $\mLambda = \frac{1}{25}\mI$. We choose a uniform sampling pattern, \emph{i.e.,} $\vp = [\xi, \ldots, \xi]^T$, for some sampling ratio $0 < \xi < 1$.

The results of this experiment are shown in \fref{harvard} and \fref{harvard,psnr}. The peak signal-to-noise ratio (PSNR) curve detailed in \fref{harvard,psnr} shows that MCNLM converges to its limiting value rapidly as the sampling ratio $\xi$ approaches 1. For example, at $\xi = 0.1$ (\emph{i.e.}, a  roughly ten-fold reduction in computational complexity), MCNLM achieves a PSNR that is only $0.2$dB away from the full NLM result. More numerical experiments will be presented in \sref{experiment}.
\end{example}

\begin{figure}[!]
\centering
\includegraphics[width=1\linewidth]{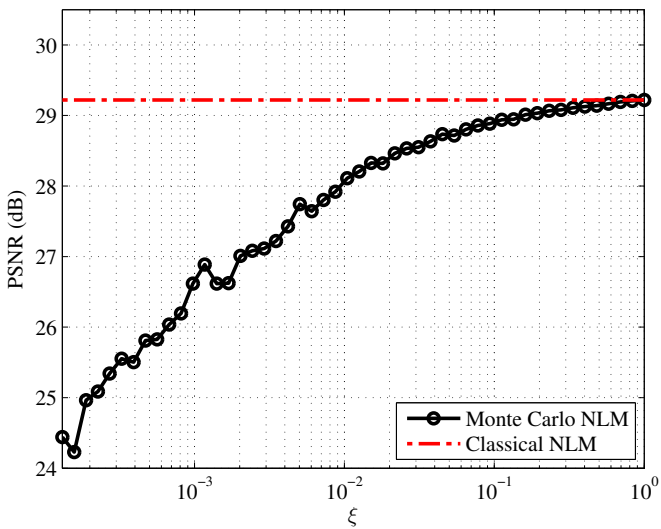}
\caption{PSNR as a function of the average sampling ratio $\xi$. The ``circled'' line indicates the result of MCNLM. The horizontal line indicates the result of the full NLM (\emph{i.e.}, MCNLM at $\xi = 1$). Note that at $\xi = 0.1$, MCNLM achieves a PSNR that is only $0.2$ dB below the full NLM result. Additional experiments are presented in \sref{experiment}.}
\label{fig:harvard,psnr}
\end{figure} 

\section{Performance Analysis}
\label{sec:performance_analysis}

One fundamental question about MCNLM is whether its random estimate $Z(\vp)$ as defined in \eref{Zhat} will be a good approximation of the full NLM solution $z$, especially when the sampling ratio $\xi$ is small. In this section, we answer this question by providing a rigorous analysis on the approximation error $\abs{Z(\vp) - z}$.

\subsection{Large Deviations Bounds}
The mathematical tool we use to analyze the proposed MCNLM algorithm comes from the probabilistic \emph{large deviations theory} \cite{DemboZ:2010}. This theory has been widely used to quantify the following phenomenon: A \emph{smooth} function $f(X_1,\ldots,X_n)$ of a large number of \emph{independent} random variables $X_1, \ldots, X_n$ tends to concentrate very tightly around its mean $\E[f(X_1,\ldots,X_n)]$. Roughly speaking, this concentration phenomenon happens because, while $X_1,\ldots,X_n$ are individually random in nature, it is unlikely for many of them to work collaboratively to alter the overall system by a significant amount. Thus, for large $n$, the randomness of these variables tends to be ``canceled out'' and the function $f(X_1,\ldots,X_n)$ stays roughly constant.

To gain insights from a concrete example, we first apply the large deviations theory to study the empirical sampling ratio $S_n$ as defined in \eref{Sn}. Here, the independent random variables are the Bernoulli random variables $\set{I_j}_{1\le j \le n}$ introduced in \eref{Bernoulli_rv}, and the smooth function $f(\cdot)$ computes their average.

It is well known from the law of large numbers (LLN) that the empirical mean $S_n$ of a large number of independent random variables stays very close to the true mean, which is equal to the average sampling ratio $\xi$ in our case. In particular, by the standard Chebyshev inequality \cite{Grimmett:2001}, we know that
\begin{equation}
\Pr[S_n - \E[S_n]>\varepsilon] \le \Pr[|S_n - \E[S_n]|>\varepsilon]<\frac{\Var[I_1]}{n\varepsilon^2},
\label{eq:LLN}
\end{equation}
for every positive $\varepsilon$.



One drawback of the bound in \eref{LLN} is that it is overly loose, providing only a linear rate of decay for the error probabilities as $n \rightarrow \infty$. In contrast, the large deviations theory provides many powerful probability inequalities which often lead to much tighter bounds with exponential decays. In this work, we will use one particular inequality in the large deviations theory, due to S. Bernstein \cite{Bernstein_1946}:

\begin{lemma}[Bernstein Inequality \cite{Bernstein_1946}]\label{lemma:Bernstein}
Let $X_1,\ldots,X_n$ be a sequence of independent random variables. Suppose that $l_j \le X_j \le u_j$ for all $j$, where $u_j$ and $l_j$ are constants. Let $S_n = (1/n)\sum_{j=1}^n X_j$, and $M = \max_{1 \le j \le n}(u_j-l_j)/2$. Then for every positive $\varepsilon$,
\begin{align}
&\Pr\left[  S_n - \E[S_n] > \varepsilon \right] \nonumber\\
&\qquad\quad\le \exp{\left\{-\frac{n\varepsilon^2}{2\left( \frac{1}{n}\sum_{j=1}^n \Var[X_j] + M\varepsilon/3\right)}\right\}}.\label{eq:Bernstein}
\end{align}
\end{lemma}
\begin{figure}[t]
\centering
\includegraphics[width=\linewidth]{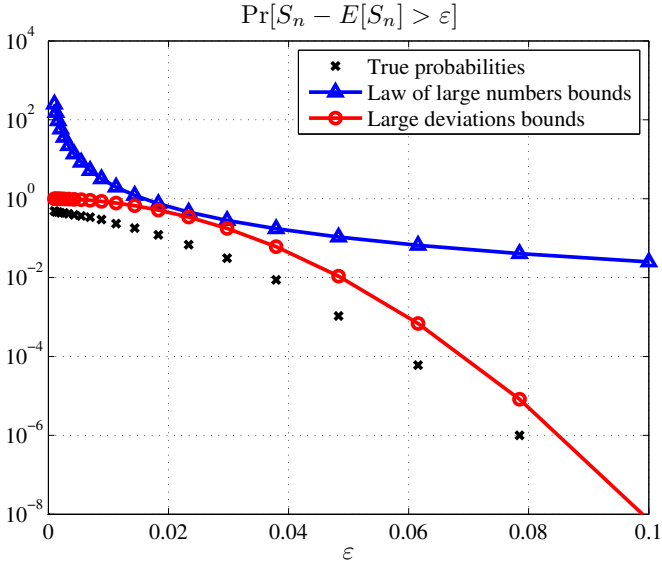}
\caption{Comparing the large deviations bound \eref{Bernstein}, the LLN bound \eref{LLN}, and the true error probability $\Pr[S_n - \E[S_n] > \varepsilon]$ as estimated by Monte Carlo simulations. Fixing $n = 10^6$, we plot the bounds and probabilities for different values of $\varepsilon$.}
\label{fig:bernoulli}
\end{figure}
To see how Bernstein's inequality can give us a better probability bound for the empirical sampling ratio $S_n$, we note that $X_j = I_j$ in our case. Thus, $M = 1$ and $\E[S_n] = \xi$. Moreover, if the sampling pattern is uniform, \emph{i.e.}, $\vp = [\xi, \ldots, \xi]^T$, we have $\frac{1}{n}\sum_{j=1}^n \Var[X_j] = \frac{1}{n} \sum_{j=1}^n p_j(1-p_j) = \xi(1-\xi)$. Substituting these numbers into \eref{Bernstein} yields an \emph{exponential} upper bound on the error probability, which is plotted and compared in \fref{bernoulli} against the LLN bound in \eref{LLN} and against the true probabilities estimated by Monte Carlo simulations. It is clear that the exponential bound provided by Bernstein's inequality is much tighter than that provided by LLN.


\subsection{General Error Probability Bound for MCNLM}

We now derive a general bound for the error probabilities of MCNLM. Specifically, for any $\varepsilon > 0$ and any sampling pattern $\vp$ satisfying the conditions that $0 < p_j \le 1$ and $\frac{1}{n}\sum_{j=1}^n p_j = \xi$, we want to study
\begin{equation}
\Pr \left[ \left| Z(\vp) - z \right| > \varepsilon \right],
\end{equation}
where $z$ is the full NLM result defined in \eref{fhat} and $Z(\vp)$ is the MCNLM estimate defined in \eref{Zhat}.

\begin{theorem}\label{thm:general_main}
Assume that $w_j > 0$ for all $j$. Then for every positive $\varepsilon$,
\begin{align}
&\Pr\left[ \left| Z(\vp) - z \right| > \varepsilon \right] \le \exp\left\{ -n\xi \right\} \nonumber \\
&\quad+ \exp\left\{ \frac{-n (\mu_B \varepsilon)^2 }{ 2 \left(\frac{1}{n}\sum_{j=1}^n \alpha_j^2 \left( \frac{1-p_j}{p_j}\right) +  (\mu_B \varepsilon) M_\alpha/6\right)}\right\} \nonumber \\
&\quad + \exp\left\{ \frac{-n (\mu_B \varepsilon)^2 }{ 2 \left(\frac{1}{n}\sum_{j=1}^n \beta_j^2  \left( \frac{1-p_j}{p_j}\right) + (\mu_B \varepsilon) M_\beta/6\right)}\right\},
\label{eq:thm1_main}
\end{align}
where $\mu_B$ is the average similarity weights defined in \eref{muB}, $\alpha_j = w_j \left(x_j - z - \varepsilon\right)$, $\beta_j = w_j \left(x_j - z + \varepsilon\right)$, and
\begin{align*}
M_\alpha = \max\limits_{1\le j \le n} \frac{|\alpha_j|}{p_j}, \quad\mathrm{and}\quad M_\beta  = \max\limits_{1\le j \le n} \frac{|\beta_j|}{p_j}.
\end{align*}
\end{theorem}
\begin{IEEEproof}
See Appendix~\ref{appendix:convergence}.
\end{IEEEproof}

\begin{figure*}[t]
\centering
\begin{tabular}{cc}
\includegraphics[width=0.45\linewidth]{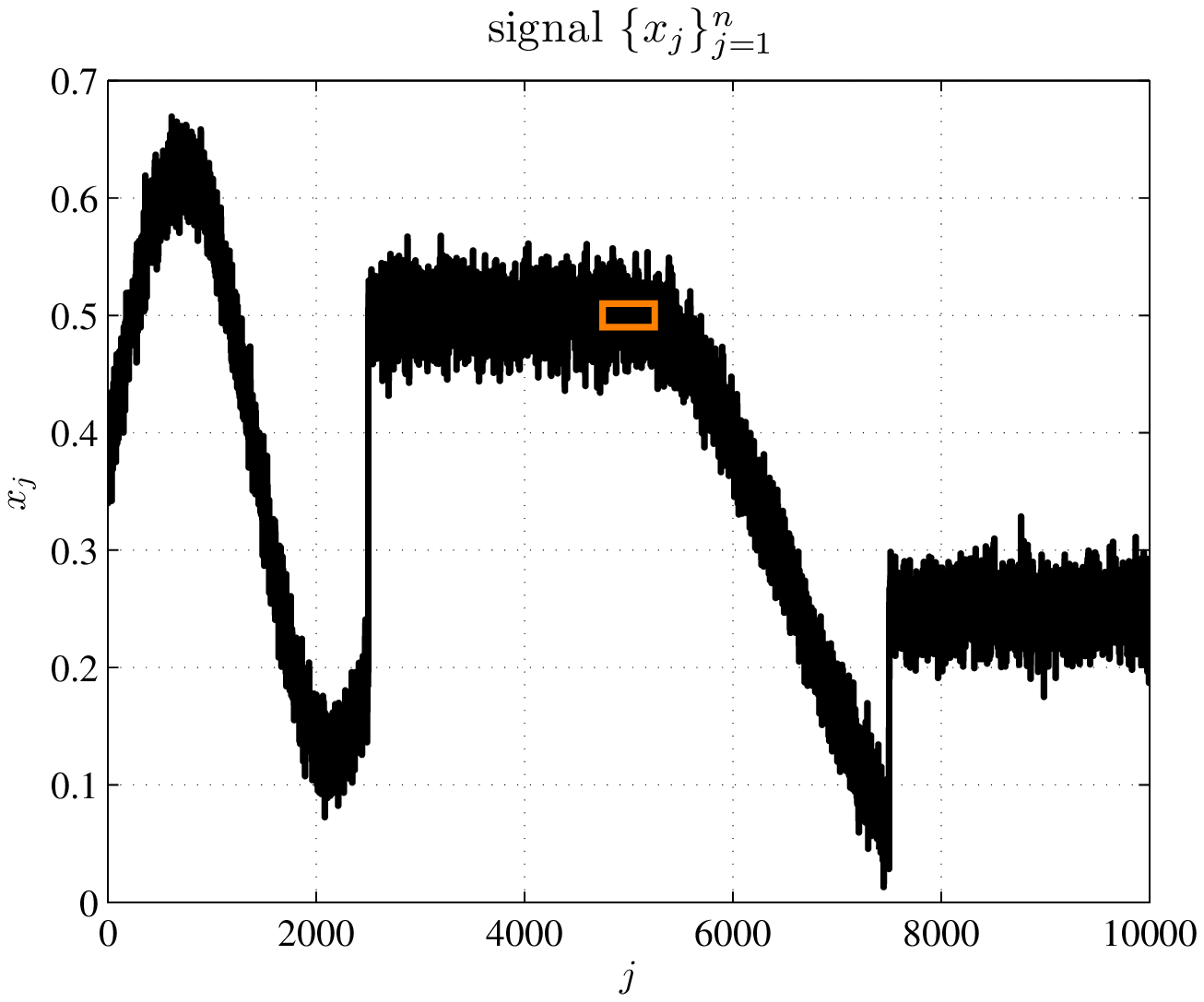}&
\includegraphics[width=0.45\linewidth]{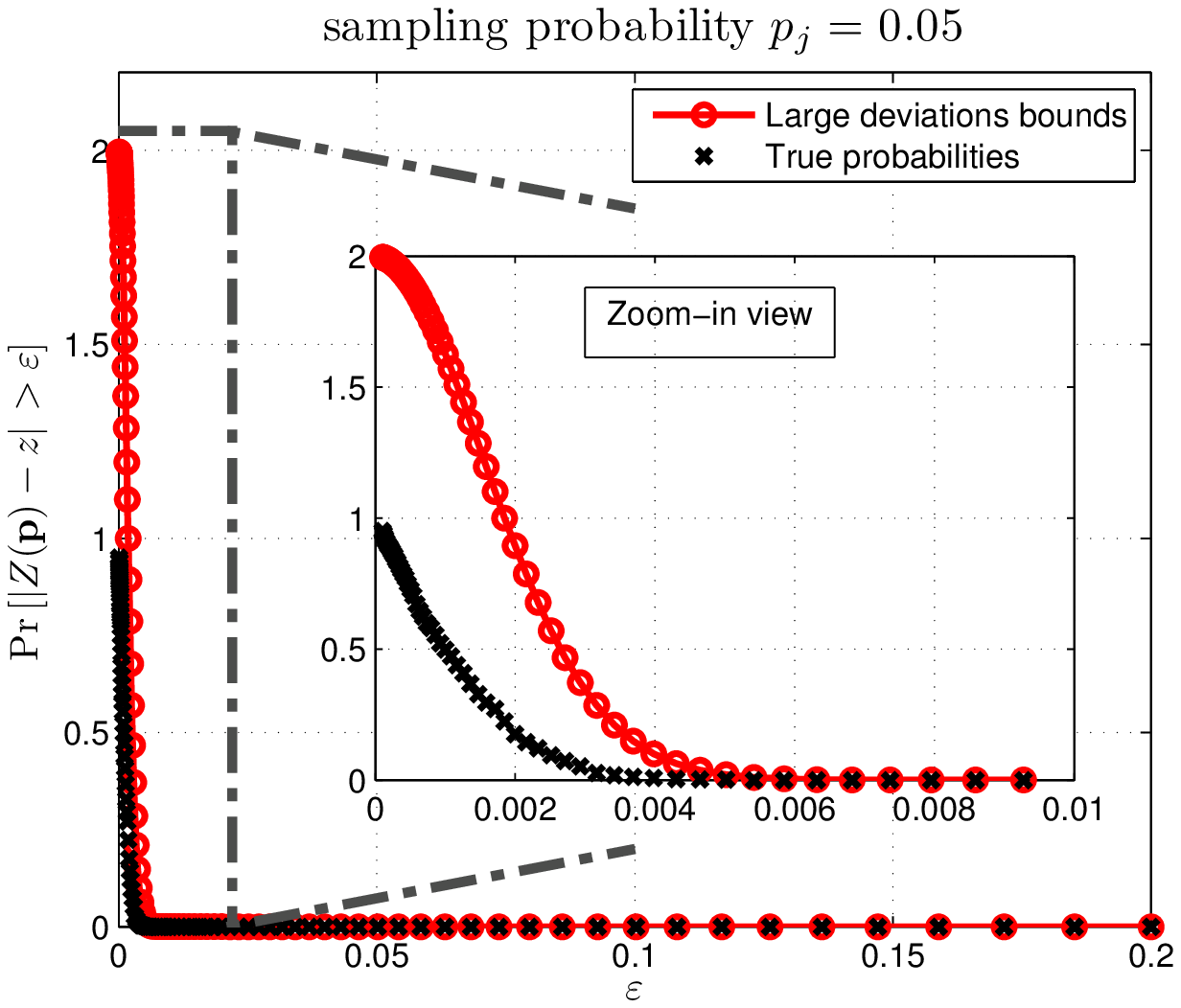}\\
(a) Noisy signal $\{x_j\}_{j=1}^n$ & (b) Probability $\Pr[|Z(\vp) - z| > \varepsilon]$
\end{tabular}
\caption{Example to illustrate \thref{thm:general_main}. (a) A one-dimensional signal with length $n = 10^4$, corrupted by  i.i.d. Gaussian noise with $\sigma = 5/255$. We use the MCNLM algorithm to denoise the signal. The patch size is $d = 5$ and the parameters are $h_r = 15/255$ and $h_s = \infty$, respectively. (b) The error probability as a function of $\varepsilon$. In this plot, the ``crosses'' denote the true probabilities as estimated by $10^5$ independent trials and the ``circles'' denote the analytical upper bound predicted by \thref{thm:general_main}. For easy comparisons, we also provide a zoomed-in version of the plot in the insert.}
\label{fig:example_thm1}
\end{figure*}

\begin{remark}
In a preliminary version of our work \cite{Chan_Zickler_Lu_2013}, we presented, based on the idea of martingales \cite{Serfling_1974}, an error probability bound for the special case when the sampling pattern is uniform. The result of \thref{thm:general_main} is more general and applies to any sampling patterns. We also note that the bound in \eref{thm1_main} quantifies the deviation of a ratio $Z(\vp) = A(\vp)/B(\vp)$, where the numerator and denominator are both weighted sums of independent random variables. It is therefore more general than the typical concentration bounds seen in the literature (see, \emph{e.g.}, \cite{Chung_Lu_2006,Drineas_Kanna_Mahoney_2006a}), where only a single weighted sum of random variables (\emph{i.e.}, either the numerator or the denominator) is considered.
\end{remark}

\begin{example}
To illustrate the result of \thref{thm:general_main}, we consider a one-dimensional signal as shown in \fref{example_thm1}(a). The signal $\{x_j\}_{j=1}^n$ is a piecewise continuous function corrupted by i.i.d. Gaussian noise. The noise standard deviation is $\sigma = 5/255$ and the signal length is $n=10^4$. We use MCNLM to denoise the $5001$-th pixel, and the sampling pattern is uniform with $p_j = \xi = 0.05$ for all $j$. For $\varepsilon = 0.01$, we can compute that $\frac{1}{n}\sum_{j=1}^n \alpha_j^2 = 1.335 \times 10^{-4}$, $\frac{1}{n}\sum_{j=1}^n \beta_j^2 = 1.452 \times 10^{-4}$, $\mu_B = 0.3015$, $M_\alpha = 0.458$, and $M_\beta = 0.617$. It then follows from \eref{thm1_main} that
\[
\Pr[\abs{Z(\vp) - z} > 0.01] \le 6.516 \times 10^{-6}.
\]
This bound shows that the random MCNLM estimate $Z(\vp)$, obtained by taking only $5\%$ of the samples, stays within one percent of the true NLM result $z$ with overwhelming probability. A complete range of results for different values of $\varepsilon$ are shown in \fref{example_thm1}(b), where we compare the true error probability as estimated by Monte Carlo simulations with the analytical upper bound predicted by \thref{thm:general_main}. We see from the ``zoomed-in'' portion of \fref{example_thm1}(b) that the analytical bound approaches the true probabilities for $\varepsilon \ge 0.005$.
\end{example}

\subsection{Special Case: Uniform Sampling Patterns}
Since the error probability bound in \eref{thm1_main} holds for all sampling patterns $\vp$, we will use \eref{thm1_main} to design optimal nonuniform sampling patterns in \sref{optimal_sampling}. But before we discuss that, we first consider the special case where $\vp$ is a uniform sampling pattern to provide a convenient and easily interpretable bound on the error probabilities.

\begin{proposition}[Uniform Sampling]
Assume that the sampling pattern is uniform, \emph{i.e.}, $\vp = \xi \vec{1}$. Then for every $\varepsilon > 0$ and every $0 < \xi \le 1$,
\begin{align}
\Pr\left[ \left| Z(\vp) - z \right| > \varepsilon \right] &\le \exp\left\{ -n\xi \right\} \notag \\
&+ 2 \exp\left\{ -n \mu_B f(\varepsilon) \xi \right\}, \label{eq:prop1_statement}
\end{align}
where $f(\varepsilon) \bydef \varepsilon^2 / \big(2(1+ \varepsilon)(1 + 7\varepsilon/6 )\big)$.
\label{proposition:uniform sampling}
\end{proposition}


To interpret \eref{prop1_statement}, we note that, for large $n$, the first term on the right-hand side of \eref{prop1_statement} is negligible. For example, when $n = 10^4$ and $\xi = 0.01$, we have $e^{-n \xi} = 3.7 \times 10^{-44}$. Thus, the error probability bound is dominated by the second term, whose negative exponent is determined by four factors:

\emph{1. The size of the reference set $\calX$.} If all other parameters are kept fixed or strictly bounded below by some positive constants, the error probability goes to zero as an exponential function of $n$. This shows that the random estimates obtained by MCNLM can be very accurate, when the size of the image (for internal denoising) or the size of the dictionary (for external denoising) is large.

\emph{2. Sampling ratio $\xi$.} To reduce the sampling ratio $\xi$ while still keeping the error probability small, a larger $n$, inversely proportional to $\xi$, is needed.

\emph{3. Precision $\varepsilon$.} Note that the function $f(\varepsilon)$ in \eref{prop1_statement} is of order $\mathcal{O}(\varepsilon^2)$ for small $\varepsilon$. Thus, with all other terms fixed, a $k$-fold reduction in $\varepsilon$ requires a $k^2$-fold increase in $n$ or $\xi$.

\emph{4. Patch redundancy $\mu_B$.} Recall that $\mu_B = \frac{1}{n}\sum_{j=1}^n w_j$, with the weights $\set{w_j}$ measuring the similarities between a noisy patch $\vy_i$ and all patches $\{\vx_j\}_{j=1}^n$ in the reference set $\calX$. Thus, $\mu_B$ serves as an indirect measure of the number of patches in $\calX$ that are similar to $\vy_i$. If $\vy_i$ can find many similar (redundant) patches in $\calX$, its corresponding $\mu_B$ will be large and so a relatively small $n$ will be sufficient to make the probability small; and vice versa.

Using the simplified expression in \eref{prop1_statement}, we derive in Appendix~\ref{appendix:prop_mse} the following upper bound on the mean squared error (MSE) of the MCNLM estimation:

\begin{proposition}[MSE]
Let the sampling pattern be uniform, with $\vp = \xi \vone$. Then for any $0 < \xi \le 1$,
\begin{equation}
\MSE_{\vp} \defequal \E_{\vp}\left[ \left( Z(\vp)-z \right)^2 \right] \le e^{-n\xi} + \frac{1}{n\xi} \left(\frac{52}{3\mu_B}\right) .
\label{eq:prop,mse,statement}
\end{equation}
\label{proposition:mse}
\end{proposition}

\begin{remark}
The above result indicates that, with a fixed average sampling ratio $\xi$ and if the patch redundancy $\mu_B$ is bounded from below by a positive constant, then the MSE of the MCNLM estimation converges to zero as $n$, the size of the reference set, goes to infinity.
\end{remark}


\begin{remark}
We note that the $\MSE_{\vp}$ stated in Proposition~\ref{proposition:mse} is a measure of the deviation between the randomized solution $Z(\vp)$ and the deterministic (full NLM) solution $z$. In other words, the expectation is taken over the different realizations of the sampling pattern, with the noise (and thus $z$) fixed. This is different from the standard MSE used in image processing (which we denote by $\MSE_\eta$), where the expectation is taken over different noise realizations.

To make this point more precise, we define $z^\ast$ as the ground truth noise free image, $Z(\vp, \eta)$ as the MCNLM solution using a random sampling pattern $\vp$ for a particular noise realization $\eta$. Note that the full NLM result can be written as $Z(\vone, \eta)$ (\emph{i.e.} when the sampling pattern $\vp$ is the all-one vector.) We consider the following two quantities:
\begin{equation}
\MSE_\eta = \E_{\eta}[(Z(\vone,\eta)-z^*)^2]
\end{equation}
and
\begin{equation}
\MSE_{\eta, \vp} = \E_{\eta, \vp}[(Z(\vp,\eta)-z^*)^2]
\end{equation}
The former is the MSE achieved by the full NLM, whereas the latter is the MSE achieved by the proposed MCNLM.
While we do not have theoretical bounds linking $\MSE_\eta$ to $\MSE_{\eta, \vp}$ (as doing so would require the knowledge of the ground truth image $z^\ast$), we refer the reader to Table~\ref{table:standard 21x21} in Sec. \ref{sec:experiment}, where numerical simulations show that, even for relatively small sampling ratios $\xi$, the MSE  achieved by MCNLM stays fairly close to the MSE achieved by the full NLM.

\end{remark}


\section{Optimal Sampling Patterns}
\label{sec:optimal_sampling}

While the uniform sampling scheme (\emph{i.e.}, $\vp = \xi\vone$) allows for easy analysis and provides useful insights, the performance of the proposed MCNLM algorithm can be significantly improved by using properly chosen nonuniform sampling patterns. We present the design of such patterns in this section.

\subsection{Design Formulation}
The starting point of seeking an optimal sampling pattern is the general probability bound provided by Theorem~\ref{thm:general_main}. A challenge in applying this probability bound in practice is that the right-hand side of \eref{thm1_main} involves the complete set of weights $\set{w_j}$ and the full NLM result $z$. One can of course compute these values, but doing so will defeat the purpose of random sampling, which is to speed up NLM by \emph{not} computing all the weights $\set{w_j}$. To address this problem, we assume that
\begin{equation}\label{eq:w_bounds}
0 < w_j \le b_j \le 1,
\end{equation}
where the upper bounds $\set{b_j}$ are either known \emph{a priori} or can be efficiently computed. We will provide concrete examples of such upper bounds in \sref{pattern_examples}. For now, we assume that the bounds $\set{b_j}$ have already been obtained.

Using \eref{w_bounds} and noting that $0 \le x_j, z \le 1$ (and thus $|x_j - z| \le 1$), we can see that the parameters $\set{\alpha_j, \beta_j}$ in \eref{thm1_main} are bounded by
\[
\abs{\alpha_j} \le b_j (1 + \varepsilon) \quad\text{and}\quad \abs{\beta_j} \le b_j (1 + \varepsilon),
\]
respectively. It then follows from \eref{thm1_main} that
\begin{align}
&\Pr\left[ \abs{Z(\vp) - z} > \varepsilon \right]\le \exp\set{-n\xi}\nonumber\\
&+2\exp\left\{\frac{-n (\mu_B \varepsilon)^2 / (1+\varepsilon)^2}{ 2 \left(\frac{1}{n}\sum_{j=1}^n b_j^2 \left( \frac{1-p_j}{p_j}\right) +  M \max\limits_{1\le j \le n} \left( \frac{b_j}{p_j} \right) \right)}\right\},  \label{eq:sampling_main}
\end{align}
where $M \bydef \left(\mu_B \varepsilon \right) / (6(1+\varepsilon)) $.

Given the average sampling ratio $\xi$, we seek sampling patterns $\vp$ to minimize the probability bound in \eref{sampling_main}, so that the random MCNLM estimate $Z(\vp)$ will be tightly concentrated around the full NLM result $z$. Equivalently, we solve the following optimization problem.

\begin{equation}\label{eq:sampling_P0}
(P): \begin{array}{ll}
\argmin{\vp}   &\ \ \frac{1}{n}\sum_{j=1}^n b_j^2 \left( \frac{1-p_j}{p_j}\right) + M \max\limits_{1\le j \le n} \left( \frac{b_j}{p_j} \right)\\
\subjectto     &\ \ \frac{1}{n} \sum\limits_{j=1}^n p_j = \xi  \ \mathrm{and}\  0 < p_j \le 1.
\end{array}
\end{equation}

The optimization formulated has a closed-form solution stated as below. The derivation is given in Appendix~\ref{appendix:thm,pj}.
\begin{theorem}[Optimal Sampling Patterns]
\label{thm:optimal p_j}
The solution to $(P)$ is given by
\begin{equation}
\label{eq:thm,optimal p_j,solution}
p_j = \max( \min(b_j \tau, 1), b_j/t), \quad \text{for } 1 \le j \le n,
\end{equation}
where $t \bydef \max\left( \frac{1}{n\xi}\sum_{j=1}^n b_j,\; \max\limits_{1 \le j\le n} b_j \right)$, and the parameter $\tau$ is chosen so that $\sum_j p_j = n \xi$.
\end{theorem}

\begin{remark}
It is easy to verify that the function
\begin{equation}\label{eq:f_tau}
g(x) = \sum_{j=1}^n \max( \min(b_j x, 1), b_j/t) - n \xi
\end{equation}
is a piecewise linear and monotonically increasing function. Moreover, $g(+\infty) = n(1-\xi) > 0$ and
$$
g(0) = \sum_{j=1}^n \frac{ b_j}{t} - n\xi \le \frac{\sum_{j=1}^n b_j}{\frac{1}{n\xi} \sum_{j=1}^n b_j} - n\xi = 0.
$$
Thus, $\tau$ can be uniquely determined as the root of $g(\tau)$.
\end{remark}

\begin{remark}
The cost function of ($P$) contains a quantity $M = \left(\mu_B \varepsilon \right) / (6(1+\varepsilon))$. One potential issue is that the two parameters ($\mu_B$ and $\varepsilon$) that are not necessarily known to the algorithm. However, as a remarkable property of the solution given in Theorem 2, the optimal sampling pattern $\vp$ does \emph{not} depend on $M$. Thus, only a single parameter, namely, the average sampling ratio $\xi$, will be needed to fully specify the optimal sampling pattern in practice.
\end{remark}

\subsection{Optimal Sampling Patterns}
\label{sec:pattern_examples}

To construct the optimal sampling pattern prescribed by Theorem~\ref{thm:optimal p_j}, we need to find $\set{b_j}$, which are the upper bounds on the true similarity weights $\set{w_j}$. At one extreme, the tightest upper bounds are $b_j = w_j$, but this oracle scheme is not realistic as it requires that we know all the weights $\set{w_j}$. At the other extreme, we can use the trivial upper bound $b_j = 1$. It is easy to verify that, under this setting, the sampling pattern in \eref{thm,optimal p_j,solution} becomes the uniform pattern, \emph{i.e.}, $p_j = \xi$ for all $j$. In what follows, we present two choices for the upper bounds that can be efficiently computed and that can utilize partial knowledge of $w_j$.

\subsubsection{Bounds from spatial information} The first upper bound is designed for internal (\emph{i.e.}, single image) denoising where there is often a spatial term in the similarity weight, \emph{i.e.},
\begin{equation}\label{eq:weight_spatial_intensity}
w_j = w^{s}_j \, w^{r}_j.
\end{equation}
One example of the spatial weight can be found in \eref{spatial_weight}. Since $w^{r}_j \le 1$, we always have $w_j \le w^{s}_j$. Thus, a possible choice is to set
\begin{equation}\label{eq:spatial_bound}
b_j^{s} = w^{s}_j.
\end{equation}
The advantage of the above upper bound is that $b_j^s$ is a function of the spatial distance $d_{i,j}$ between a pair of pixels, which is independent of the image data $\calX$ and $\calY$. Therefore, it can be pre-computed before running the MCNLM algorithm. Moreover, since $\{b_j\}$ is \emph{spatially invariant}, they can be reused at all pixel locations.

\subsubsection{Bounds from intensity information} For external image denoising, the patches in $\calX$ and $\calY$ do not have any spatial relationship, as they can come from different images. In this case, the similarity weight $w_j$ is only due to the difference in pixel intensities (\emph{i.e.}, $w_j = w_j^r$), and thus we cannot use the spatial bounds given in \eref{spatial_bound}. To derive a new bound for this case, we first recall the Cauchy-Schwartz inequality: For any two vectors $\vu, \vv \in \R^d$ and for any positive-definite weight matrix $\mLambda \in \R^{d\times d}$, it holds that
\[
\abs{\vu^T \mLambda \vv} \le \norm{\vu}_{\mLambda} \, \norm{\vv}_{\mLambda}.
\]
Setting $\vu = \vy - \vx_j$, we then have
\begin{align}
w_j = e^{-\|\vy - \vx_j\|_{\mLambda}^2/(2h_r^2)}
&\le e^{-\left( (\vx_j - \vy)^T \mLambda \vv \right)^2 / \left({2h_r^2 \norm{\vv}_{\mLambda}^2}\right)} \notag \\
&\le e^{-\left(\vx_j^T \vs - \vy^T \vs\right)^2} = b_j^{r}, \label{eq:mean sampling bound}
\end{align}
where $\vs \bydef \mLambda \vv / \left(\sqrt{2} h_r \norm{\vv}_{\mLambda}\right)$. The vector $\vv$ can be any nonzero vector. In practice, we choose $\vv = \vone$ with $\mLambda = \diag{1/d,\ldots,1/d}$ and we find this choice effective in our numerical experiments. In this case, $b_j^r = \exp\left\{-(\vx_j^T\vone - \vy^T\vone)^2/(2d^2h_r^2)\right\}$.

\begin{remark}
To obtain the upper bound $b_j^r$ in \eref{mean sampling bound}, we need to compute the terms $\vy^T\vs$ and $\vx_j^T\vs$, which are the projections of the vectors $\vy$ and $\vx_j$ onto the one-dimensional space spanned by $\vs$. These projections can be efficiently computed by convolving the noisy image and the images in the reference set with a spatially-limited kernel corresponding to $\vs$. To further reduce the computational complexity, we also adopt a two-stage importance sampling procedure in our implementation, which allows us to avoid the computation of the exact values of $\set{b_j}$ at most pixels. Details of our implementation are given in a supplementary technical report \cite{Chan_Zickler_Lu_2013_TR}.
\end{remark}

\begin{figure}[t]
  \begin{minipage}{1\linewidth}
    \centering
    \includegraphics[width=0.4\linewidth]{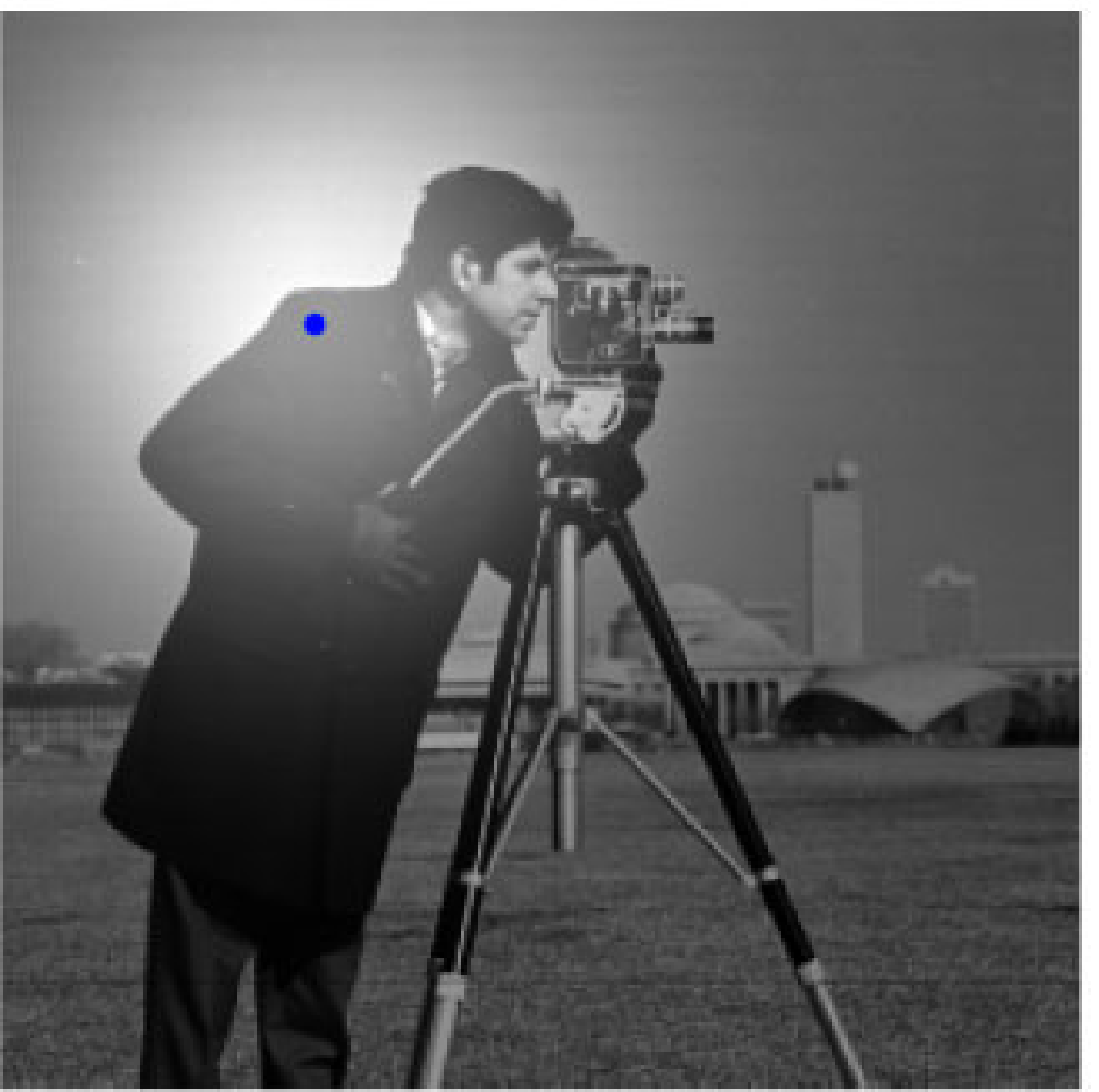}\\
    (a) Target pixel to be denoised
  \end{minipage}\hfill
  \begin{minipage}{0.5\linewidth}
    \centering
    \includegraphics[width=1\linewidth]{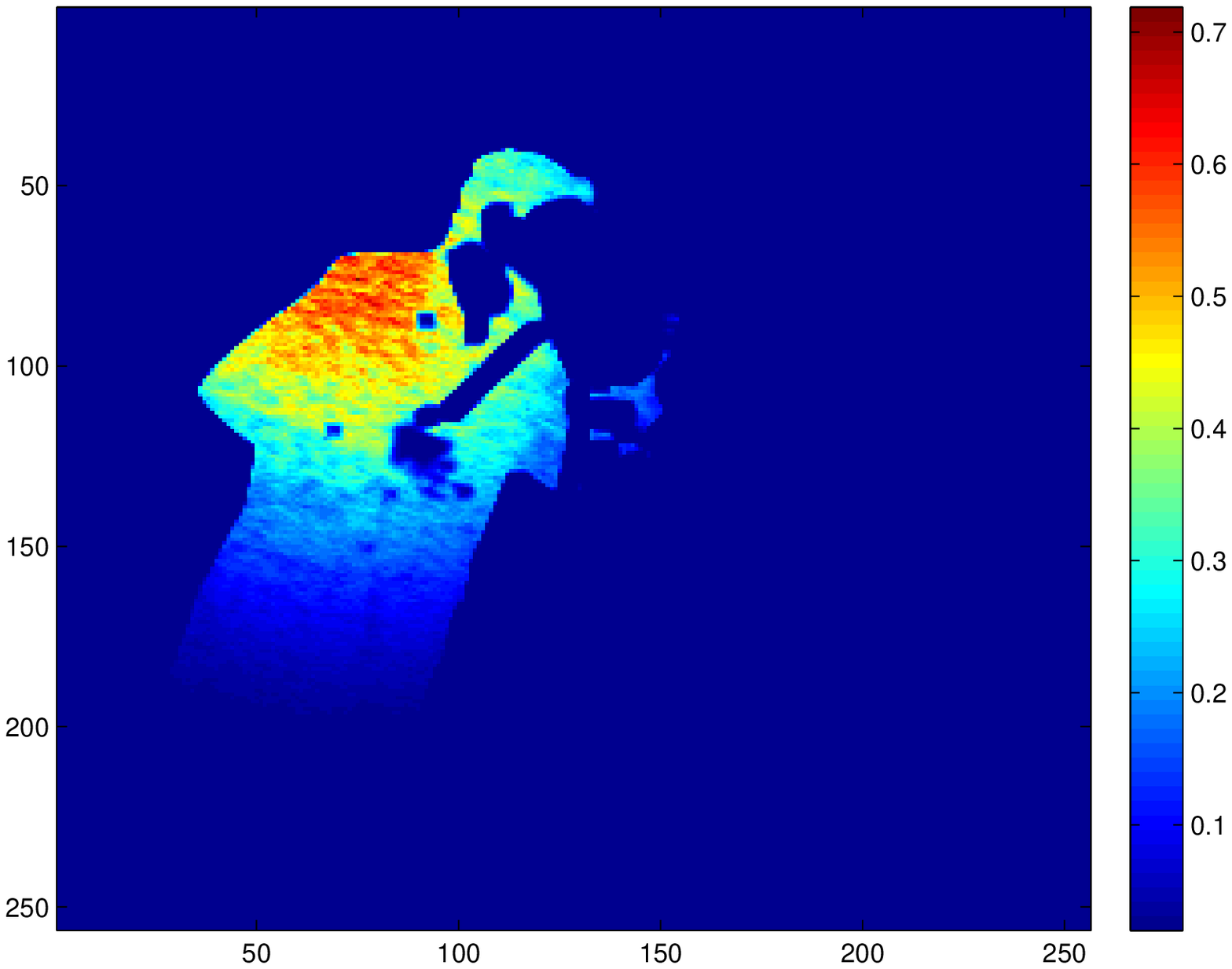}
    (b) Oracle sampling pattern
    \includegraphics[width=1\linewidth]{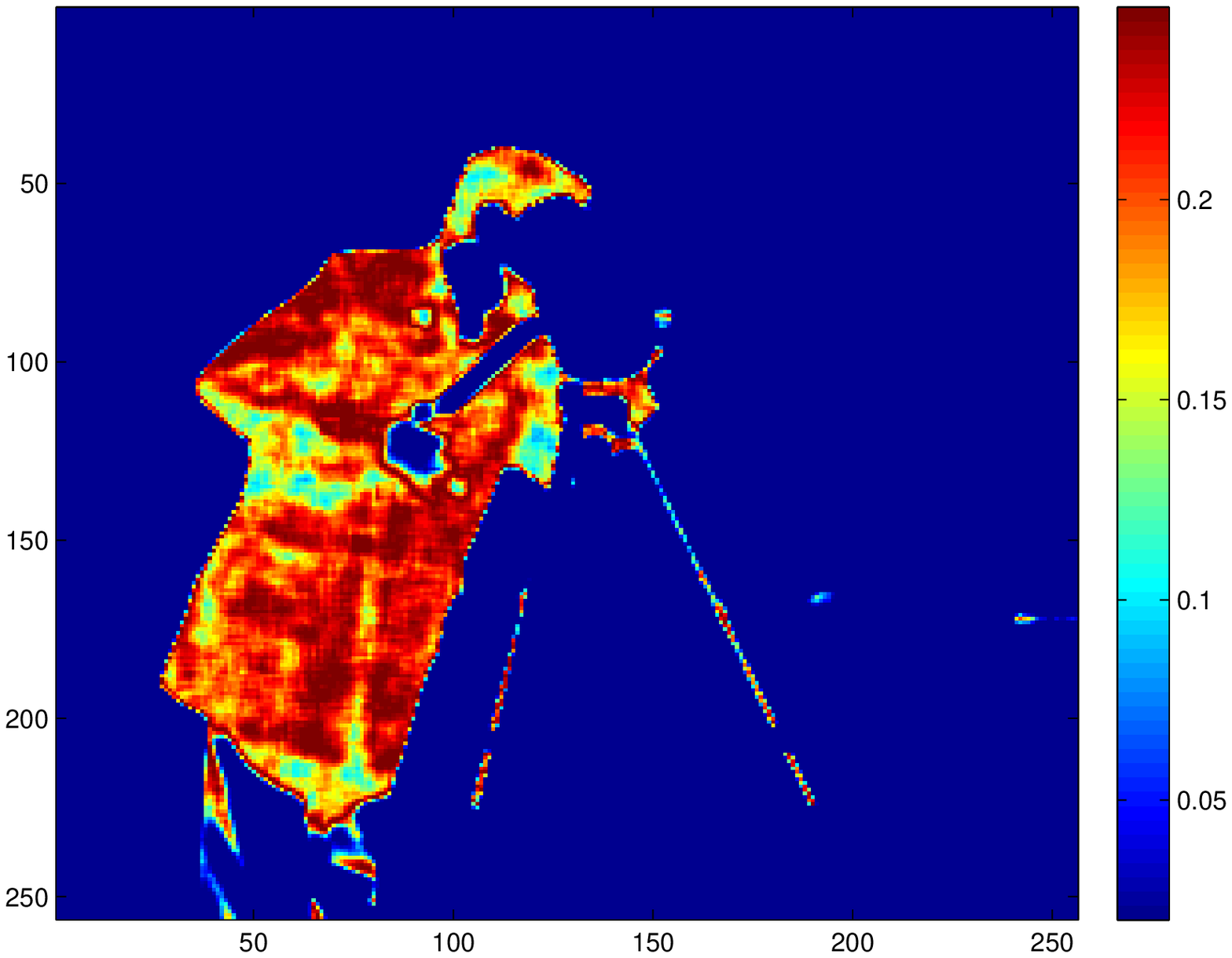}
    (d) Intensity
  \end{minipage}\hfill
  \begin{minipage}{0.5\linewidth}
    \centering
    \includegraphics[width=1\linewidth]{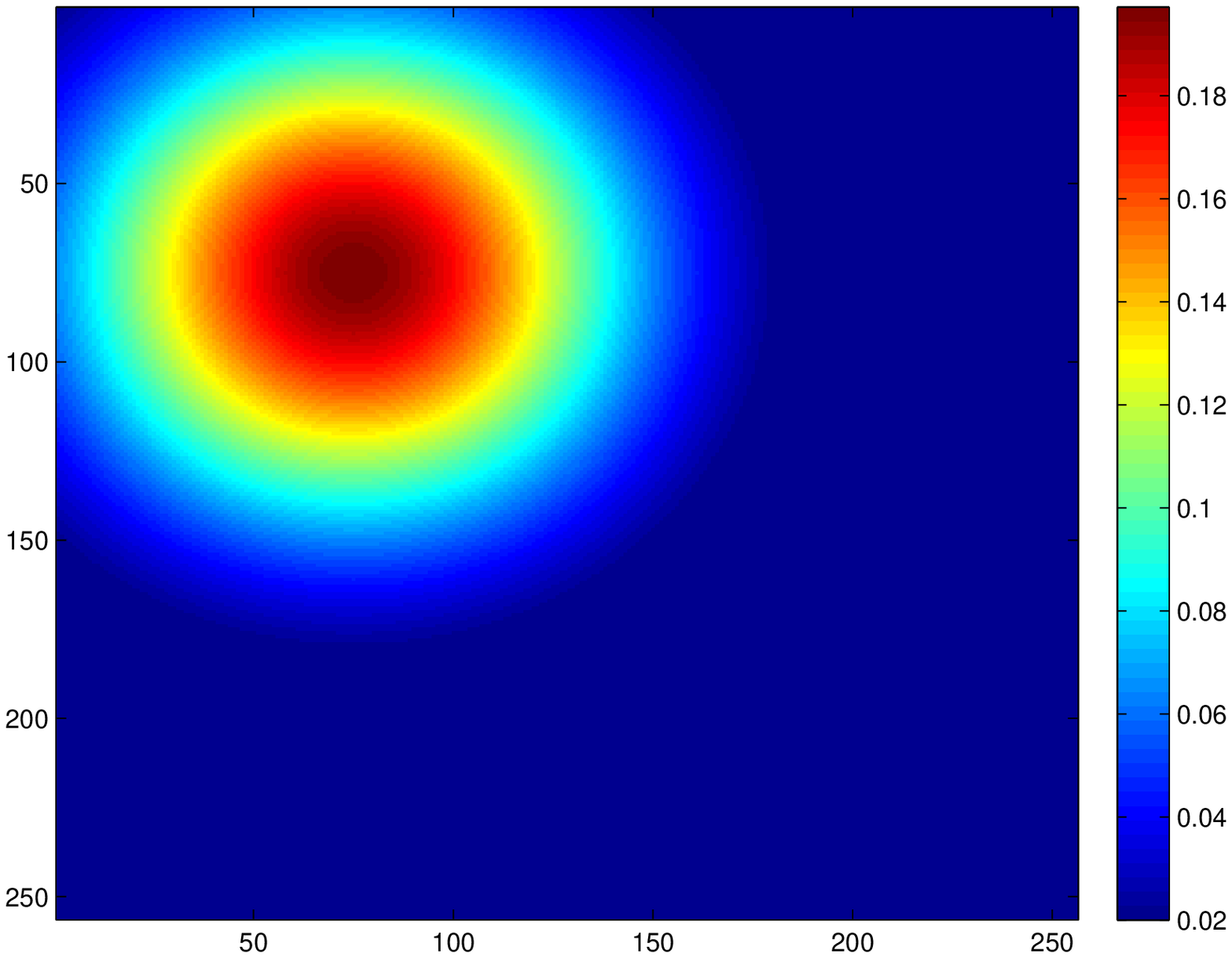}
    (c) Spatial
    \includegraphics[width=1\linewidth]{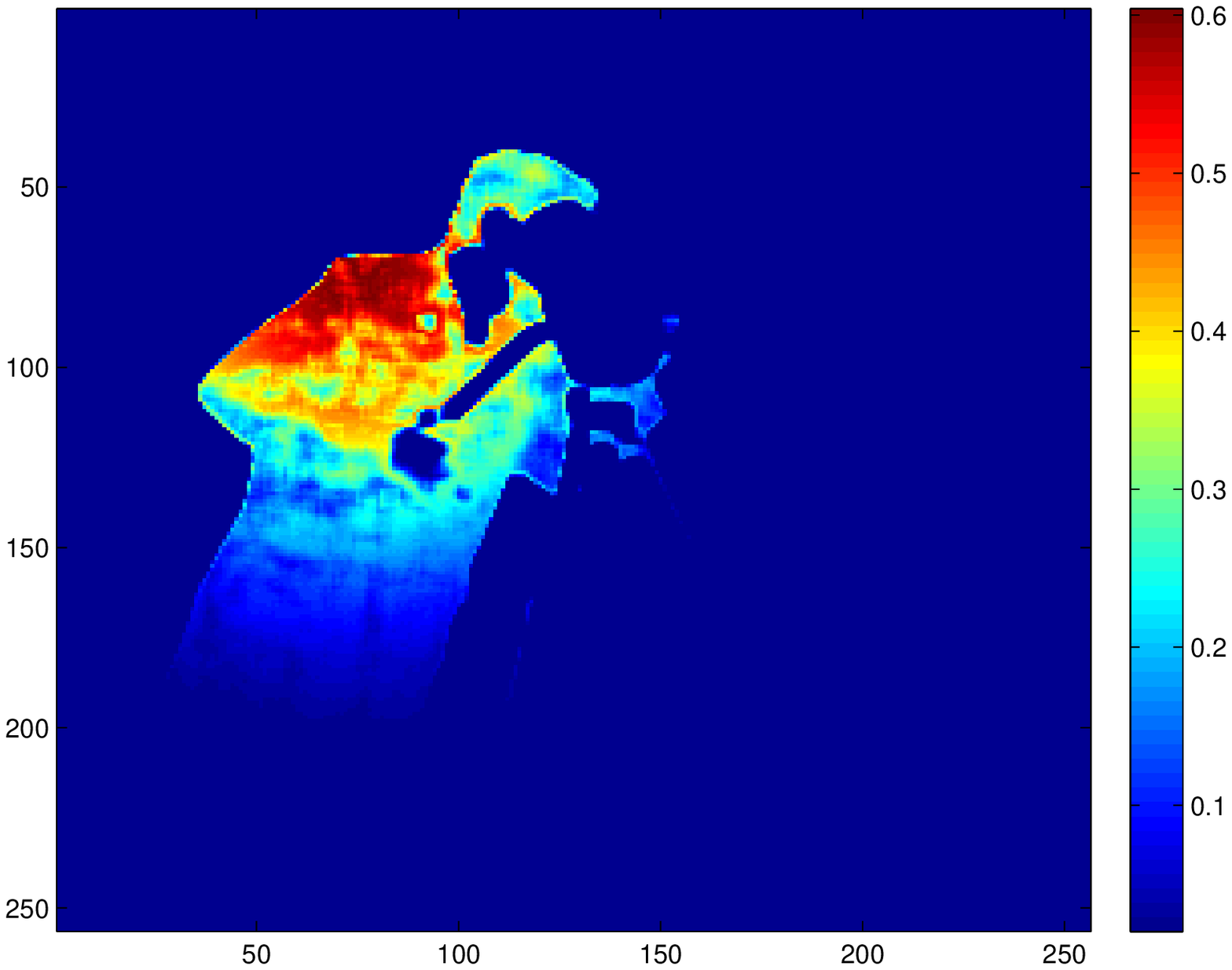}
    (e) Spatial + Intensity
  \end{minipage}\hfill
\caption{Illustration of optimal sampling probability for the case $h_r = 15/255$, $h_s = 50$. (a) Cameraman image and the target pixel. We overlay the spatial weight on top of \emph{cameraman} for visualization. (b) Optimal sampling pattern w.r.t. $w_j$ (oracle scheme). (c) Spatial upper bound $b_j^{s}$. (d) Intensity upper bound $b_j^{r}$. (e) Spatial and intensity upper bound $b_j^{s}\cdot b^{r}_j$.}
\label{fig:optimal_sampling}
\end{figure}

\begin{figure}[t]
\centering
\includegraphics[width=1\linewidth]{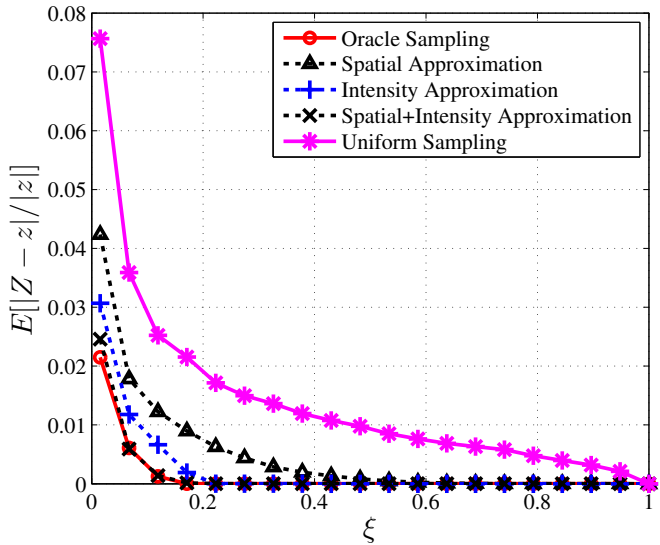}
\caption{Denoising results of using different sampling schemes shown in \fref{optimal_sampling}. Setting of experiment: noise $\sigma = 15/255$, $h_s = 50$, $h_r = 15/255$, patch size $5 \times 5$.}
\label{fig:approx_comparison}
\end{figure}

\begin{remark}
Given the oracle sampling pattern, it is possible to improve the performance of NLM by \emph{deterministically} choosing the weights according to the oracle sampling pattern. We refer the reader to \cite{Kervrann_Boulanger_2006}, where similar approaches based on spatial adaptations were proposed.
\end{remark}

\begin{example} To demonstrate the performance of the various sampling patterns presented above, we consider denoising one pixel of the \emph{Cameraman} image as shown in \fref{optimal_sampling}(a). The similarity weights are in the form of \eref{weight_spatial_intensity}, consisting of both a spatial and a radiance-related part. Applying the result of Theorem~\ref{thm:optimal p_j}, we derive four optimal sampling patterns, each associated with a different choice of the upper bound, namely, $b_j = w_j, b_j = b_j^{s}, b_j = b_j^{r}$, and $b_j = b_j^{s} b_j^{r}$. Note that the first choice corresponds to an \emph{oracle} setting, where we assume that the weights $\set{w_j}$ are known. The latter three are practically achievable sampling patterns, where $b_j^{s}$ and $b_j^{r}$ are defined in \eref{spatial_bound} and \eref{mean sampling bound}, respectively.

\fref{optimal_sampling}(b)--(e) show the resulting sampling patterns. As can be seen in the figures, various aspects of the oracle sampling pattern are reflected in the approximated patterns. For instance, the spatial approximation has more emphasis at the center than the peripherals whereas the intensity approximation has more emphasis on pixels that are similar to the target pixel.

To compare these sampling patterns quantitatively, we plot in \fref{approx_comparison} the reconstruction relative error associated with different patterns as functions of the average sampling ratio $\xi$. Here, we set $h_r = 15/255$ and $h_s = 50$. For benchmark, we also show the performance of the uniform sampling pattern. It is clear from the figure that all the optimal sampling patterns outperform the uniform pattern. In particular, the pattern obtained by incorporating both the spatial and intensity information approaches the performance of the oracle scheme.
\end{example}


\pagebreak

\section{Experimental Results}
\label{sec:experiment}
In this section we present additional numerical experiments to evaluate the performance of the MCNLM algorithm and compare it with several other accelerated NLM algorithms.

\subsection{Internal Denoising}

A benchmark of ten standard test images are used for this experiment. For each image, we add zero-mean Gaussian noise with standard deviations equal to $\sigma = \frac{10}{255},\frac{20}{255},\frac{30}{255},\frac{40}{255},\frac{50}{255}$ to simulate noisy images at different PSNR levels. Two choices of the spatial search window size are used: $21 \times 21$ and $35 \times 35$, following the original configurations used in \cite{Buades_Coll_2005_Journal}.

The parameters of MCNLM are as follows: The patch size is $5 \times 5$ (\emph{i.e.}, $d = 25$) and $\mLambda = \mI/d$. For each choice of the spatial search window size (\emph{i.e.}, $\rho = 21$ or $\rho = 35$), we define $h_s = (\lfloor \rho/2 \rfloor)/3$ so that three standard deviations of the spatial Gaussian will be inside the spatial search window. The intensity parameter is set to $h_r = 1.3\sigma/255$.

In this experiment, we use the spatial information bound \eref{spatial_bound} to compute the optimal sampling pattern in \eref{thm,optimal p_j,solution}. Incorporating additional intensity information as in \eref{mean sampling bound} would further improve the performance, but we choose not to do so because the PSNR gains are found to be moderate in this case due to the relatively small size of the spatial search window. Five average sampling ratios, $\xi = 0.05, 0.1, 0.2, 0.5, 1$, are evaluated. We note that when $\xi = 1$, MCNLM is identical to the full NLM.

For comparisons, we test the Gaussian KD tree (GKD) algorithm \cite{Adams_Gelfand_2009} with a C++ implementation (ImageStack \cite{ImageStack}) and the adaptive manifold (AM) algorithm \cite{Gastal_Oliveira_2012} with a MATLAB implementation provided by the authors. To create a meaningful common ground for comparison, we adapt MCNLM as follows: First, since both GKD and AM use SVD projection \cite{VanDeVille_Kocher_2010} to reduce the dimensionality of patches, we also use in MCNLM  the same SVD projection method by computing the 10 leading singular values. The implementation of this SVD step is performed using an off-the-shelf MATLAB code \cite{Peyre_2009}.  We also tune the major parameters of GKD and AM for their best performance, \emph{e.g.}, for GKD we set $h_r = 1.3 \sigma/255$ and for AM we set $h_r = 2\sigma/255$. Other parameters are kept at their default values as reported in \cite{Adams_Gelfand_2009}, \cite{Gastal_Oliveira_2012}. For completeness, we also show the results of BM3D \cite{Dabov_Foi_Katkovnik_2007}.

\tref{table:standard 21x21} and \tref{table:average_psnr} summarize the results of the experiment. Additional results, with visual comparison of denoised images, can be found in the supplementary technical report \cite{Chan_Zickler_Lu_2013_TR}. Since MCNLM is a randomized algorithm, we report the average PSNR values of MCNLM over 24 independent runs using random sampling pattern and 20 independent noise realizations. The standard deviations of the PSNR values over the 24 random sampling patterns are shown in Table~\ref{table:average_psnr}. The results show that for the 10 images, even at a very low sampling ratio, \emph{e.g.}, $\xi = 0.1$, the averaged performance of MCNLM (over 10 testing images) is only about $0.35$ dB to $0.7$ dB away (depending on $\sigma$) from the full NLM solution. When the sampling ratio is further increased to $\xi = 0.2$, the PSNR values become very close (about a $0.09$ dB to $0.2$ dB drop depending on $\sigma$) to those of the full solution.

In \tref{table:run time} we report the runtime of MCNLM, GKD and AM. Since the three algorithms are implemented in different environments, namely, MCNLM in MATLAB/C++ (.mex), GKD in C++ with optimized library and data-structures, and AM in MATLAB (.m), we caution that \tref{table:run time} is only meant to provide some rough references on computational times. For MCNLM, its speed improvement over the full NLM can be reliably estimated by the average sampling ratio $\xi$.

We note that the classical NLM algorithm is no longer the state-of-the-art in image denoising. It has been outperformed by several more recent approaches, \emph{e.g.}, BM3D \cite{Dabov_Foi_Katkovnik_2007} (See Table I and II) and global image denoising \cite{Talebi_Milanfar_2013}. Thus, for internal (\emph{i.e.}, single-image) denoising, the contribution of MCNLM is mainly of a theoretical nature: It provides the theoretical foundation and a proof-of-concept demonstration to show the effectiveness of a simple random sampling scheme to accelerate the NLM algorithm. More work is needed to explore the application of similar ideas to more advanced image denoising algorithms.

\begin{table*}[!]
\small
\setlength{\extrarowheight}{1.2pt}
\setlength{\tabcolsep}{5pt}
\centering
\caption{Single image denoising by MCNLM, using the optimal Gaussian sampling pattern. The case when $\xi = 1$ is equivalent to the standard NLM \cite{Buades_Coll_2005_AVSS}. GKD refers to \cite{Adams_Gelfand_2009}. AM refers to \cite{Gastal_Oliveira_2012}. BM3D refers to \cite{Dabov_Foi_Katkovnik_2007}. Shown in the table are PSNR values (in dB). The results of MCNLM is averaged over 24 independent trials of using different sampling patterns, and over 20 independent noise realizations.}
%
\begin{tabular}{|c|cccccccc||cccccccc|}
\hline
    $\xi $& 0.05 & 0.1     & 0.2      &   0.5     &    1     & GKD & AM & BM3D &   0.05 & 0.1     &   0.2 &       0.5     &    1      & GKD  & AM  & BM3D \\
\hline
$\sigma$  & \multicolumn{8}{c||}{\emph{Baboon} $512 \times 512$} & \multicolumn{8}{c|}{\emph{Barbara} $512 \times 512$} \\
\hline
10	 & 	30.70	 & 	31.20	 & 	31.56	 & 	31.60	 & 	31.60	 & 	31.13	 & 	28.88	 & 	33.14	 & 	32.14	 & 	32.68	 & 	 33.05	 & 	33.19	 & 	33.19	 & 	32.72	 & 	30.47	 & 	34.95	\\
20	 & 	26.85	 & 	27.12	 & 	27.23	 & 	27.31	 & 	27.31	 & 	26.68	 & 	25.80	 & 	29.07	 & 	28.19	 & 	28.76	 & 	 29.09	 & 	29.24	 & 	29.24	 & 	28.38	 & 	26.87	 & 	31.74	\\
30	 & 	24.59	 & 	24.86	 & 	25.01	 & 	25.10	 & 	25.10	 & 	24.59	 & 	24.28	 & 	26.83	 & 	25.82	 & 	26.37	 & 	 26.71	 & 	26.85	 & 	26.86	 & 	25.90	 & 	24.91	 & 	29.75	\\
40	 & 	23.17	 & 	23.58	 & 	23.81	 & 	23.92	 & 	23.93	 & 	23.27	 & 	23.37	 & 	25.26	 & 	24.14	 & 	24.75	 & 	 25.12	 & 	25.27	 & 	25.27	 & 	24.28	 & 	23.77	 & 	28.05	\\
50	 & 	22.16	 & 	22.71	 & 	23.02	 & 	23.16	 & 	23.16	 & 	22.29	 & 	22.70	 & 	24.21	 & 	22.85	 & 	23.52	 & 	 23.92	 & 	24.08	 & 	24.08	 & 	23.08	 & 	22.96	 & 	26.85	\\
\hline
$\sigma$  & \multicolumn{8}{c||}{\emph{Boat} $512 \times 512$}     & \multicolumn{8}{c|}{\emph{Bridge} $512 \times 512$} \\
\hline
10	 & 	32.16	 & 	32.58	 & 	32.86	 & 	32.93	 & 	32.93	 & 	32.51	 & 	30.94	 & 	33.90	 & 	29.47	 & 	29.25	 & 	 29.03	 & 	29.07	 & 	29.07	 & 	29.61	 & 	28.49	 & 	30.71	\\
20	 & 	28.63	 & 	29.23	 & 	29.58	 & 	29.70	 & 	29.70	 & 	28.68	 & 	28.00	 & 	30.84	 & 	25.41	 & 	25.36	 & 	 25.33	 & 	25.38	 & 	25.38	 & 	25.68	 & 	25.41	 & 	26.75	\\
30	 & 	26.47	 & 	27.15	 & 	27.53	 & 	27.68	 & 	27.68	 & 	26.46	 & 	26.11	 & 	29.02	 & 	23.60	 & 	23.72	 & 	 23.81	 & 	23.88	 & 	23.88	 & 	23.89	 & 	23.72	 & 	24.99	\\
40	 & 	24.82	 & 	25.58	 & 	26.03	 & 	26.20	 & 	26.21	 & 	24.89	 & 	24.82	 & 	27.60	 & 	22.33	 & 	22.59	 & 	 22.76	 & 	22.84	 & 	22.84	 & 	22.68	 & 	22.63	 & 	23.87	\\
50	 & 	23.50	 & 	24.35	 & 	24.86	 & 	25.06	 & 	25.06	 & 	23.66	 & 	23.86	 & 	26.36	 & 	21.36	 & 	21.73	 & 	 21.95	 & 	22.06	 & 	22.06	 & 	21.75	 & 	21.86	 & 	22.96	\\
\hline
$\sigma$  & \multicolumn{8}{c||}{\emph{Couple} $512 \times 512$}  & \multicolumn{8}{c|}{\emph{Hill} $256 \times 256$}\\
\hline
10	 & 	31.97	 & 	32.39	 & 	32.65	 & 	32.72	 & 	32.72	 & 	32.39	 & 	30.85	 & 	34.01	 & 	30.54	 & 	30.48	 & 	 30.41	 & 	30.46	 & 	30.46	 & 	30.85	 & 	30.10	 & 	31.88	\\
20	 & 	28.14	 & 	28.56	 & 	28.78	 & 	28.88	 & 	28.89	 & 	28.19	 & 	27.58	 & 	30.70	 & 	26.98	 & 	27.12	 & 	 27.19	 & 	27.26	 & 	27.26	 & 	27.17	 & 	26.95	 & 	28.55	\\
30	 & 	25.91	 & 	26.41	 & 	26.69	 & 	26.82	 & 	26.82	 & 	25.99	 & 	25.77	 & 	28.74	 & 	25.11	 & 	25.45	 & 	 25.65	 & 	25.75	 & 	25.75	 & 	25.34	 & 	25.36	 & 	26.93	\\
40	 & 	24.36	 & 	25.00	 & 	25.37	 & 	25.52	 & 	25.52	 & 	24.51	 & 	24.58	 & 	27.29	 & 	23.83	 & 	24.34	 & 	 24.65	 & 	24.78	 & 	24.78	 & 	24.09	 & 	24.33	 & 	25.82	\\
50	 & 	23.18	 & 	23.94	 & 	24.40	 & 	24.58	 & 	24.58	 & 	23.37	 & 	23.71	 & 	26.07	 & 	22.81	 & 	23.48	 & 	 23.87	 & 	24.02	 & 	24.02	 & 	23.11	 & 	23.56	 & 	24.89	\\
\hline
$\sigma$  & \multicolumn{8}{c||}{\emph{House} $256 \times 256$}   & \multicolumn{8}{c|}{\emph{Lena} $512 \times 512$}\\
\hline
10	 & 	33.95	 & 	34.77	 & 	35.35	 & 	35.48	 & 	35.48	 & 	34.46	 & 	33.03	 & 	36.70	 & 	34.76	 & 	35.54	 & 	 36.02	 & 	36.15	 & 	36.15	 & 	34.90	 & 	34.03	 & 	37.04	\\
20	 & 	30.37	 & 	31.53	 & 	32.26	 & 	32.48	 & 	32.48	 & 	30.37	 & 	29.57	 & 	33.82	 & 	30.98	 & 	31.94	 & 	 32.52	 & 	32.72	 & 	32.72	 & 	30.96	 & 	30.52	 & 	33.95	\\
30	 & 	27.89	 & 	29.05	 & 	29.78	 & 	30.03	 & 	30.03	 & 	27.80	 & 	27.24	 & 	32.13	 & 	28.44	 & 	29.55	 & 	 30.24	 & 	30.48	 & 	30.49	 & 	28.53	 & 	28.44	 & 	31.83	\\
40	 & 	26.04	 & 	27.25	 & 	28.01	 & 	28.28	 & 	28.29	 & 	26.07	 & 	25.78	 & 	30.80	 & 	26.53	 & 	27.73	 & 	 28.48	 & 	28.75	 & 	28.76	 & 	26.71	 & 	26.95	 & 	30.10	\\
50	 & 	24.53	 & 	25.76	 & 	26.55	 & 	26.83	 & 	26.84	 & 	24.69	 & 	24.71	 & 	29.52	 & 	25.00	 & 	26.24	 & 	 27.03	 & 	27.31	 & 	27.32	 & 	25.31	 & 	25.81	 & 	28.59	\\
\hline
$\sigma$  & \multicolumn{8}{c||}{\emph{Man} $512 \times 512$}     &  \multicolumn{8}{c|}{\emph{Pepper} $512 \times 512$}\\
\hline
10	 & 	32.28	 & 	32.57	 & 	32.71	 & 	32.78	 & 	32.78	 & 	32.53	 & 	31.49	 & 	33.95	 & 	32.83	 & 	33.52	 & 	 33.97	 & 	34.06	 & 	34.06	 & 	33.42	 & 	31.68	 & 	34.69	\\
20	 & 	28.67	 & 	29.13	 & 	29.38	 & 	29.49	 & 	29.49	 & 	28.76	 & 	28.36	 & 	30.56	 & 	28.98	 & 	29.81	 & 	 30.28	 & 	30.42	 & 	30.42	 & 	29.31	 & 	28.35	 & 	31.22	\\
30	 & 	26.64	 & 	27.30	 & 	27.68	 & 	27.83	 & 	27.83	 & 	26.72	 & 	26.59	 & 	28.83	 & 	26.57	 & 	27.39	 & 	 27.86	 & 	28.02	 & 	28.03	 & 	26.76	 & 	25.87	 & 	29.15	\\
40	 & 	25.16	 & 	25.98	 & 	26.48	 & 	26.67	 & 	26.67	 & 	25.27	 & 	25.40	 & 	27.61	 & 	24.73	 & 	25.54	 & 	 26.03	 & 	26.20	 & 	26.21	 & 	24.95	 & 	24.22	 & 	27.56	\\
50	 & 	23.95	 & 	24.90	 & 	25.49	 & 	25.71	 & 	25.71	 & 	24.12	 & 	24.49	 & 	26.60	 & 	23.23	 & 	24.04	 & 	 24.52	 & 	24.70	 & 	24.70	 & 	23.56	 & 	23.07	 & 	26.11	\\
\hline
\end{tabular}
\label{table:standard 21x21}
\end{table*}

\begin{table*}[ht]
\small
\setlength{\extrarowheight}{1.5pt}
\centering
\caption{Mean and standard deviations of the PSNRs over 24 independent sampling patterns. Reported are the average values over 10 testing images. Bold values are the minimum PSNRs that surpass GKD and AM.}
\begin{tabular}{|c|ccccc|ccc|}
\hline
$\sigma$ & 0.05                      & 0.1                       & 0.2                       & 0.5                       & 1             & GKD           & AM            & BM3D\\
\hline
 10  	 & 	 32.08 $\pm$ 1.01e-03 	 & 	 \textbf{32.50 $\pm$ 6.95e-04} 	 & 	 32.76 $\pm$ 1.67e-04 	 & 	 32.84 $\pm$ 3.52e-05 	 & 	 32.84  	 & 	 32.45  	 & 	 31.00  	 & 	 34.10\\
 20  	 & 	 28.32 $\pm$ 1.09e-03 	 & 	 \textbf{28.86 $\pm$ 8.55e-04} 	 & 	 29.17 $\pm$ 4.10e-04 	 & 	 29.29 $\pm$ 5.56e-05 	 & 	 29.29  	 & 	 28.42  	 & 	 27.74  	 & 	 30.72\\
 30  	 & 	 26.10 $\pm$ 1.27e-03 	 & 	 \textbf{26.72 $\pm$ 8.84e-04} 	 & 	 27.10 $\pm$ 3.46e-04 	 & 	 27.24 $\pm$ 4.25e-05 	 & 	 27.25  	 & 	 26.20  	 & 	 25.83  	 & 	 28.82\\
 40  	 & 	 24.51 $\pm$ 8.07e-04 	 & 	 \textbf{25.23 $\pm$ 7.20e-04} 	 & 	 25.67 $\pm$ 3.63e-04 	 & 	 25.84 $\pm$ 5.57e-05 	 & 	 25.85  	 & 	 24.67  	 & 	 24.59  	 & 	 27.40\\
 50  	 & 	 23.26 $\pm$ 8.67e-04 	 & 	 \textbf{24.07 $\pm$ 9.69e-04} 	 & 	 24.56 $\pm$ 3.49e-04 	 & 	 24.75 $\pm$ 6.75e-05 	 & 	 24.75  	 & 	 23.49  	 & 	 23.67  	 & 	 26.22\\
\hline
\end{tabular}
\label{table:average_psnr}
\end{table*}

As we will show in the following, the practical usefulness of the proposed MCNLM algorithm is more significant in the setting of external dictionary-based denoising, for which the classical NLM is still a leading algorithm enjoying theoretical optimality as the dictionary size grows to infinity \cite{Levin_Nadler_2011}.

\subsection{External Dictionary-based Image Denoising}
\label{section:external}
To test MCNLM for external dictionary-based image denoising, we consider the dataset of Levin and Nadler \cite{Levin_Nadler_2011}, which contains about 15,000 training images (about $n \approx 10^{10}$ image patches) from the LabelMe dataset \cite{Russel_Torralba_Murphy_2008}. For testing, we use a separate set of 2000 noisy patches, which are mutually exclusive from the training images. The results are shown in \fref{external results}.

\begin{figure}[ht]
\centering
\includegraphics[width=1\linewidth]{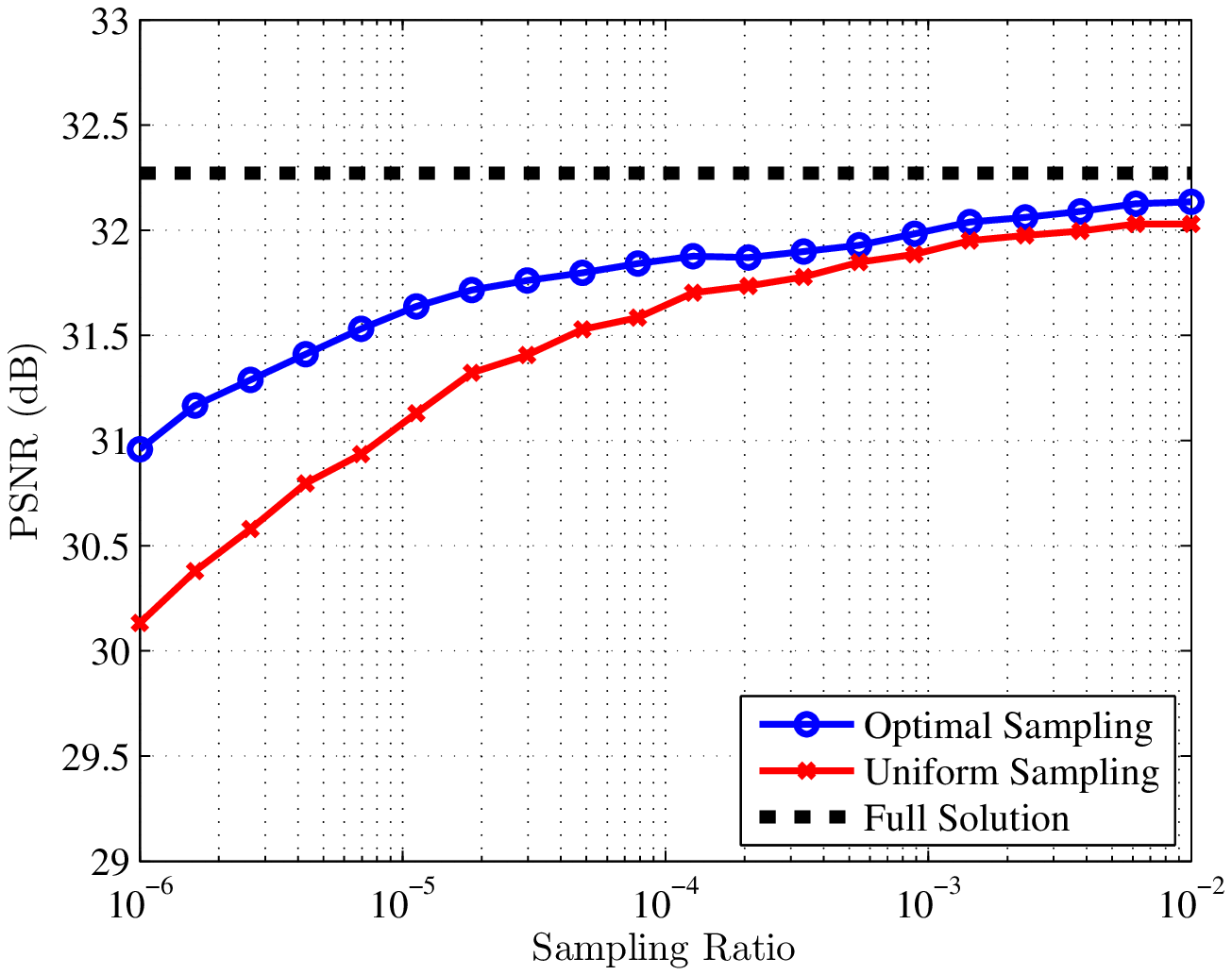}
\caption{External denoising using MCNLM. The external dataset contains $n = 10^{10}$ patches. 2000 testing patches are used to compute the PSNR. The ``dotted'' line indicates the full NLM result reported in \cite{Levin_Nadler_2011}. The ``crossed'' line indicates the MCNLM result using uniform sampling pattern, and the ``circled'' line indicates the MCNLM result using the intensity approximated sampling pattern.}
\label{fig:external results}
\end{figure}

\begin{table*}[!]
\small
\setlength{\extrarowheight}{1.5pt}
\centering
\caption{Runtime (in seconds) of MCNLM, GKD and AM. Implementations: MCNLM: MATLAB/C++ (.mex) on Windows 7, GKD: C++ on Windows 7, AM: MATLAB on Windows 7.}
\begin{tabular}{|c|c|ccccccc|}
\hline
Image Size        & Search Window / Patch Size / PCA dimension     &  0.05     & 0.1       &  0.2      & 0.5       & 1         & GKD       &  AM    \\
\hline
$512 \times 512$  & $21 \times 21$ / $5 \times 5$ / 10           & 0.495	 & 	0.731	 & 	1.547	 & 	3.505	 & 	7.234	 & 	 3.627	 & 	0.543 \\
(\emph{Man})      & $35 \times 35$ / $9 \times 9$ / 10           & 1.003	 & 	1.917	 & 	3.844	 & 	9.471	 & 	19.904	 & 	 4.948	 & 	0.546 \\
\hline
$256 \times 256$  & $21 \times 21$ / $5 \times 5$ / 10           & 0.121	 & 	0.182	 & 	0.381	 & 	0.857	 & 	1.795	 & 	 0.903	 & 	0.242 \\
(\emph{House})    & $35 \times 35$ / $9 \times 9$ / 10           & 0.248	 & 	0.475	 & 	0.954	 & 	2.362	 & 	4.851	 & 	 1.447	 & 	0.244 \\
\hline
\end{tabular}
\label{table:run time}
\end{table*}

Due to the massive size of the reference set, full evaluation of \eref{fhat} requires about one week on a 100-CPU cluster, as reported in \cite{Levin_Nadler_2011}. To demonstrate how MCNLM can be used to speed up the computation, we repeat the same experiment on a 12-CPU cluster. The testing conditions of the experiment are identical to those in \cite{Levin_Nadler_2011}. Each of the 2000 test patches is corrupted by i.i.d. Gaussian noise of standard deviation $\sigma = 18/255$. Patch size is fixed at $5 \times 5$. The weight matrix is $\mLambda = \mI$. We consider a range of sampling ratios, from $\xi = 10^{-6}$ to $\xi = 10^{-2}$. For each sampling ratio, 20 independent trials are performed and their average is recorded. Here, we show the results of the uniform sampling pattern and the optimal sampling pattern obtained using the upper bound in \eref{mean sampling bound}. The results in \fref{external results} indicate that MCNLM achieves a PSNR within $0.2$dB of the full computation at a sampling ratio of $10^{-3}$, a speed-up of about $1000$-fold.


\section{Conclusion}
\label{sec:conclusion}
We proposed Monte Carlo non-local means (MCNLM), a randomized algorithm for large-scale patch-based image filtering. MCNLM randomly chooses a fraction of the similarity weights to generate an approximated result. At any fixed sampling ratio, the probability of having large approximation errors decays exponentially with the problem size, implying that the approximated solution of MCNLM is tightly concentrated around its limiting value. Additionally, our analysis allows deriving optimized sampling patterns that exploit partial knowledge of weights of the types that are readily available in both internal and external denoising applications. Experimentally, MCNLM is competitive with other state-of-the-art accelerated NLM algorithms for single-image denoising in standard tests. When denoising with a large external database of images, MCNLM returns an approximation close to the full solution with speed-up of three orders of magnitude, suggesting its utility for large-scale image processing.

\section*{Acknowledgement}
The authors thank Anat Levin for sharing the experimental settings and datasets reported in \cite{Levin_Nadler_2011}. 

\appendix
\section{Proofs}
\subsection{Proof of \thref{thm:general_main}}
\label{appendix:convergence}
For notational simplicity, we shall drop the argument $\vp$ in $A(\vp), B(\vp)$ and $Z(\vp)$, since the sampling pattern $\vp$ remains fixed in our proof. We also define $A/B = 1$ for the case when $B = 0$. We observe that
\begin{align}
\Pr\left[ \left| Z - z \right| > \varepsilon \right] &= \Pr\left[ \left| A/B - z \right| > \varepsilon \right] \notag \\
&= \Pr\left[ \left| A/B - z \right| > \varepsilon \;\cap\; B = 0\right] \nonumber\\
&\qquad\quad + \Pr\left[ \left| A/B - z \right| > \varepsilon \;\cap\; B > 0\right] \notag \\
&\le \Pr\left[B = 0\right]\nonumber\\
&\qquad\quad  + \Pr\left[ \left| A - z B\right| > \varepsilon B \;\cap\; B > 0\right] \notag \\
&\le \Pr\left[B = 0\right] + \Pr\left[ \left| A - z B\right| > \varepsilon B \right]. \label{eq:thm1,proof,eq1}
\end{align}

By assumption, $w_j > 0$ for all $j$. It then follows from the definition in \eref{A and B} that $B = 0$ if and only if $I_j = 0$ for all $j$. Thus,
\begin{align}
\Pr[B = 0]
&= \prod_{j=1}^n (1-p_j)
= \exp\left\{ \sum_{j=1}^n \log(1-p_j) \right\} \notag \\
&\overset{(b1)}{\le} \exp\left\{ -\sum_{j=1}^n p_j \right\}
\overset{(b2)}{=} \exp\left\{ -n\xi \right\}. \label{eq:thm1,proof,term1}
\end{align}
Here, $(b1)$ holds because $\log(1-p) \le -p$ for $0 \le p \le 1$; and $(b2)$ is due to the definition that $\xi = \frac{1}{n}\sum_{j=1}^n p_j$.

Next, we provide an upper bound for $\Pr\left[ \left| A - z B\right| > \varepsilon B \right]$ in \eref{thm1,proof,eq1} by considering the two tail probabilities $\Pr\left[  A - z B > \varepsilon B \right]$ and $\Pr\left[  A - z B < - \varepsilon B \right]$ separately. Our goal here is to rewrite the two inequalities so that Bernstein's inequality in Lemma~\ref{lemma:Bernstein} can be applied. To this end, we define
\begin{equation*}
\alpha_j \bydef w_j(x_j - z - \varepsilon) \quad\mbox{and}\quad Y_j \bydef \alpha_j\left(\frac{I_j}{p_j}-1\right).
\end{equation*}
We note that $z = \mu_A/\mu_B$, where $\mu_A$ and $\mu_B$ are defined in \eref{muA} and \eref{muB}, respectively. It is easy to verify that
\begin{align}
\Pr\left[ A - z B > \varepsilon B \right]
&= \Pr\left[ \frac{1}{n}\sum_{j=1}^n Y_j > -\frac{1}{n}\sum_{j=1}^n \alpha_j\right] \notag \\
&= \Pr\left[ \frac{1}{n}\sum_{j=1}^n Y_j > \varepsilon \mu_B \right]. \label{eq:thm1,proof,eq2}
\end{align}

The random variables $Y_j$ are of zero-mean, with variance
\begin{align*}
\Var\left[ Y_j \right] = \frac{\alpha_j^2}{p_j^2} \Var[I_j] = \alpha_j^2 \frac{1-p_j}{p_j}.
\end{align*}
Using Bernstein's inequality in Lemma~\ref{lemma:Bernstein}, we can then bound the probability in \eref{thm1,proof,eq2} as
\begin{align}
&\Pr\left[ A - z B > \varepsilon B \right] \notag \\
&\le \exp\left\{ \frac{-n (\mu_B \varepsilon)^2 }{ 2 \left(\frac{1}{n}\sum_{j=1}^n \alpha_j^2 \left( \frac{1-p_j}{p_j}\right) + M_\alpha' (\mu_B\varepsilon) /3\right)} \right\},
\label{eq:thm1,proof,term2}
\end{align}
where the constant $M_\alpha'$ can be determined as follows. Since
\begin{align*}
Y_j &=\begin{cases}
\alpha_j \left(\frac{1-p_j}{p_j}\right), &\mbox{ if } I_j = 1,\\
\alpha_j , &\mbox{ if } I_j = 0,
\end{cases}
\end{align*}
it holds that
\begin{align*}
M_\alpha' &=
\max\limits_{1 \le j \le n} \left( \frac{1}{2}\left| \alpha_j \left(\frac{1-p_j}{p_j}\right) - \alpha_j \right| \right) = \max\limits_{1 \le j \le n} \left( \frac{|\alpha_j|}{2p_j} \right).
\end{align*}

The other tail probability, \emph{i.e.}, $\Pr\left[  A - z B < - \varepsilon B \right]$, can be bounded similarly. In this case, we let
\begin{align*}
\beta_j \bydef w_j(x_j - z + \varepsilon) \quad\mbox{and}\quad \Ytilde_j \bydef -\beta_j\left( \frac{I_j}{p_j} - 1 \right).
\end{align*}
Then, following the same derivations as above, we can show that
\begin{align}
&\Pr\left[  A - z B < - \varepsilon B \right] \notag \\
&\le \exp\left\{ \frac{-n (\mu_B \varepsilon)^2 }{ 2 \left(\frac{1}{n}\sum_{j=1}^n \beta_j^2 \left( \frac{1-p_j}{p_j}\right) + M_\beta' (\mu_B\varepsilon) /3\right)} \right\},\label{eq:thm1,proof,term3}
\end{align}
where $M_\beta' = \max_{1 \le j \le n} \left( \frac{|\beta_j|}{2p_j} \right)$. Substituting \eref{thm1,proof,term1}, \eref{thm1,proof,term2} and \eref{thm1,proof,term3} into \eref{thm1,proof,eq1}, and defining $M_\alpha = 2M_\alpha'$, $M_\beta = 2M_\beta'$, we are done.

\subsection{Proof of Proposition \ref{proposition:uniform sampling}}
\label{appendix:prop uniform sampling}
The goal of the proof is to simplify \eref{thm1_main} by utilizing the fact that $0\le x_j \le 1$, $0 \le z \le 1$, $0 < w_j \le 1$ and $\vp = \xi\vone$. To this end, we first observe the following:
\[
|\alpha_j| = w_j |x_j - z - \varepsilon| \le w_j||x_j - z|+\varepsilon| \le w_j(1+\varepsilon).
\]
Consequently, $M_\alpha$ is bounded as
\[
M_\alpha = \max_{1 \le j \le n} \left( \frac{|\alpha_j|}{p_j} \right) \overset{(a)}{\le} \frac{1+\varepsilon}{\xi},
\]
where in $(a)$ we used the fact that $w_j\le 1$. Similarly,
\[
|\beta_j| \le w_j(1+\varepsilon) \quad \mbox{and}\quad M_\beta \le \frac{1+\varepsilon}{\xi}.
\]
Therefore, the two negative exponents in \eref{thm1_main} are lower bounded by the following common quantity:
\begin{align*}
&\;\frac{n(\mu_B \varepsilon)^2}{2\left(\frac{1}{n}\sum_{j=1}^n w_j^2(1+\varepsilon)^2\left(\frac{1-\xi}{\xi}\right)+ \mu_B \varepsilon(1+\varepsilon)\left(\frac{1}{\xi}\right)/6\right)}\\
&\overset{(b)}{\ge} \frac{n \xi (\mu_B \varepsilon)^2 }{2\left(\frac{1}{n}\sum_{j=1}^n w_j(1+\varepsilon)^2+ \mu_B \varepsilon(1+\varepsilon)/6\right)}\\
&= \frac{n \xi (\mu_B \varepsilon)^2 }{2\left(\mu_B(1+\varepsilon)^2+ \mu_B \varepsilon(1+\varepsilon)/6\right)}\\
&= \frac{n \xi \mu_B \varepsilon^2 }{2 (1+\varepsilon)(1+7\varepsilon/6) },
\end{align*}
where in $(b)$ we used the fact that $0 \le w_j \le 1 \Rightarrow 0 \le w_j^2 \le w_j$. Defining $f(\varepsilon) \defequal \varepsilon^2/(2(1+\varepsilon)(1+7\varepsilon/6))$ yields the desired result.

\subsection{Proof of Proposition \ref{proposition:mse}}
\label{appendix:prop_mse}
The MSE can be computed as
\begin{align*}
\E\left[ \left( Z(\vp)-z \right)^2 \right]
&\overset{(a)}{=} \int_0^\infty \Pr\left[ \left( Z(\vp)-z \right)^2 > \varepsilon \right] d \varepsilon\\
&\overset{(b)}{=} \int_0^1 \Pr\left[ \left( Z(\vp)-z \right)^2 > \varepsilon \right] d \varepsilon,
\end{align*}
where $(a)$ is due to the ``layer representation'' of the expectations (See, \emph{e.g.}, \cite[Chapter 5.6]{Feller_1971}), and $(b)$ is due to the fact that $|Z(\vp) - z| \le 1$. Then, by \eref{prop1_statement}, we have that
\begin{align*}
&\int_0^1 \Pr\left[ \left( Z(\vp) - z \right)^2 > \varepsilon \right] d \varepsilon \\
&\qquad\le e^{-n\xi} + 2 \int_0^1 \exp\left\{ -n\mu_B f(\sqrt{\varepsilon}) \xi \right\} d\varepsilon.
\end{align*}
By the definition of $f(\varepsilon)$, it is easy to verify $f(\sqrt{\varepsilon}) \ge 3\varepsilon/26$. Thus,
\begin{align*}
&\int_0^1 \Pr\left[ \left( Z(\vp) - z \right)^2 > \varepsilon \right] d \varepsilon \\
&\qquad\le e^{-n\xi} + 2 \int_0^1 \exp\left\{ -n\mu_B \left(3\varepsilon/26\right) \xi  \right\} d\varepsilon\\
&\qquad\le e^{-n\xi} + \frac{1}{n\xi}\left(\frac{52}{3\mu_B}\right).
\end{align*}

\subsection{Proof of Theorem \ref{thm:optimal p_j}}
\label{appendix:thm,pj}
By introducing an auxiliary variable $t>0$, we rewrite $(P)$ as the following equivalent problem
\begin{equation}
\begin{array}{ll}
\minimize{\vp,t} &\;\; \frac{1}{n}\sum_{j=1}^n \frac{b_j^2}{p_j} + M t\\
\subjectto       &\;\; \frac{b_j}{p_j} \le t, \;\; \sum_{j=1}^n p_j = n\xi, \;\; 0 < p_j \le 1.
\end{array}
\label{eq:thm,pj,proof,equivalent}
\end{equation}
Combining the first and the third constraint, \eref{thm,pj,proof,equivalent} becomes
\begin{equation}
\begin{array}{ll}
\minimize{\vp,t} &\;\; \frac{1}{n}\sum_{j=1}^n \frac{b_j^2}{p_j} + M t\\
\subjectto       &\;\; \sum_{j=1}^n p_j = n\xi, \;\; \frac{b_j}{t} \le p_j \le 1.
\end{array}
\label{eq:thm,pj,proof,equivalent2}
\end{equation}
We note that the optimal solution of $(P)$ can be found by first minimizing \eref{thm,pj,proof,equivalent2} over $\vp$ while keeping $t$ fixed.

For fixed $t$, the lower bound $b_j/t$ in \eref{thm,pj,proof,equivalent2} is a constant with respect to $\vp$. Therefore, by applying Lemma \ref{lemma:auxiliary result 1} in Appendix E, we obtain that, for any fixed $t$, the solution of \eref{thm,pj,proof,equivalent2} is
\begin{equation}
p_j(t) = \max\left( \min\left(b_j \tau(t) , 1 \right), b_j/t \right),
\label{eq:thm,pj in t}
\end{equation}
where $\tau(t)$ is the unique solution of the following equation with respect to the variable $x$:
\begin{equation*}
\sum_{j=1}^n \max\left( \min\left(b_j x, 1 \right), b_j/t\right) = n\xi
\end{equation*}

In order to make \eref{thm,pj,proof,equivalent2} feasible, we note that it is necessary to have
\begin{equation}
t \ge \max\limits_{1\le j \le n} b_j  \quad\mbox{and}\quad   \frac{1}{n\xi}\sum_{j=1}^n b_j.
\label{eq:thm,t bound 1}
\end{equation}
The first constraint is due to the fact that $b_j/t \le p_j \le 1$ for all $j$, and the second constraint is an immediate consequence by substituting the lower bound constraint $b_j/t \le p_j$ into the equality constraint $\sum_{j=1}^n p_j = n\xi$. For $t$ satisfying \eref{thm,t bound 1}, Lemma \ref{lemma:auxiliary result 2} in Appendix E shows that $\tau_t = \tau^\ast$ is a constant with respect to $t$. Therefore, \eref{thm,pj in t} can be simplified as
\begin{equation}
p_j(t) = \max\left( c_j, b_j/t \right),
\label{eq:thm p_j independent of t}
\end{equation}
where $c_j \bydef \min(b_j \tau^\ast, 1)$ is a constant in $t$.

Substituting \eref{thm p_j independent of t} into \eref{thm,pj,proof,equivalent2}, the minimization of \eref{thm,pj,proof,equivalent2} with respect to $t$ becomes
\begin{equation}
\begin{array}{ll}
\minimize{t} &\;\; \varphi(t) \bydef \frac{1}{n}\sum\limits_{j=1}^n \frac{b_j^2}{\max(c_j, b_j/t)} + Mt\\
\subjectto   &\;\; t \ge \frac{1}{n\xi} \sum\limits_{j=1}^n b_j, \;\mbox{and}\; t \ge \max\limits_{1 \le j \le n}b_j.
\end{array}
\label{eq:thm,t optimization}
\end{equation}
Here, the inequality constraints follow from \eref{thm,t bound 1}.

Finally, the two inequality constraints in \eref{thm,t optimization} can be combined to yield
\begin{equation}
t \ge \max\left( \frac{1}{n\xi}\sum_{j=1}^n b_j, \; \max\limits_{1\le j \le n} b_j \right) \bydef  t^*.
\label{eq:thm,t ast}
\end{equation}
Since the function $f(x) = \max(c,x)$ is non-decreasing for any $c \in \R$, it follows that $\max\left( c_j,\; b_j/t \right) \le \max\left(c_j ,\; b_j/t^* \right)$, and hence $\varphi(t) \ge \varphi\left(t^*\right)$. Therefore, the minimum of $\varphi$ is attained at $t = \max\left( \frac{1}{n\xi}\sum_{j=1}^n b_j, \; \max\limits_{1\le j \le n} b_j \right)$.

\subsection{Auxiliary Results for Theorem \ref{thm:optimal p_j}}
\begin{lemma}
\label{lemma:auxiliary result 1}
Consider the optimization problem
\begin{equation*}
(P'):
\begin{array}{ll}
\minimize{\vp} &\ \ \sum\limits_{j=1}^n \frac{b_j^2}{p_j} \\
\subjectto     &\ \ \sum\limits_{j=1}^n p_j = \xi  \ \mbox{and}\  \delta_j \le p_j \le 1.
\end{array}
\end{equation*}
The solution to $(P')$ is
\begin{equation}
\label{eq:thm,optimal p_j,solution}
p_j = \max( \min(b_j \tau, 1), \delta_j), \quad \text{for } 1 \le j \le n,
\end{equation}
where the parameter $\tau$ is chosen so that $\sum_j p_j = n \xi$.
\end{lemma}

\begin{proof}
The Lagrangian of $(P')$ is
\begin{align}
\calL(\vp,\vlambda,\veta,\nu) &= \sum\limits_{j=1}^n \frac{b_j^2}{p_j} + \nu \left(\sum_{j=1}^n p_j - n\xi\right) \notag \\
&\quad  + \sum_{j=1}^n \lambda_j (p_j-1) + \sum_{j=1}^n \eta_j(\delta_j-p_j),
\end{align}
where $\vp = [p_1,\ldots,p_n]^T$ are the primal variables, $\vlambda = [\lambda_1,\ldots,\lambda_n]^T$, $\veta = [\eta_1,\ldots,\eta_n]^T$ and $\nu$ are the Lagrange multipliers associated with the constraints $p_j \le 1$, $p_j \ge \delta_j$ and $\sum_{j=1}^n p_j = n\xi$, respectively.

The first order optimality conditions imply the following:
\begin{densitemize}
\item \emph{Stationarity}: $\nabla_{\vp}\; \calL = 0$. That is, $-\frac{b_j^2}{p_j^2}+\lambda_j-\eta_j + \nu = 0$.
\item \emph{Primal feasibility}: $\sum_{j=1}^n p_j = n\xi$, $p_j \le 1$, and $p_j \ge \delta_j$.
\item \emph{Dual feasibility}: $\lambda_j \ge 0$, $\eta_j \ge 0$, and $\nu \ge 0$.
\item \emph{Complementary slackness}: $\lambda_j(p_j-1) = 0$, $\eta_j(\delta_j-p_j) = 0$.
\end{densitemize}
The first part of the complementary slackness implies that for each $j$, one of the following cases always holds: $\lambda_j = 0$ or $p_j = 1$.

Case 1: $\lambda_j = 0$. In this case, we need to further consider the condition that $\eta_j(\delta_j-p_j) = 0$. First, if $\eta_j = 0$, then $p_j \ge \delta_j$. Substituting $\lambda_j = \eta_j = 0$ into the stationarity condition yields $p_j = b_j/\sqrt{\nu}$. Since $\delta_j \le p_j \le 1$, we must have $b_j \le \sqrt{\nu} \le b_j/\delta_j$. Second, if $p_j = \delta_j$, then $\eta_j > 0$. Substituting $p_j = \delta_j$ and $\lambda_j = 0$ into the stationarity condition yields $\eta_j = \nu - b_j^2/\delta_j^2$. Since $\eta_j > 0$, we have $\sqrt{\nu} > b_j/\delta_j$.

Case 2: $p_j = 1$. In this case, $\eta_j(\delta_j-p_j) = 0$ implies that $\eta_j = 0$ because $p_j = 1 > \delta_j$. Substituting $p_j = 1$, $\eta_j = 0$ into the stationarity condition suggests that $\lambda_j = b_j^2 - \nu$. Since $\lambda_j > 0$, we have$\sqrt{\nu} < b_j$.

Combining these two cases, we obtain
\begin{equation*}
p_j = \begin{cases}
\delta_j,                         &\quad\quad \mbox{ if }\quad b_j < \delta_j \sqrt{\nu},\\
b_j / \sqrt{\nu},               &\quad\quad \mbox{ if }\quad \delta_j \sqrt{\nu} \le  b_j  \le \sqrt{\nu},\\
1,                              &\quad\quad \mbox{ if }\quad b_j > \sqrt{\nu}.
\end{cases}
\end{equation*}
By defining $\tau = 1/\sqrt{\nu}$, we prove \eref{thm,optimal p_j,solution}.

It remains to determine $\nu$. This can be done by using the primal feasibility condition that $\frac{1}{n}\sum_{j=1}^n p_j = \xi$. In particular, consider the function $g(\tau)$ defined in \eref{f_tau}, where $\tau = 1/\sqrt{\nu}$. The desired value of $\nu$ can thus be obtained by finding the root of the equation $g(\tau)$. Since $g(\tau)$ is a monotonically increasing piecewise-linear function, the parameter $\nu$ is uniquely determined, so is $\vp$.
\end{proof}

\begin{lemma}
\label{lemma:auxiliary result 2}
Let $g_t(x) = \sum_{j=1}^n \max\left( \min\left(b_j x, 1 \right), b_j/t\right)$, and for any fixed $t$, let $\tau_t$ be the solution of the equation $g_t(x) = n\xi$. For any $t \ge t^{\ast}$, where $t^\ast$ is defined in \eref{thm,t ast}, $\tau_t = \tau^\ast$ for some constant $\tau^\ast$.
\end{lemma}

\begin{proof}
First, we claim that
\begin{equation}
g_t(x)
=
\begin{cases}
\left(\sum_{j=1}^n b_j\right)/t,        &\quad\quad x \le 1/t,\\
\sum_{j=1}^n \min\left(b_j x, 1\right), &\quad\quad x > 1/t.
\end{cases}
\label{eq:lemma,g t tau}
\end{equation}

To show the first case, we observe that $b_j/t \le 1$ implies $b_j x \le b_j/t \le 1$. Thus,
\begin{align*}
g_t(x) = \sum_{j=1}^n \max\left(  b_j x, b_j/t \right) = \left(\sum_{j=1}^n b_j\right)/t.
\end{align*}
For the second case, since $x > 1/t$, it follows that $b_j/t < b_j x$. Also, because $b_j/t \le 1$, we have $b_j/t \le \min\left(b_j x, 1 \right)$. Thus,
\begin{align*}
g_t(x) = \sum_{j=1}^n \min\left(b_j x, 1 \right).
\end{align*}

Now, by assumption that $t \ge \frac{1}{n\xi}\sum_{j=1}^n b_j$, it follows from \eref{lemma,g t tau} that
\begin{equation}
g_t\left(\frac{1}{t}\right) \le n\xi.
\label{eq:lemma,g(t,t)}
\end{equation}
Since $g_t(x)$ is a constant for $x \le 1/t$ or $x \ge 1/\min_j b_j$, the only possible range for $g_t(x)=n\xi$ to have a solution is when $1/t < x < 1/\min_j b_j$. In this case, $g_t(x) = \sum_{j=1}^n \min\left(b_j x, 1\right)$ is a strictly increasing function in $x$ and so the solution is unique. Let $\tau^\ast$ be the solution of $g_t(x) = n\xi$. Since $\sum_{j=1}^n \min\left(b_j x, 1\right)$ does not involve $t$, it follows that $\tau^\ast$ is a constant in $t$.
\end{proof}

\bibliographystyle{IEEEbib}
\bibliography{ref_MCNLM}

\begin{thebibliography}{10}

\bibitem{Buades_Coll_2005_Journal}
A.~Buades, B.~Coll, and J.~Morel,
\newblock ``A review of image denoising algorithms, with a new one,''
\newblock {\em Multiscale Model. Simul.}, vol. 4, no. 2, pp. 490--530, 2005.

\bibitem{Buades_Coll_2005_AVSS}
A.~Buades, B.~Coll, and J.~Morel,
\newblock ``Denoising image sequences does not require motion estimation,''
\newblock in {\em Proc. IEEE Conf. Advanced Video and Signal Based Surveillance
  (AVSS)}, Sep. 2005, pp. 70--74.

\bibitem{Dabov_Foi_Katkovnik_2007}
K.~Dabov, A.~Foi, V.~Katkovnik, and K.~Egiazarian,
\newblock ``Image denoising by sparse {3D} transform-domain collaborative
  filtering,''
\newblock {\em IEEE Trans. Image Process.}, vol. 16, no. 8, pp. 2080--2095,
  Aug. 2007.

\bibitem{Talebi_Milanfar_2013}
H.~Talebi and P.~Milanfar,
\newblock ``Global image denoising,''
\newblock {\em IEEE Trans. Image Process.}, vol. 23, no. 2, pp. 755--768, Feb.
  2014.

\bibitem{Protter_Elad_Takeda_2009}
M.~Protter, M.~Elad, H.~Takeda, and P.~Milanfar,
\newblock ``Generalizing the non-local-means to super-resolution
  reconstruction,''
\newblock {\em IEEE Trans. Image Process.}, vol. 18, no. 1, pp. 36--51, Jan.
  2009.

\bibitem{Mairal_Bach_Ponce_2009}
J.~Mairal, F.~Bach, J.~Ponce, G.~Sapiro, and A.~Zisserman,
\newblock ``Non-local sparse models for image restoration,''
\newblock in {\em Proc. IEEE Int. Conf. Computer Vision (ICCV)}, Oct. 2009, pp.
  2272--2279.

\bibitem{Chaudhury_Singer_2013}
K.~Chaudhury and A.~Singer,
\newblock ``Non-local patch regression: Robust image denoising in patch
  space,''
\newblock in {\em Proc. IEEE Int. Conf. Acoustics, Speech and Signal Process.
  (ICASSP)}, 2013,
\newblock available online at http://arxiv.org/abs/1211.4264.

\bibitem{Milanfar_2013a}
P.~Milanfar,
\newblock ``A tour of modern image filtering,''
\newblock {\em {IEEE} Signal Processing Magazine}, vol. 30, pp. 106--128, Jan.
  2013.

\bibitem{Dong_Zhang_Shi_2013}
W.~Dong, L.~Zhang, G.~Shi, and X.~Li,
\newblock ``Nonlocally centralized sparse representation for image
  restoration,''
\newblock {\em IEEE Trans. Image Process.}, vol. 22, no. 4, pp. 1620--1630,
  Apr. 2013.

\bibitem{VanDeVille_Kocher_2009}
D.~Van~De Ville and M.~Kocher,
\newblock ``{SURE}-based non-local means,''
\newblock {\em IEEE Signal Process. Lett.}, vol. 16, no. 11, pp. 973--976, Nov.
  2009.

\bibitem{Kervrann_Boulanger_2006}
C.~Kervrann and J.~Boulanger,
\newblock ``Optimal spatial adaptation for patch-based image denoising,''
\newblock {\em IEEE Trans. Image Process.}, vol. 15, no. 10, pp. 2866--2878,
  Oct. 2006.

\bibitem{Zontak_Irani_2011}
M.~Zontak and M.~Irani,
\newblock ``Internal statistics of a single natural image,''
\newblock in {\em Proc. IEEE Conf. Computer Vision and Pattern Recognition
  (CVPR)}, Jun. 2011, pp. 977--984.

\bibitem{Levin_Nadler_2011}
A.~Levin and B.~Nadler,
\newblock ``Natural image denoising: Optimality and inherent bounds,''
\newblock in {\em Proc. IEEE Conf. Computer Vision and Pattern Recognition
  (CVPR)}, Jun. 2011, pp. 2833--2840.

\bibitem{Levin_Nadler_Durand_2012}
A.~Levin, B.~Nadler, F.~Durand, and W.~Freeman,
\newblock ``Patch complexity, finite pixel correlations and optimal
  denoising,''
\newblock in {\em Proc. 12th European Conf. Computer Vision (ECCV)}, Oct. 2012,
  vol. 7576, pp. 73--86.

\bibitem{Mahmoudi_Sapiro_2005}
M.~Mahmoudi and G.~Sapiro,
\newblock ``Fast image and video denoising via nonlocal means of similar
  neighborhoods,''
\newblock {\em IEEE Signal Process. Lett.}, vol. 12, no. 12, pp. 839--842, Dec.
  2005.

\bibitem{Coupe_Yger_2006}
P.~Coupe, P.~Yger, and C.~Barillot,
\newblock ``Fast non local means denoising for {3D MR} images,''
\newblock in {\em Proc. Medical Image Computing and Computer-Assisted
  Intervention (MICCAI)}, 2006, pp. 33--40.

\bibitem{Brox_Kleinschmidt_2008}
T.~Brox, O.~Kleinschmidt, and D.~Cremers,
\newblock ``Efficient nonlocal means for denoising of textural patterns,''
\newblock {\em IEEE Trans. Image Process.}, vol. 17, no. 7, pp. 1083--1092,
  Jul. 2008.

\bibitem{Tasdizen_2008}
T.~Tasdizen,
\newblock ``Principal components for non-local means image denoising,''
\newblock in {\em Proc. IEEE Int. Conf. Image Process. (ICIP)}, Oct. 2008, pp.
  1728 --1731.

\bibitem{Orchard_Ebrahimi_2008}
J.~Orchard, M.~Ebrahimi, and A.~Wong,
\newblock ``Efficient nonlocal-means denoising using the {SVD},''
\newblock in {\em Proc. IEEE Int. Confe. Image Process. (ICIP)}, Oct. 2008, pp.
  1732 --1735.

\bibitem{VanDeVille_Kocher_2010}
D.~Van~De Ville and M.~Kocher,
\newblock ``Nonlocal means with dimensionality reduction and {SURE}-based
  parameter selection,''
\newblock {\em IEEE Trans. Image Process.}, vol. 20, no. 9, pp. 2683 --2690,
  Sep. 2011.

\bibitem{Darbon_Cunha_2008}
J.~Darbon, A.~Cunha, T.~Chan, S.~Osher, and G.~Jensen,
\newblock ``Fast nonlocal filtering applied to electron cryomicroscopy,''
\newblock in {\em Proc. IEEE Int. Sym. Biomedical Imaging}, 2008, pp.
  1331--1334.

\bibitem{Wang_Guo_2006}
J.~Wang, Y.~Guo, Y.~Ying, Y.~Liu, and Q.~Peng,
\newblock ``Fast non-local algorithm for image denoising,''
\newblock in {\em Proc. IEEE Int.Conf. Image Process. (ICIP)}, Oct. 2006, pp.
  1429--1432.

\bibitem{Karnati_Uliyar_2009}
V.~Karnati, M.~Uliyar, and S.~Dey,
\newblock ``Fast non-local algorithm for image denoising,''
\newblock in {\em Proc. IEEE Int. Conf. Image Process. (ICIP)}, 2009, pp.
  3873--3876.

\bibitem{Vignesh_Oh_2010}
R.~Vignesh, B.~Oh, and J.~Kuo,
\newblock ``Fast non-local means {(NLM)} computation with probabilistic early
  termination,''
\newblock {\em IEEE Signal Process. Lett.}, vol. 17, no. 3, pp. 277--280, Mar.
  2010.

\bibitem{Paris_Durand_2009}
S.~Paris and F.~Durand,
\newblock ``A fast approximation of the bilateral filter using a signal
  processing approach,''
\newblock {\em Int. J. Computer Vision}, vol. 81, no. 1, pp. 24--52, Jan. 2009.

\bibitem{Yang_Duraiswami_2003}
C.~Yang, R.~Duraiswami, N.~Gumerov, and L.~Davis,
\newblock ``Improved fast {Gauss} transform and efficient kernel density
  estimation,''
\newblock in {\em Proc. Int. Conf. Computer Vision (ICCV)}, Oct. 2003, pp.
  664--671.

\bibitem{Adams_Baek_Davis_2010}
A.~Adams, J.~Baek, and M.~A. Davis,
\newblock ``Fast high-dimensional filtering using the permutohedral lattice,''
\newblock in {\em Proc. EUROGRAPHICS}, 2010, vol.~29, pp. 753--762.

\bibitem{Adams_Gelfand_2009}
A.~Adams, N.~Gelfand, J.~Dolson, and M.~Levoy,
\newblock ``Gaussian {KD}-trees for fast high-dimensional filtering,''
\newblock in {\em Proc. of ACM SIGGRAPH}, 2009,
\newblock Article No. 21.

\bibitem{Gastal_Oliveira_2012}
E.~Gastal and M.~Oliveira,
\newblock ``Adaptive manifolds for real-time high-dimensional filtering,''
\newblock {\em ACM Trans. Graphics}, vol. 31, no. 4, pp. 33:1--33:13, 2012.

\bibitem{Bhujle_Chaudhuri_2014}
H.~Bhujle and S.~Chaudhuri,
\newblock ``Novel speed-up strategies for non-local means denoising with patch
  and edge patch based dictionaries,''
\newblock {\em IEEE Trans. Image Process.}, vol. 23, no. 1, pp. 356--365, Jan.
  2014.

\bibitem{Arietta_Lawrence_2011}
S.~Arietta and J.~Lawrence,
\newblock ``Building and using a database of one trillion natural-image
  patches,''
\newblock {\em IEEE Computer Graphics and Applications}, vol. 31, no. 1, pp.
  9--19, Jan. 2011.

\bibitem{DemboZ:2010}
A.~Dembo and O.~Zeitouni,
\newblock {\em Large deviations techniques and applications},
\newblock Springer, Berlin, 2010.

\bibitem{Grimmett:2001}
G.~R. Grimmett and D.~R. Stirzaker,
\newblock {\em Probability and Random Processes},
\newblock Oxford University Press, 3rd edition, 2001.

\bibitem{Bernstein_1946}
S.~Bernstein,
\newblock {\em The Theory of Probabilities},
\newblock Gastehizdat Publishing House, Moscow, 1946.

\bibitem{Chan_Zickler_Lu_2013}
S.~H. Chan, T.~Zickler, and Y.~M. Lu,
\newblock ``Fast non-local filtering by random sampling: It works, especially
  for large images,''
\newblock in {\em Proc. IEEE Int. Conf. Acoust., Speech, Signal Process.
  (ICASSP)}, 2013, pp. 1603--1607.

\bibitem{Serfling_1974}
R.~Serfling,
\newblock ``Probability inequalities for the sum in sampling without
  replacement,''
\newblock {\em The Annals of Statistics}, vol. 2, pp. 39--48, 1974.

\bibitem{Chung_Lu_2006}
F.~Chung and L.~Lu,
\newblock ``Concentration inequalities and martingale inequalities: a survey,''
\newblock {\em Internet Mathematics}, vol. 3, no. 1, pp. 79--127, 2006.

\bibitem{Drineas_Kanna_Mahoney_2006a}
P.~Drineas, R.~Kannan, and M.~Mahoney,
\newblock ``Fast {Monte Carlo} algorithms for matrices {I}: Approximating
  matrix multiplication,''
\newblock {\em SIAM J. Computing}, vol. 36, pp. 132--157, 2006.

\bibitem{Chan_Zickler_Lu_2013_TR}
S.~H. Chan, T.~Zickler, and Y.~M. Lu,
\newblock ``{Monte-Carlo} non-local means: Random sampling for large-scale
  image filtering --- {Supplementary material},''
\newblock Tech. {R}ep., Harvard University, 2013,
\newblock [Online] http://arxiv.org/abs/1312.7366.

\bibitem{ImageStack}
A.~Adams,
\newblock ``Imagestack,'' \url{https://code.google.com/p/imagestack/}.

\bibitem{Peyre_2009}
G.~Peyre,
\newblock ``Non-local means {MATLAB} toolbox,''
  http://www.mathworks.com/matlabcentral/fileexchange/13619.

\bibitem{Russel_Torralba_Murphy_2008}
B.~Russell, A.~Torralba, K.~Murphy, and W.~Freeman,
\newblock ``{LabelMe}: a database and web-based tool for image annotation,''
\newblock {\em Int. J. Computer Vision}, vol. 77, no. 1-3, pp. 157--173, May
  2008.

\bibitem{Feller_1971}
W.~Feller,
\newblock {\em An Introduction to Probability Theory and its Applications},
  vol.~2,
\newblock John Wiley \& Sons, 2nd edition, 1971.

\end{thebibliography}


\begin{thebibliography}{1}

\bibitem{Mahmoudi_Sapiro_2005}
M.~Mahmoudi and G.~Sapiro,
\newblock ``Fast image and video denoising via nonlocal means of similar
  neighborhoods,''
\newblock {\em IEEE Signal Process. Lett.}, vol. 12, no. 12, pp. 839--842, Dec.
  2005.

\bibitem{Murphy_2012}
K.~Murphy,
\newblock {\em Machine Learning: A Probabilistic Perspective},
\newblock MIT Press, 2012.

\bibitem{Buades_Coll_2005_AVSS}
A.~Buades, B.~Coll, and J.~Morel,
\newblock ``Denoising image sequences does not require motion estimation,''
\newblock in {\em Proc. IEEE Conf. Advanced Video and Signal Based Surveillance
  (AVSS)}, Sep. 2005, pp. 70--74.

\bibitem{Adams_Gelfand_2009}
A.~Adams, N.~Gelfand, J.~Dolson, and M.~Levoy,
\newblock ``Gaussian {KD}-trees for fast high-dimensional filtering,''
\newblock in {\em Proc. of ACM SIGGRAPH}, 2009,
\newblock Article No. 21.

\bibitem{Gastal_Oliveira_2012}
E.~Gastal and M.~Oliveira,
\newblock ``Adaptive manifolds for real-time high-dimensional filtering,''
\newblock {\em ACM Trans. Graphics}, vol. 31, no. 4, pp. 33:1--33:13, 2012.

\bibitem{Dabov_Foi_Katkovnik_2007}
K.~Dabov, A.~Foi, V.~Katkovnik, and K.~Egiazarian,
\newblock ``Image denoising by sparse {3D} transform-domain collaborative
  filtering,''
\newblock {\em IEEE Trans. Image Process.}, vol. 16, no. 8, pp. 2080--2095,
  Aug. 2007.

\end{thebibliography}
\end{document}


\maketitle

\begin{abstract}
This supplementary document provides the following additional information of the main article.
\begin{itemize}
  \item Implementation of Theorem 2 (General Sampling Case)
  \item Implementation of Uniform Sampling Patterns
  \item Implementation of Spatially Approximated Sampling Patterns (for Internal Denoising)
  \item Implementation of Intensity Approximated Sampling Patterns (for External Denoising)
  \item Additional Experimental Results
\end{itemize}
\end{abstract}

\section{Implementation of Theorem 2 (General Sampling Case)}
The optimal sampling pattern presented in Theorem 2 of the main article is
\begin{equation}
p_j = \max\left( \min \left(b_j \tau, 1\right), b_j/t \right), \quad \mathrm{for } \; 1\le j \le n,
\end{equation}
where $t = \max\left( \frac{1}{n\xi}\sum_{j=1}^n b_j, \; \max_{1 \le j \le n} b_j \right)$, and the parameter $\tau$ is the root of the function
\begin{equation}
g_t(\tau) = \sum_{j=1}^n \max( \min(b_j \tau, 1), b_j/t) - n \xi.
\label{eq:g}
\end{equation}
It is easy to verify that $g_t(\cdot)$ is piecewise linear and monotonically increasing for $1/t < \tau < 1/\min_j b_j$. Thus, the solution of $g_t(\tau) = 0$ is unique. In this section, we discuss an efficient way to determine $\tau$.

\subsection{The Bisection Method}
To determine the unique root $\tau$, we apply the bisection method because of its efficiency and robustness. Gradient-based and Newton type of algorithms are not recommended because these algorithms require regions of convergence, which could be challenging to identify for the function $g_t(\cdot)$ defined in \eref{g}.

\begin{algorithm}[h]
\caption{The Bisection Method}
\begin{algorithmic}
\STATE Input: $\tau_a$, $\tau_b$.
\STATE Output: $\tau_c$.
\STATE Initialize: $F_a = g_t(\tau_a)$, $F_b = g_t(\tau_b)$, $F_c = \infty$.
\STATE
\WHILE{$|\tau_a-\tau_b| > \mbox{\texttt{tol}}$ and $|F_c - 0| > \mbox{\texttt{tol}}$}
    \STATE Define $\tau_c = (\tau_a + \tau_b)/2$, and evaluate $F_c = g_t(\tau_c)$.
    \IF{$F_a<0$ and $F_c>0$}
        \STATE Set $\tau_b = \tau_c$, and $F_b = F_c$.
    \ELSE
        \STATE Set $\tau_a = \tau_c$, and $F_a = F_c$.
    \ENDIF
\ENDWHILE
\end{algorithmic}
\label{alg:bisection}
\end{algorithm}

The bisection method is an iterative procedure that checks the signs of the two points $\tau_a,\tau_b$ and their midpoint $\tau_c = (\tau_a+\tau_b)/2$. If $\tau_c$ has the same sign as $\tau_a$, then $\tau_c$ replaces $\tau_a$. Otherwise, $\tau_c$ replaces $\tau_b$. The iteration continues until the residue $|\tau_a-\tau_b|$ is less than a tolerance level, or when $g_t(\tau_c)$ is sufficiently close to 0.

A piece of pseudo-code of the bisection method is shown in Algorithm~\ref{alg:bisection}. For a small $\xi$, we find that by setting $\tau_a = 1/t$ and $\tau_b = 1/\min_j b_j$, the bisection method typically converges in 10 iterations with a precision $|g_t(\tau_c)-0| < 10^{-1}$, which is sufficient for our problem.

\subsection{Cost Reduction by Quantization}
The cost of evaluating the function $g_t(\tau)$ for a fixed $\tau$ is $\calO(n)$: there are $n$ multiplications of $b_j \cdot \tau$, $n$ multiplications of $b_j \cdot 1/t$, and $n$ minimum/maximum operations. To reduce the cost, we note that it is possible to quantize the weights $b_j$ by constructing a histogram of $b_j$ as follows.

Let $Q$ be a predefined integer denoting the number of bins of the histogram. We define two sequences $\{u_q\}_{q=1}^Q$ and $\{l_q\}_{q=1}^Q$ such that $l_q \le b_q \le u_q$ for some $q$, and for $1\le j \le n$. That is, $u_q$ and $l_q$ are the upper and lower bounds of the values in the $q$-th bin, respectively. Also, we let the center of each bin be
\begin{align*}
b^c_q &= \frac{u_q + l_q}{2},
\end{align*}
and we let the number of elements in the $q$-th bin be
\begin{align*}
n_q &= \Big|\{ b_j \;|\; l_q \le w_j \le u_q, \quad\mbox{for } j = 1,\ldots,n \} \Big|.
\end{align*}
Then, the value $g_t(\tau)$ can be approximated by
\begin{equation}
g_t(\tau) \approx \sum_{q=1}^Q n_q \max(\min\left\{ 1, b^c_q \tau \right\}, b^c_q/t) - n \xi.
\label{eq:g(tau) approx}
\end{equation}
Essentially, the idea of quantization is to partition the weights $\{b_j \;|\; j = 1,\ldots,n\}$ into $Q$ bins, and approximate all weights in the same bin to a common value. The advantage of using \eref{g(tau) approx} instead of \eref{g} is that the cost of evaluating \eref{g(tau) approx} is $\calO(Q)$, which is significantly smaller than $\calO(n)$.

\section{Implementation of Uniform Sampling Patterns}
Uniform sampling is the fundamental building block of MCNLM's optimal sampling patterns. In this section, we discuss the implementation of uniform sampling for MCNLM. The techniques presented here will be used in other sections of this report.

For clarity we present the pseudo-codes using MATLAB language, although in practice the codes are implemented in C++.

\subsection{Naive Implementation}
To begin with, we consider the following naive implementation of uniform sampling:
\begin{quote}
\begin{verbatim}
if (rand(1)<xi)
  I(j) = 1;
else
  I(j) = 0;
end
\end{verbatim}
\end{quote}
where \texttt{rand(1)} is the MATLAB command for generating a random number from Uniform$[0,1]$. The output of the above procedure is a sequence of i.i.d. Bernoulli random variables $\{I_j\}_{j=1}^n$ with probability $\xi$.

The problem of this naive implementation is that the random number \texttt{rand(1)} has to be generated on-the-fly for $n$ times. Then, the random numbers will be compared against a double precision number $\xi$ for $n$ times. Finally, this process is repeated for $m$ times, where $m$ is the number of pixels in the noisy image. Therefore, the naive implementation is computationally expensive, although it is theoretically valid.

\emph{Remark}: In practice, the line \texttt{I(j)=1} is replaced by the actual denoising steps, \emph{e.g.} \texttt{A = A + w(j)*x(j)/xi} and \texttt{B = B + w(j)/xi}. This avoids the need of using another ``\texttt{IF I(j) == 1}'' statement when performing the denoising step.

\subsection{Fast Implementation}
Our implementation replaces the online Bernoulli sampling by a predefined (fixed) sequence of sampling indices. More precisely, we define
\begin{quote}
\begin{verbatim}
k   = round(xi*n);
idx = randi(n,k,1);
\end{verbatim}
\end{quote}
The command \texttt{k = round(xi*n)} returns the \emph{average} number of samples to be picked, and the command \texttt{idx = randi(n,k,1)} returns a list of $k$ random indices drawn uniformly from $\{1,\ldots,n\}$. Different from the naive implementation, the indices \texttt{idx} are \emph{reused} for denoising all $m$ pixels, where $m$ is the number of pixels in the noisy image. Therefore, the overall cost of the new implementation is $\calO(k)$ (for generating the random indices), as compared with $\calO(nm)$ operations in the naive implementation. The pseudo-code of the alternative implementation is shown in Algorithm~\ref{alg:uniform sampling pattern}.

\begin{algorithm}[h]
\caption{Uniform Sampling}
\begin{algorithmic}
\STATE Input: \texttt{xi,n}.
\STATE Determine \texttt{k = round(xi*n)}.
\STATE Construct \texttt{idx = randi(n,k,1)}.
\FOR{\texttt{i=1:m}}
    \FOR{\texttt{t=1:k}}
        \STATE Set \texttt{j = idx(t)}.
        \STATE Compute \texttt{w(j)} and perform other steps of NLM.
    \ENDFOR
\ENDFOR
\end{algorithmic}
\label{alg:uniform sampling pattern}
\end{algorithm}

The sampling pattern produced by the new implementation is an \emph{approximation} of the naive implementation, because a fixed sampling pattern is used for all $m$ pixels. The potential problem of such implementation is that there will be correlation between the denoised pixels because of the shared sampling pattern. However, in practice, we find that the impact of this correlation is small to the denoising quality. One way to minimize the correlation is to define multiple sampling patterns and use different patterns within certain spatial neighborhood.

\section{Implementation of Spatially Approximated Sampling Patterns (for Internal Denoising)}
In this section we discuss the implementation of the spatially approximated sampling patterns presented in Section IV.B.1 of the main article. To begin with, we recall that the spatially approximated sampling pattern is derived from the spatial weight
\begin{equation}
b_j^s = e^{-d_{i,j}^2/(2h_s^2)},
\end{equation}
where $d_{i,j}$ is the Euclidean distance between the spatial locations of the $i$-th and $j$-th pixels. Since $d_{i,j}$ is the distance, without loss of generality we can set $i = 0$.

Since $b_j^s \approx 0$ when $d_{i,j} > 3 h_s$, we set a cutoff $\rho = 3 h_s$ so that any pixel $j$ located at a position farther than $\rho$ from the $i$-th pixel will be discarded. (See Section I.B.1 for the definition of $\rho$.) We let the number of nonzero elements of $\{b_j^s\}$ be $n_s$.

\subsection{Pre-Defined Sampling Indices}
The goal of the fast implementation is to generate a sequence of sampling indices $j_1,\ldots,j_k$ by exploiting $\{b_j^s\}$. To this end, we first compute the parameter $\tau$ using the bisection method:
\begin{quote}
\begin{verbatim}
tau = bisection_method(bs, xi, n_s);
\end{verbatim}
\end{quote}
where the limit $n_s$ is the number of non-zero $\{b_j^s\}$. The computed parameter $\tau$ determines the sampling pattern $\{p_j\}$:
\begin{equation*}
p_j = \max\left( \min\left(b_j \tau, 1 \right), b_j/t \right).
\end{equation*}
The sampling pattern thus returns a sequence of indices for denoising:
\begin{verbatim}
for j=1:n_s
    if (rand(1)<p(j))
        I(j) = 1;
    end
end
\end{verbatim}
Similar to the uniform sampling case, the random indices generated by the above procedure are reused.

\subsection{Comparisons}
The performance of the spatially approximated sampling pattern is useful for small $h_s$. In \fref{gaussian comparison 1} and \fref{gaussian comparison 2}, we show two denoising examples of using the oracle sampling pattern, the uniform sampling pattern, and the spatially approximated sampling pattern. The algorithm is implemented on MATLAB/C++ (\texttt{.mex}), and supports multi-core processing. The run time shown in the figures are recorded based on a 4-CPU 3.5GHz PC.

\begin{figure}[h]
\centering
\begin{tabular}{ccc}
\includegraphics[width=0.3\linewidth]{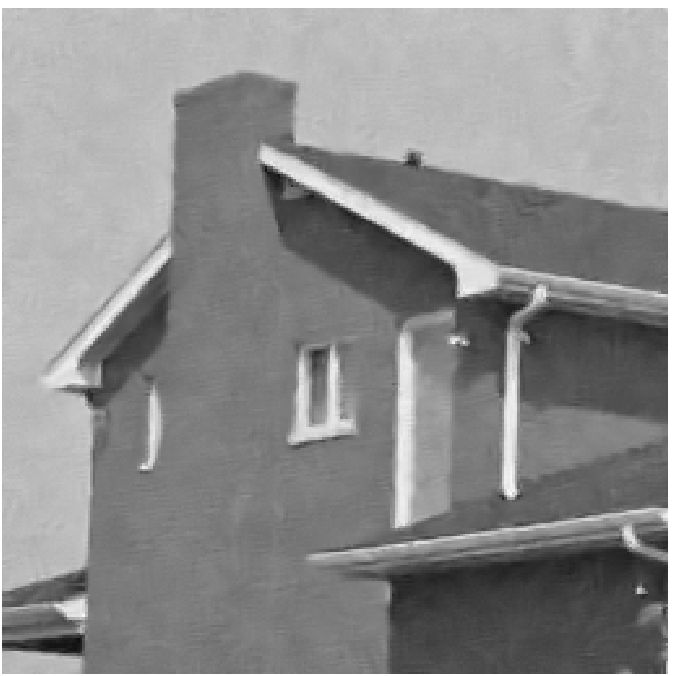}&
\includegraphics[width=0.3\linewidth]{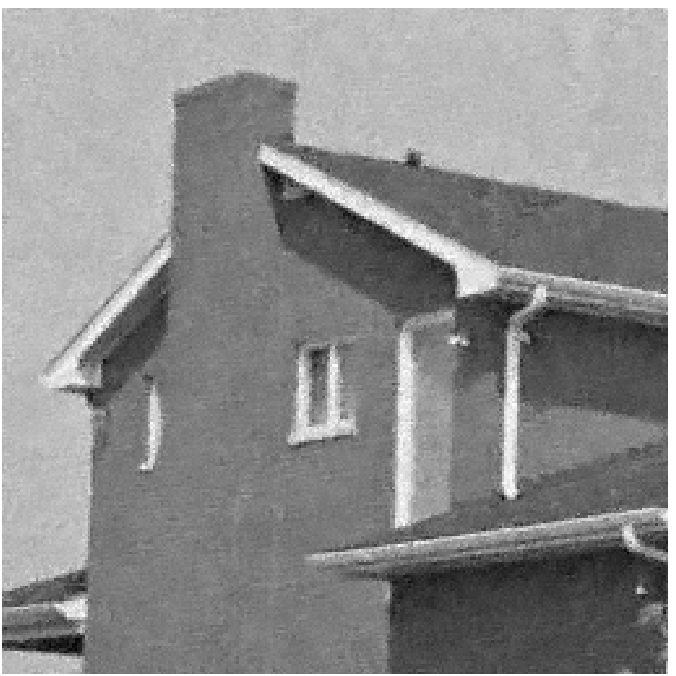}&
\includegraphics[width=0.3\linewidth]{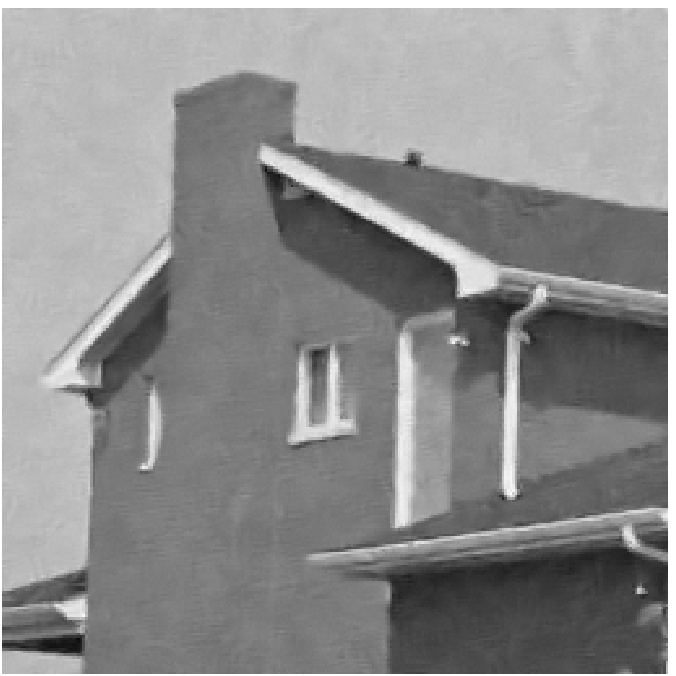}\\
(a) Oracle sampling & (b) Uniform sampling & (c) Spatially approx. sampling \\
2.662 sec, 32.5081 dB   & 0.7422 sec, 29.3738 dB   & 0.7624 sec, 32.4189 dB
\end{tabular}
\caption{\emph{House} ($256 \times 256$). Noise level is $\sigma = 20/255$. Search radius $=21 \times 21$. Parameters are $h_r = 20/255$, $h_s = 10/3$, $\rho = 10$. Patch size $=5 \times 5$. Sampling Ratio $\xi = 0.2$.}
\label{fig:gaussian comparison 1}
\end{figure}

\begin{figure}[h]
\centering
\begin{tabular}{ccc}
\includegraphics[width=0.3\linewidth]{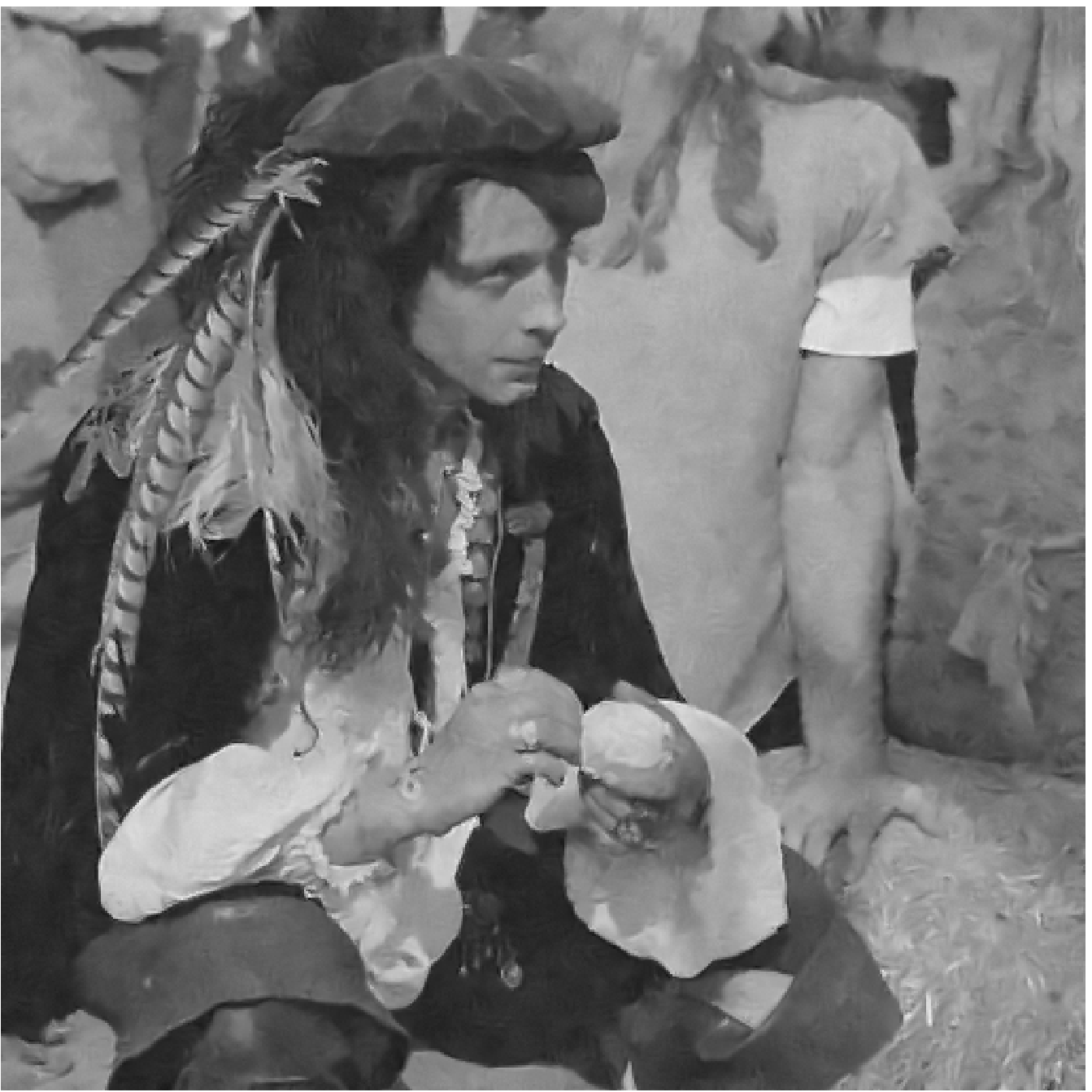}&
\includegraphics[width=0.3\linewidth]{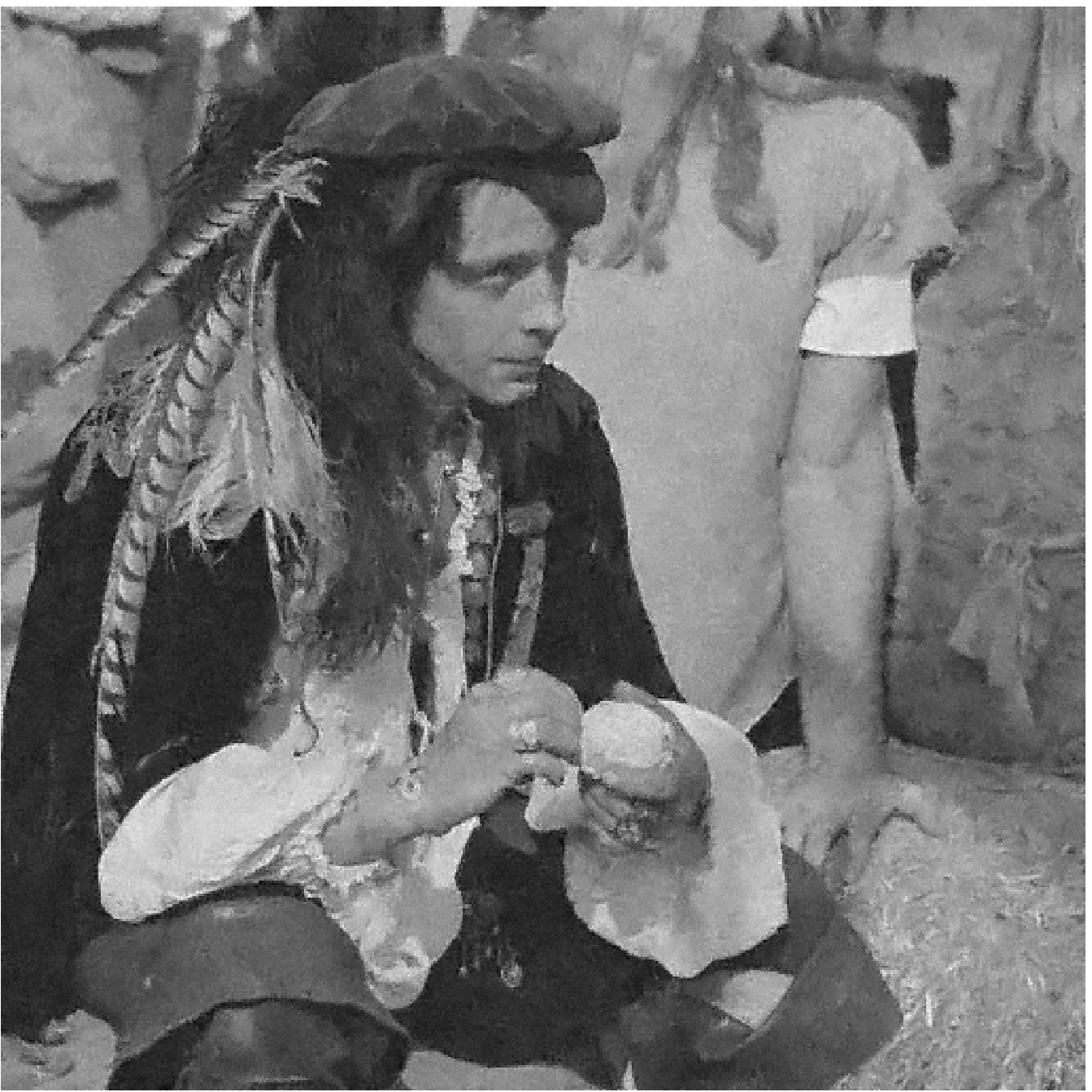}&
\includegraphics[width=0.3\linewidth]{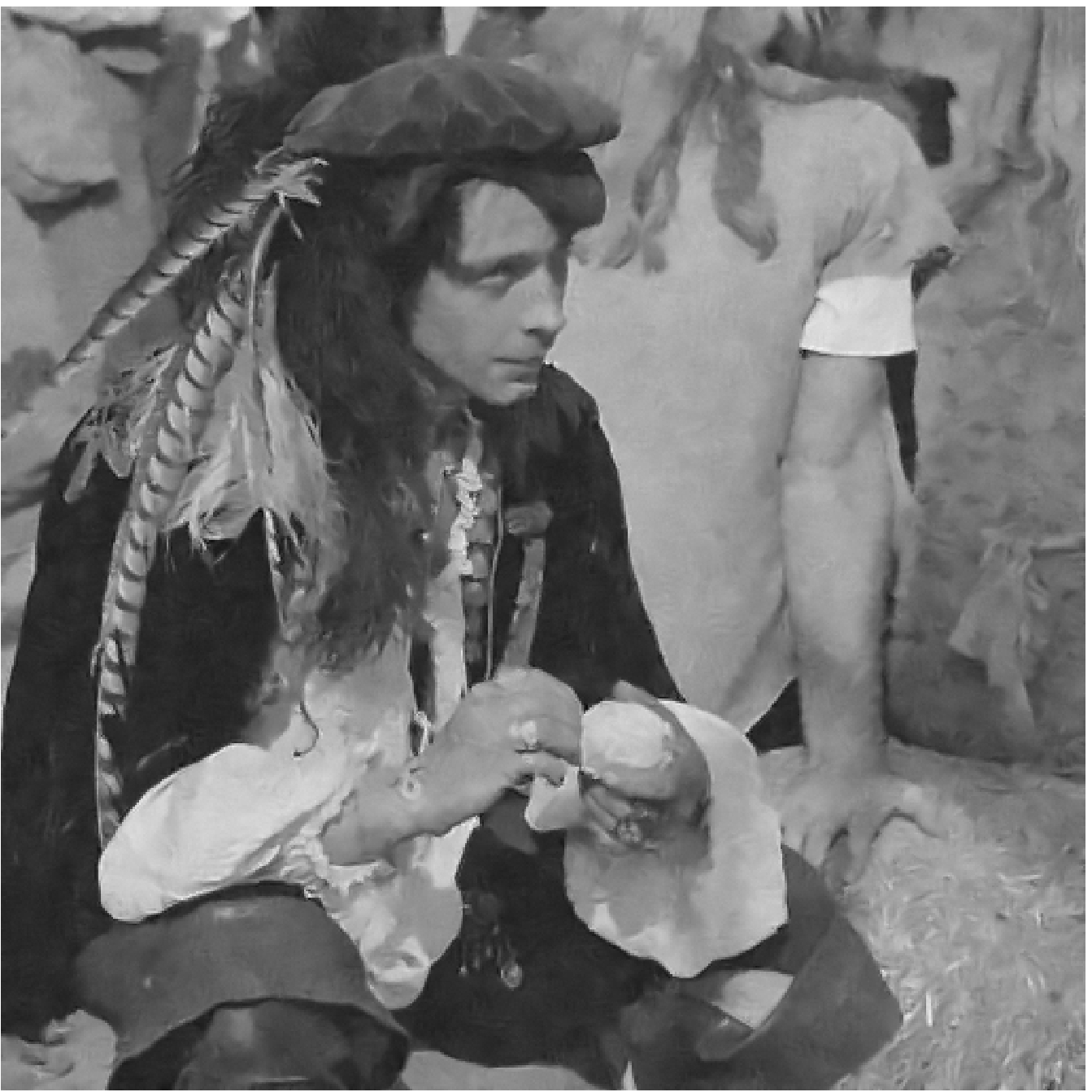}\\
(a) Oracle sampling & (b) Uniform sampling & (c) Spatially approx. sampling \\
10.7943 sec, 30.0919 dB   & 3.0603 sec, 28.2572 dB   & 3.0875 sec, 30.0283 dB
\end{tabular}
\caption{\emph{Man} ($512 \times 512$). Noise level is $\sigma = 20/255$. Search radius $=21 \times 21$. Parameters are $h_r = 20/255$, $h_s = 10/3$, $\rho = 10$. Patch size $=5 \times 5$. Sampling Ratio $\xi = 0.2$.}
\label{fig:gaussian comparison 2}
\end{figure}

\section{Implementation of Intensity Approximated Sampling Patterns (for External Denoising)}
The problem of the spatially approximated sampling pattern is that it is not applicable to external denoising, because patches in external databases do not necessarily have spatial correlations. In this case, the intensity approximated sampling pattern described in Section IV.B.2 can be used.

The idea of intensity approximated sampling is to realize that
\[
w_j \le e^{-(\vx_j^T\vs - \vy^T\vs)^2} = b_j^r,
\]
where $\vs = \mLambda\vone/(\sqrt{2}h_r \|\vone\|_{\mLambda})$. (See Section IV.B.2 for details.) The quantities $\vx_j^T\vs$ and $\vy^T\vs$ can be effectively computed by projecting $\vx_j$ (and $\vy$) onto the one-dimensional space spanned by $\vs$. If $\{\vx_j\}$ are patches collected from an image, then $\vx_j^T\vs$ can be computed through convolution \cite{Mahmoudi_Sapiro_2005}.

An implementation challenge about the projection is that since the number of $w_j$ (\emph{i.e.}, $n$) is large for external denoising, it will be inefficient to compute projections $\vx_j^T\vs$ and $\vy^T\vs$ for all $j = 1,\ldots,n$. In this section, we present a fast method that implements the intensity approximated sampling pattern without computing all projections.

\subsection{Overview}
The overall idea of the method is to use a two-stage importance sampling procedure \cite{Murphy_2012}. The motivation is that if sampling a probability distribution $p_j$ (which is $b_j^r$ in our problem) is difficult, we can first sample an easy-to-compute distribution $r_j$ such that
\begin{equation}
p_j \le r_j,
\end{equation}
and then re-sample the already picked samples according to the probability $\frac{p_j}{r_j}$. This two-stage sampling procedure is identical to the original sampling scheme. The reason is that for any Bernoulli random variable $I_j$, the probability of getting $I_j = 1$ is
\begin{equation}
p_j = r_j \cdot \frac{p_j}{r_j}.
\end{equation}
Therefore, as long as $r_j$ is an upper bound of $p_j$ for all $j$, then the two-stage sampling procedure is valid. In what follows we discuss a procedure to find a valid and efficient upper bound $r_j$.

\subsection{Quantization of $\{\xbar_j\}$}
For notational simplicity we define
\[
\xbar_j = \vx_j^T\vs, \quad\mbox{and} \quad
\ybar = \vy^T\vs
\]
as the projected signals. Then, we quantize the sequence $\{\xbar_j\}_{j=1}^n$ into a $Q$-bin histogram with bins $\calB_1,\ldots,\calB_Q$. Each bin $\calB_q$ ($q = 1,\ldots,Q$) contains a lower boundary $l_q$ and an upper boundary $u_q$. In other words, $\calB_q$ is the set of indices such that
\begin{equation}
\calB_q \defequal \{ j \;|\; l_q \le \xbar_j \le u_q \},
\end{equation}
for $q = 1,\ldots,Q$. To illustrate this idea pictorially, in \fref{xbar} we show a sorted sequence $\{\xbar_j\}$. The dotted horizontal lines are the bin boundaries. In this plot, there are $Q = 16$ bins.

Remark: the quantization is independent of the denoising process. Therefore, it can be executed off-line when preparing the dataset.

\begin{figure}[h]
\centering
\includegraphics[width=0.5\linewidth]{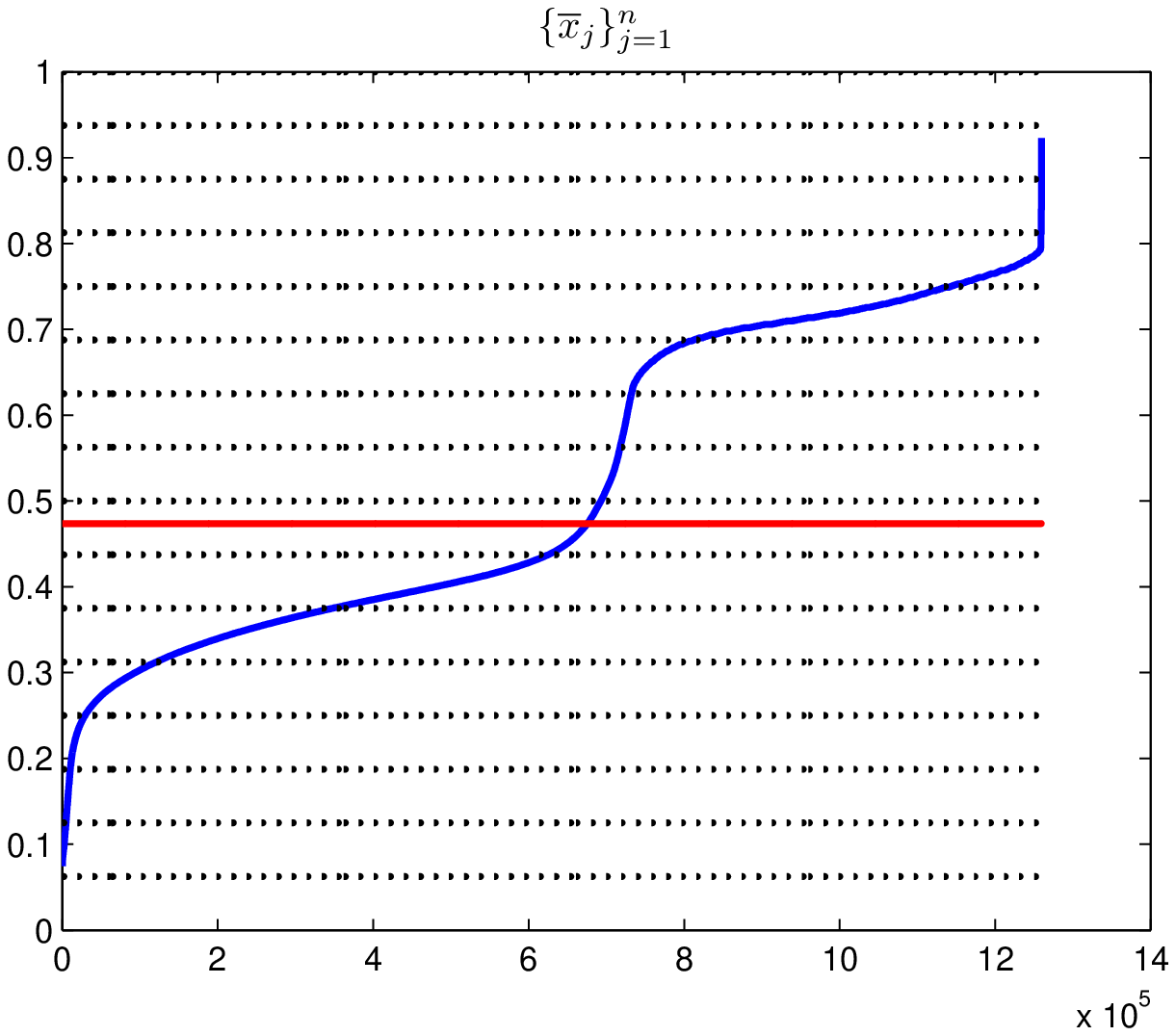}
\caption{Illustration of the quantization process. The dataset $\calX$ used in this example contains $n = 1.26 \times 10^6$ samples. The blue line is the sequence $\{\xbar_j\}_{j=1}^n$ (sorted). The black dotted lines are the quantization boundaries. The red solid line is $\ybar$.}
\label{fig:xbar}
\end{figure}

\subsection{Quantization of $\{b_j^r\}$}
Our next step is to determine an upper bound $r_j$ of $p_j$ using $\{\calB_q\}_{q=1}^Q$. First, we determine an index $1 \le q_0 \le Q$ such that the point $\ybar$ in contained in the bin $\calB_{q_0}$:
\[
\quad l_{q_0} \le \ybar \le u_{q_0},
\]
where $l_q$ and $u_q$ are the lower and upper boundaries of the histogram bins. This search procedure of finding $q_0$ can be done by sweeping through $q_0 = 1,\ldots, Q$.

Then, for all $j \in \{1,\ldots,n\}$, we define $r_j$ as
\begin{align}
r_j =
\begin{cases}
e^{-( \ybar - u_q )^2}, &\quad q \in \{1,\ldots,q_0-1\}, \,\mbox{and}\, j \in \calB_q,\\
1,                      &\quad j \in \calB_{q_0},\\
e^{-( \ybar - l_q )^2}, &\quad q \in \{q_0+1,\ldots,Q\}, \,\mbox{and}\, j \in \calB_q.
\end{cases}
\label{eq:wbar}
\end{align}
A pictorial illustration of $r_j$ is shown in \fref{wbar}, where the red color piecewise constant is $r_j$ and the blur curve is $b_j$. From both \eref{wbar} and \fref{wbar}, it can be observed that
\begin{equation}
b_j^r \le r_j,
\end{equation}
hence justifying the validity of the bound $r_j$.

\begin{figure}[h]
\centering
\includegraphics[width=0.5\linewidth]{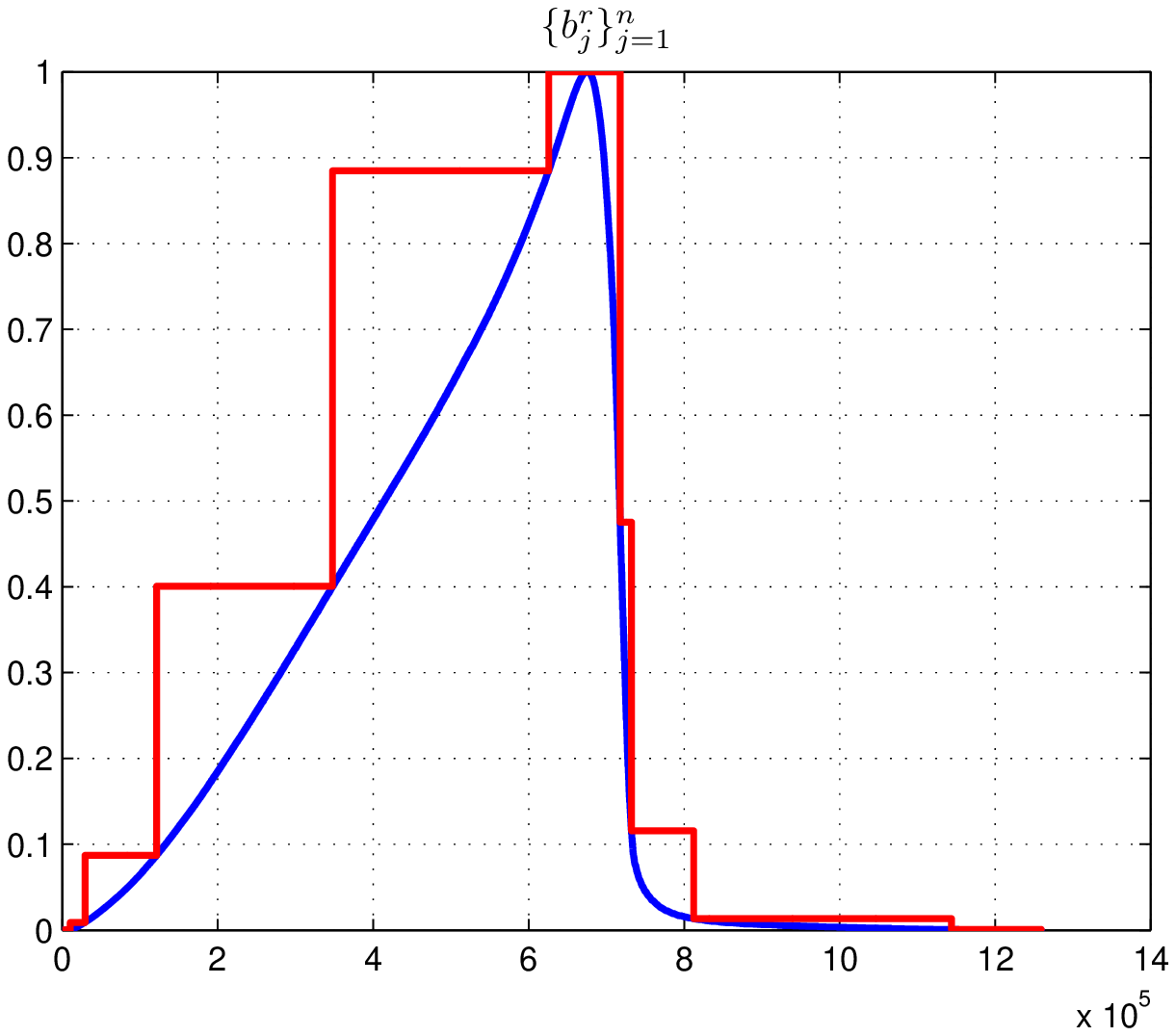}
\caption{Illustration of the quantization of $b_j^r$. The blue solid line is $\{b_j^r\}_{j=1}^n$. The red solid line is the upper bound $r_j$ defined by \eref{wbar}.}
\label{fig:wbar}
\end{figure}

\subsection{Drawing Samples}
Once $r_j$ is defined, the two-stage sampling procedure can be described as follows. We compute $\tau$ for the sequence $\{r_j\}$ to determine a probability distribution
\begin{equation}
\overline{r}_j = r_j \tau.
\end{equation}
Because $r_j$ is piecewise constant, $\overline{r}_j$ is also piecewise constant. Therefore, drawing samples according to $\overline{r}_j$ is equivalent to drawing \emph{uniformly} random samples at a probability $\overline{r}_j$. Thus, the fast implementation presented in Section II.B above can be used.

Since $\overline{r}_j$ is an upper bound of $b_j^r$, the number of samples collected at the Stage-1 sampling is guaranteed to be more than $n \xi$. However, an excessively large number of samples is undesirable as it requires more computation for Stage-2. In order to control the number of samples, we can choose an appropriate number of quantization levels $Q$. In our experiment, we find that a $Q$ ranging from 8 to 64 is sufficient for most cases.

In the Stage-2 sampling, we compute the weight $b_j^r$
\begin{equation}
b_j^r = e^{-( \xbar_j - \ybar )^2},
\end{equation}
for all $j$'s that are picked in Stage-1. Then we define the probability
\begin{equation}
p_j = b_j^r \tau',
\end{equation}
by computing an appropriate $\tau'$. Finally, we pick the weights at a probability
\[
p_j / \overline{r}_j = \frac{b_j^r \tau'}{r_j \tau}.
\]

\section{Additional Experimental Results}
In this section we provide additional numerical results for Section V of the main article. The 10 testing images are shown in \fref{standard images}.

\begin{figure}[ht]
\centering
\includegraphics[width=0.75\linewidth]{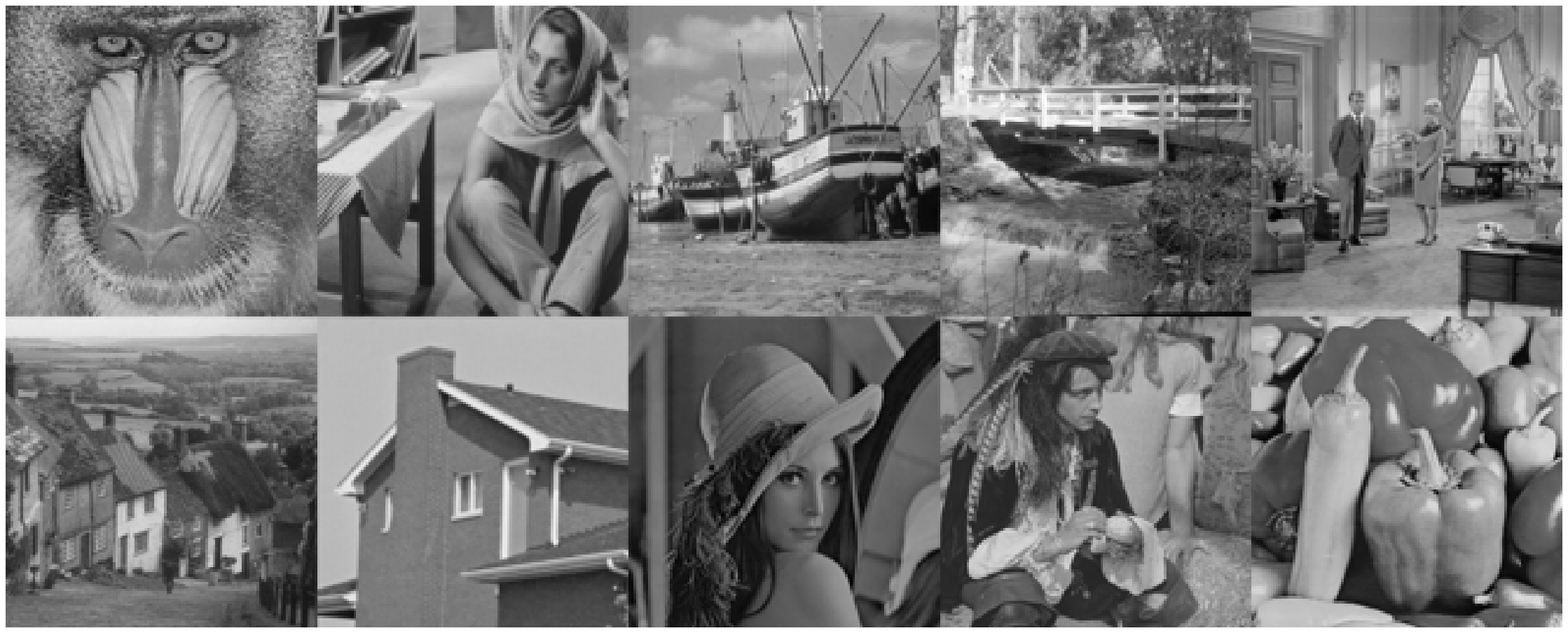}
\caption{Ten ``standard'' testing images for experiments.}
\label{fig:standard images}
\end{figure}

Table \ref{table:standard 35x35} and \ref{table:average_psnr} show additional results of Table I and II in the main article. In Table \ref{table:standard 35x35}, we show the PSNR values of the denoised images by using patches of size $7 \times 7$ and search window of size $35 \times 35$, both are larger than the one used in the main article. The results shown in Table \ref{table:standard 35x35} and \ref{table:average_psnr} are averaged over 24 independent realizations of the random sampling patterns, and over 10 independent noise realizations (totally 240 independent trials). Comparing the results to that of the main article, we observe that the results are consistent. For example, MCNLM typically has a higher PSNR than GKD and AM at $\xi = 0.1$, and has lower PSNR than BM3D even at $\xi = 1$.

In Table \ref{table:average_psnr}, we show the average PSNR and standard deviation of MCNLM over 24 random sampling patterns for a fixed noise realization. The result indicates that the fluctuation of MCNLM's result is small, which verifies the strong concentration behavior of MCNLM.

\begin{table}[!]
\footnotesize
\setlength{\extrarowheight}{1.2pt}
\setlength{\tabcolsep}{5pt}
\centering
\caption{Single image denoising by MCNLM, using the optimal Gaussian sampling pattern. Patch size is $7 \times 7$, window size is $35 \times 35$. The case when $\xi = 1$ is equivalent to the standard NLM \cite{Buades_Coll_2005_AVSS}. GKD refers to \cite{Adams_Gelfand_2009}. AM refers to \cite{Gastal_Oliveira_2012}. BM3D refers to \cite{Dabov_Foi_Katkovnik_2007}. Shown in the table are PSNR values (in dB). The results of MCNLM is averaged over 24 independent trials of using different sampling patterns, and over 10 independent noise realizations.}
\begin{tabular}{|c|cccccccc||cccccccc|}
\hline
    $\xi $& 0.05 & 0.1     & 0.2      &   0.5     &    1     & GKD & AM & BM3D &   0.05 & 0.1     &   0.2 &       0.5     &    1      & GKD  & AM  & BM3D \\
\hline
$\sigma$  & \multicolumn{8}{c||}{\emph{Baboon} $512 \times 512$} & \multicolumn{8}{c|}{\emph{Barbara} $512 \times 512$} \\
\hline
10	 & 	30.41	 & 	30.86	 & 	31.22	 & 	31.24	 & 	31.24	 & 	30.78	 & 	27.87	 & 	33.15	 & 	31.39	 & 	31.51	 & 	31.56	 & 	31.62	 & 	31.63	 & 	32.44	 & 	29.40	 & 	34.94	\\
20	 & 	26.89	 & 	27.13	 & 	27.19	 & 	27.23	 & 	27.24	 & 	26.03	 & 	24.90	 & 	29.07	 & 	27.98	 & 	28.17	 & 	28.24	 & 	28.31	 & 	28.32	 & 	27.86	 & 	26.07	 & 	31.75	\\
30	 & 	24.75	 & 	24.78	 & 	24.78	 & 	24.83	 & 	24.83	 & 	23.81	 & 	23.46	 & 	26.83	 & 	26.01	 & 	26.21	 & 	26.31	 & 	26.38	 & 	26.38	 & 	25.50	 & 	24.12	 & 	29.77	\\
40	 & 	23.37	 & 	23.46	 & 	23.51	 & 	23.56	 & 	23.56	 & 	22.64	 & 	22.66	 & 	25.26	 & 	24.58	 & 	24.81	 & 	24.93	 & 	25.00	 & 	25.01	 & 	23.90	 & 	23.01	 & 	28.02	\\
50	 & 	22.44	 & 	22.60	 & 	22.68	 & 	22.74	 & 	22.74	 & 	21.82	 & 	22.12	 & 	24.19	 & 	23.42	 & 	23.68	 & 	23.82	 & 	23.89	 & 	23.89	 & 	22.81	 & 	22.28	 & 	26.83	\\
\hline
$\sigma$  & \multicolumn{8}{c||}{\emph{Boat} $512 \times 512$}     & \multicolumn{8}{c|}{\emph{Bridge} $512 \times 512$} \\
\hline
10	 & 	31.97	 & 	32.30	 & 	32.52	 & 	32.55	 & 	32.55	 & 	31.99	 & 	29.52	 & 	33.90	 & 	29.25	 & 	29.05	 & 	28.82	 & 	28.83	 & 	28.83	 & 	29.15	 & 	27.35	 & 	30.71	\\
20	 & 	28.75	 & 	29.13	 & 	29.33	 & 	29.39	 & 	29.39	 & 	28.02	 & 	26.88	 & 	30.85	 & 	25.23	 & 	25.15	 & 	25.08	 & 	25.11	 & 	25.11	 & 	25.00	 & 	24.45	 & 	26.76	\\
30	 & 	26.87	 & 	27.16	 & 	27.30	 & 	27.37	 & 	27.37	 & 	25.78	 & 	25.06	 & 	29.01	 & 	23.53	 & 	23.54	 & 	23.53	 & 	23.57	 & 	23.57	 & 	23.18	 & 	22.70	 & 	24.97	\\
40	 & 	25.39	 & 	25.67	 & 	25.81	 & 	25.89	 & 	25.89	 & 	24.26	 & 	23.90	 & 	27.60	 & 	22.35	 & 	22.40	 & 	22.43	 & 	22.47	 & 	22.48	 & 	22.03	 & 	21.65	 & 	23.86	\\
50	 & 	24.16	 & 	24.47	 & 	24.63	 & 	24.72	 & 	24.72	 & 	23.16	 & 	23.08	 & 	26.35	 & 	21.45	 & 	21.55	 & 	21.61	 & 	21.66	 & 	21.66	 & 	21.23	 & 	20.98	 & 	22.97	\\
\hline
$\sigma$  & \multicolumn{8}{c||}{\emph{Couple} $512 \times 512$}  & \multicolumn{8}{c|}{\emph{Hill} $256 \times 256$}\\
\hline
10	 & 	31.91	 & 	32.31	 & 	32.54	 & 	32.58	 & 	32.58	 & 	31.86	 & 	29.43	 & 	34.01	 & 	30.25	 & 	30.14	 & 	30.00	 & 	30.03	 & 	30.03	 & 	30.29	 & 	29.06	 & 	31.87	\\
20	 & 	28.38	 & 	28.64	 & 	28.75	 & 	28.80	 & 	28.80	 & 	27.45	 & 	26.42	 & 	30.70	 & 	26.88	 & 	26.87	 & 	26.84	 & 	26.88	 & 	26.88	 & 	26.46	 & 	25.94	 & 	28.53	\\
30	 & 	26.24	 & 	26.43	 & 	26.51	 & 	26.57	 & 	26.57	 & 	25.26	 & 	24.71	 & 	28.73	 & 	25.20	 & 	25.28	 & 	25.32	 & 	25.36	 & 	25.36	 & 	24.65	 & 	24.42	 & 	26.93	\\
40	 & 	24.80	 & 	25.01	 & 	25.12	 & 	25.19	 & 	25.19	 & 	23.90	 & 	23.70	 & 	27.28	 & 	24.04	 & 	24.20	 & 	24.28	 & 	24.34	 & 	24.34	 & 	23.50	 & 	23.49	 & 	25.84	\\
50	 & 	23.70	 & 	23.97	 & 	24.11	 & 	24.19	 & 	24.19	 & 	22.91	 & 	22.98	 & 	26.09	 & 	23.15	 & 	23.37	 & 	23.49	 & 	23.55	 & 	23.55	 & 	22.64	 & 	22.87	 & 	24.90	\\
\hline
$\sigma$  & \multicolumn{8}{c||}{\emph{House} $256 \times 256$}   & \multicolumn{8}{c|}{\emph{Lena} $512 \times 512$}\\
\hline
10	 & 	34.01	 & 	34.66	 & 	35.09	 & 	35.17	 & 	35.17	 & 	34.11	 & 	31.46	 & 	36.72	 & 	34.93	 & 	35.53	 & 	35.87	 & 	35.93	 & 	35.93	 & 	34.35	 & 	32.87	 & 	37.03	\\
20	 & 	31.08	 & 	31.98	 & 	32.50	 & 	32.63	 & 	32.63	 & 	29.92	 & 	28.30	 & 	33.83	 & 	31.74	 & 	32.26	 & 	32.51	 & 	32.61	 & 	32.61	 & 	30.35	 & 	29.36	 & 	33.95	\\
30	 & 	29.02	 & 	29.76	 & 	30.13	 & 	30.27	 & 	30.28	 & 	27.29	 & 	25.97	 & 	32.15	 & 	29.46	 & 	29.97	 & 	30.24	 & 	30.36	 & 	30.36	 & 	28.09	 & 	27.41	 & 	31.80	\\
40	 & 	27.32	 & 	27.95	 & 	28.28	 & 	28.43	 & 	28.43	 & 	25.54	 & 	24.62	 & 	30.82	 & 	27.68	 & 	28.22	 & 	28.50	 & 	28.62	 & 	28.63	 & 	26.46	 & 	26.18	 & 	30.11	\\
50	 & 	25.77	 & 	26.34	 & 	26.64	 & 	26.78	 & 	26.78	 & 	24.26	 & 	23.71	 & 	29.48	 & 	26.16	 & 	26.71	 & 	26.99	 & 	27.12	 & 	27.13	 & 	25.20	 & 	25.28	 & 	28.62	\\
\hline
$\sigma$  & \multicolumn{8}{c||}{\emph{Man} $512 \times 512$}     &  \multicolumn{8}{c|}{\emph{Pepper} $512 \times 512$}\\
\hline
10	 & 	32.04	 & 	32.25	 & 	32.34	 & 	32.37	 & 	32.37	 & 	31.95	 & 	30.23	 & 	33.95	 & 	32.59	 & 	33.21	 & 	33.61	 & 	33.65	 & 	33.65	 & 	32.89	 & 	29.71	 & 	34.69	\\
20	 & 	28.73	 & 	28.93	 & 	29.01	 & 	29.06	 & 	29.07	 & 	28.01	 & 	27.22	 & 	30.56	 & 	29.10	 & 	29.75	 & 	30.07	 & 	30.14	 & 	30.14	 & 	28.54	 & 	26.79	 & 	31.27	\\
30	 & 	26.96	 & 	27.20	 & 	27.31	 & 	27.38	 & 	27.38	 & 	26.03	 & 	25.51	 & 	28.83	 & 	26.97	 & 	27.45	 & 	27.67	 & 	27.75	 & 	27.75	 & 	25.96	 & 	24.31	 & 	29.17	\\
40	 & 	25.70	 & 	26.01	 & 	26.17	 & 	26.25	 & 	26.26	 & 	24.68	 & 	24.46	 & 	27.61	 & 	25.25	 & 	25.63	 & 	25.82	 & 	25.90	 & 	25.91	 & 	24.21	 & 	22.71	 & 	27.58	\\
50	 & 	24.62	 & 	24.99	 & 	25.18	 & 	25.27	 & 	25.27	 & 	23.67	 & 	23.70	 & 	26.60	 & 	23.76	 & 	24.08	 & 	24.25	 & 	24.33	 & 	24.33	 & 	22.90	 & 	21.63	 & 	26.11	\\
\hline
\end{tabular}
\label{table:standard 35x35}
\end{table}

\begin{table*}[ht]
\small
\setlength{\extrarowheight}{1.5pt}
\centering
\caption{Mean and standard deviations of the PSNRs over 24 independent sampling patterns. Reported are the average values over 10 testing images. Bold values are the minimum PSNRs that surpass GKD and AM.}
\begin{tabular}{|c|ccccc|ccc|}
\hline
$\sigma$ & 0.05                      & 0.1                       & 0.2                       & 0.5                       & 1             & GKD           & AM            & BM3D\\
\hline
10	 & 	 31.87 $\pm$ 7.41e-04 	 & 	 \textbf{32.18 $\pm$ 7.81e-04} 	 & 	 32.36 $\pm$ 3.28e-04 	 & 	 32.40 $\pm$ 2.99e-05 	 & 	32.40	 & 	31.98	 & 	29.69	 & 	34.10	\\
20	 & 	 28.48 $\pm$ 1.05e-03 	 & 	 \textbf{28.80 $\pm$ 1.15e-03} 	 & 	 28.95 $\pm$ 4.45e-04 	 & 	 29.02 $\pm$ 3.11e-05 	 & 	29.02	 & 	27.77	 & 	26.63	 & 	30.73	\\
30	 & 	 26.50 $\pm$ 1.09e-03 	 & 	 \textbf{26.78 $\pm$ 5.87e-04} 	 & 	 26.91 $\pm$ 3.80e-04 	 & 	 26.98 $\pm$ 4.36e-05 	 & 	26.99	 & 	25.55	 & 	24.77	 & 	28.82	\\
40	 & 	 25.05 $\pm$ 1.10e-03 	 & 	 \textbf{25.34 $\pm$ 6.96e-04} 	 & 	 25.49 $\pm$ 3.29e-04 	 & 	 25.57 $\pm$ 5.82e-05 	 & 	25.57	 & 	24.11	 & 	23.64	 & 	27.40	\\
50	 & 	 23.86 $\pm$ 1.04e-03 	 & 	 \textbf{24.17 $\pm$ 8.40e-04} 	 & 	 24.34 $\pm$ 3.44e-04 	 & 	 24.42 $\pm$ 6.02e-05 	 & 	24.43	 & 	23.06	 & 	22.86	 & 	26.21	\\
\hline
\end{tabular}
\label{table:average_psnr}
\end{table*}

\begin{figure}[!]
\centering
\begin{subfigure}{0.3\linewidth}
\centering
\includegraphics[width=\linewidth]{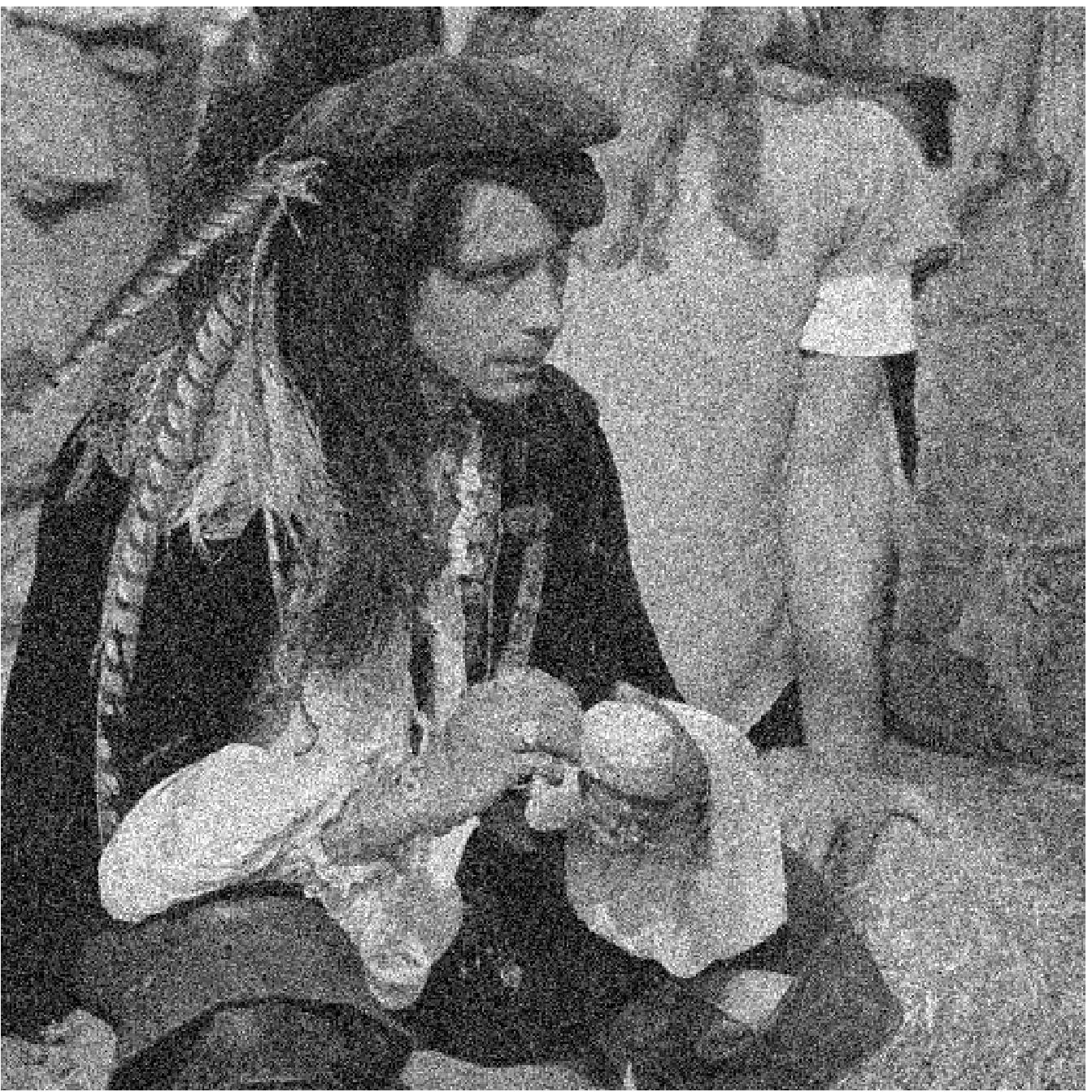}
\caption{noisy, 18.7367dB}
\end{subfigure}
\begin{subfigure}{0.3\linewidth}
\centering
\includegraphics[width=\linewidth]{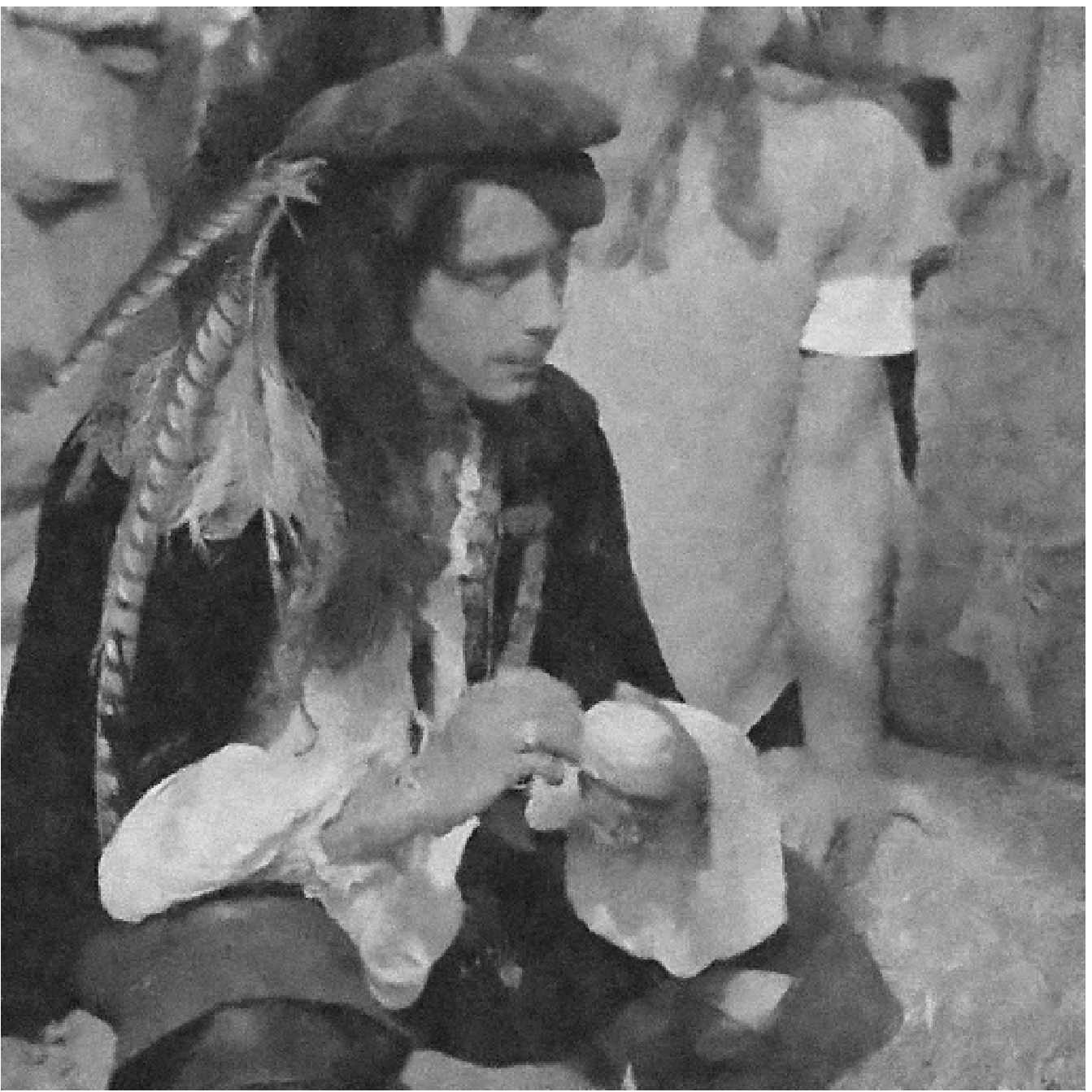}
\caption{$\xi = 0.05$, 26.6160dB}
\end{subfigure}
\begin{subfigure}{0.3\linewidth}
\centering
\includegraphics[width=\linewidth]{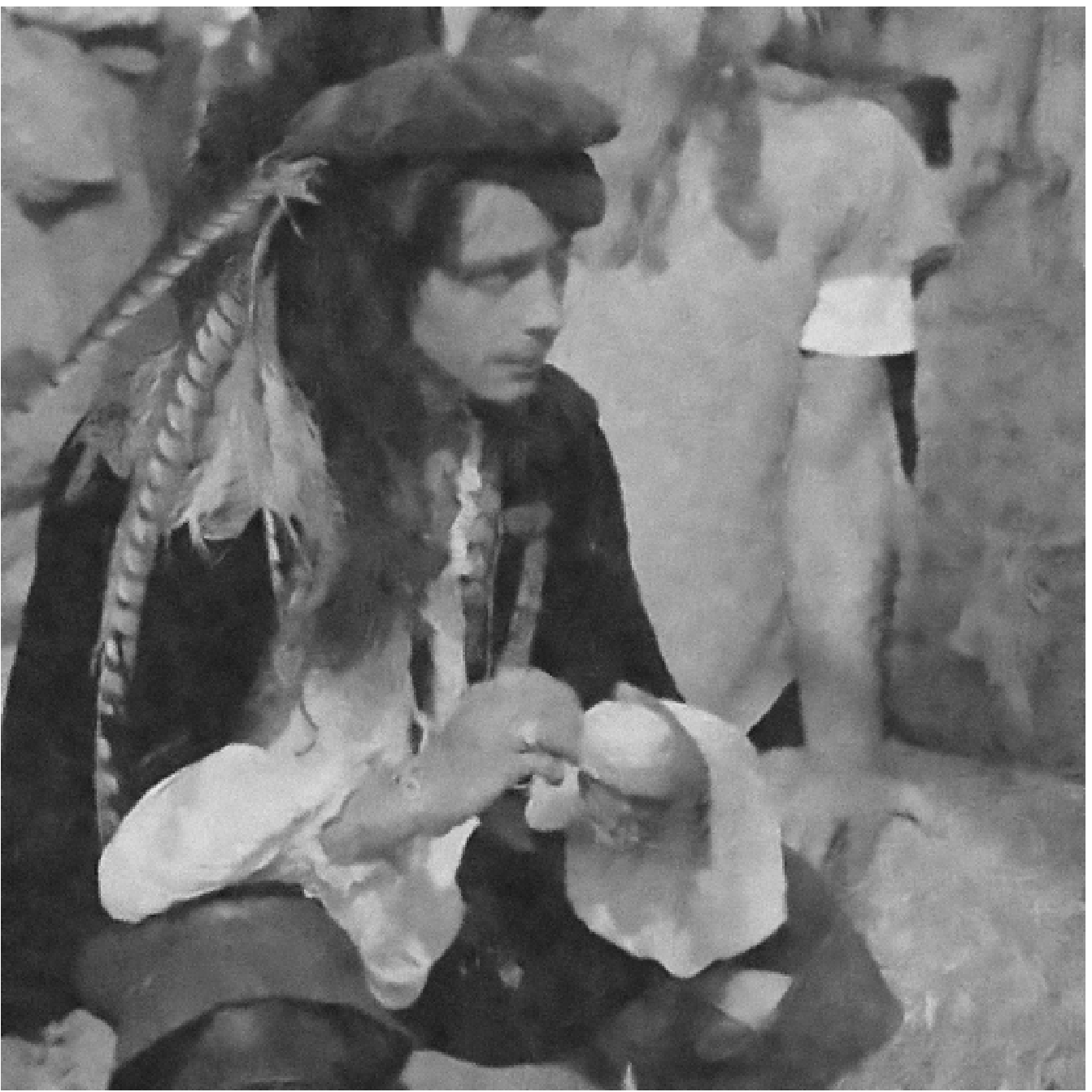}
\caption{$\xi = 0.1$, 27.2808dB}
\end{subfigure}
\begin{subfigure}{0.3\linewidth}
\centering
\includegraphics[width=\linewidth]{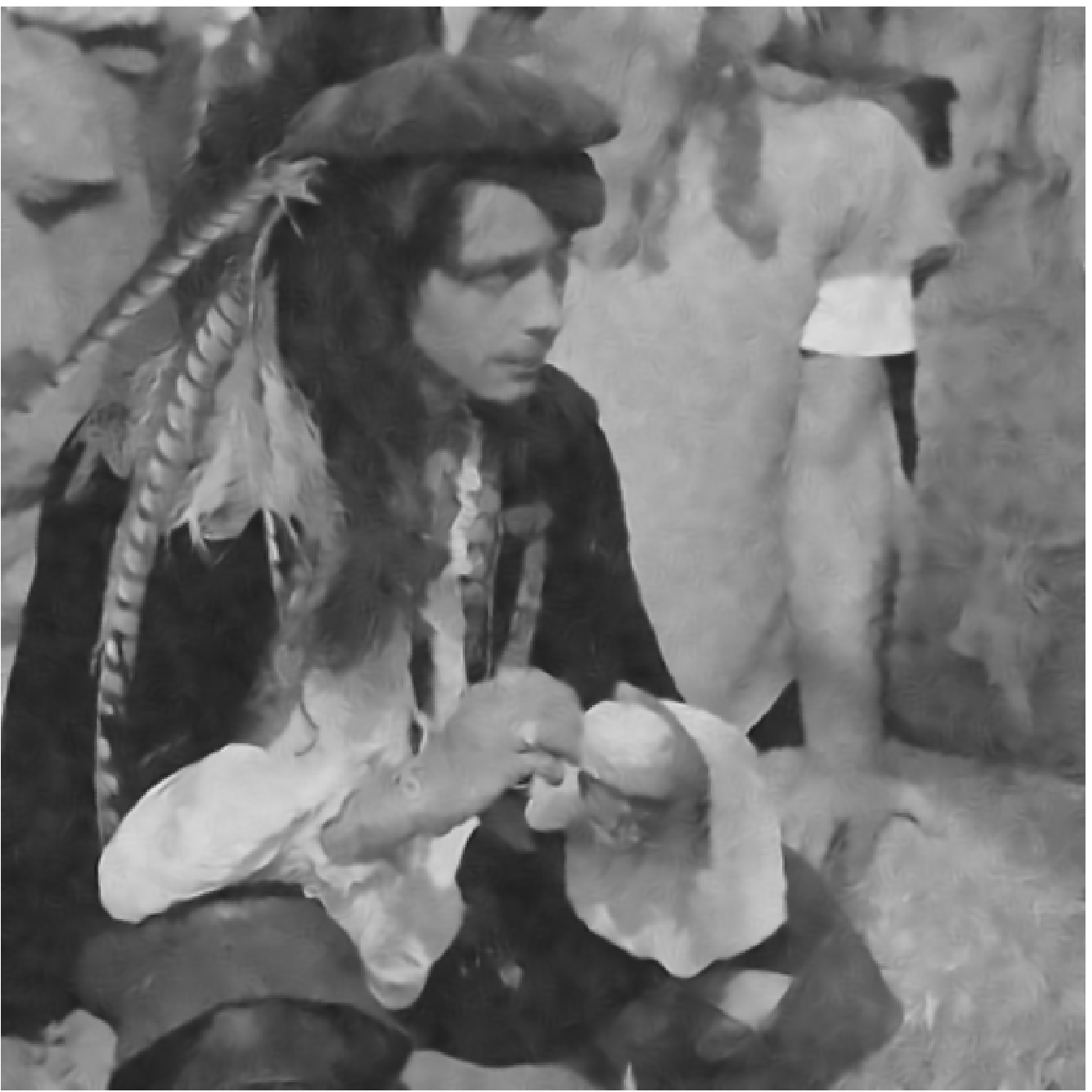}
\caption{$\xi = 1$, NLM \cite{Buades_Coll_2005_AVSS}, 27.8143dB}
\end{subfigure}
\begin{subfigure}{0.3\linewidth}
\centering
\includegraphics[width=\linewidth]{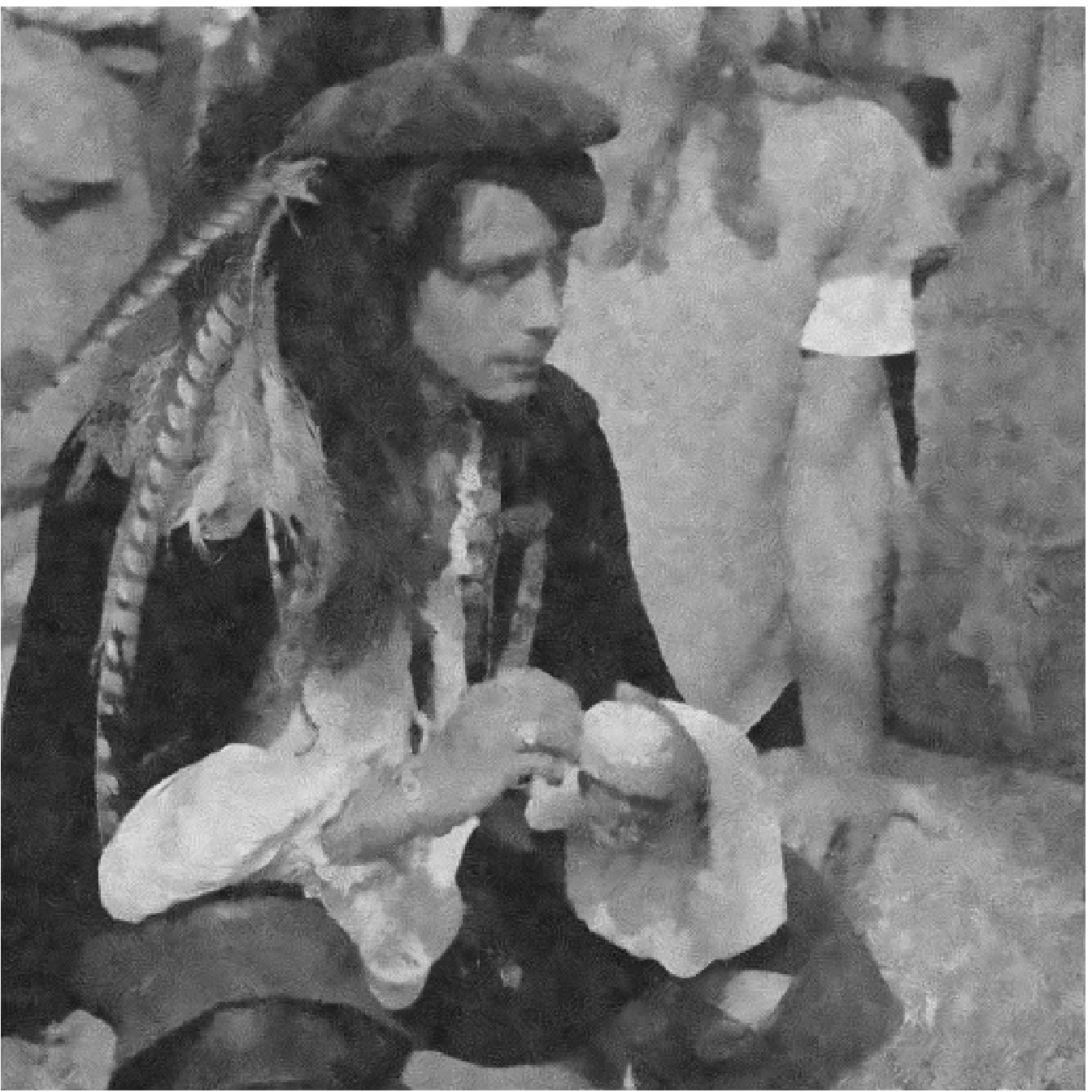}
\caption{GKD \cite{Adams_Gelfand_2009}, $26.7219$dB}
\end{subfigure}
\begin{subfigure}{0.3\linewidth}
\centering
\includegraphics[width=\linewidth]{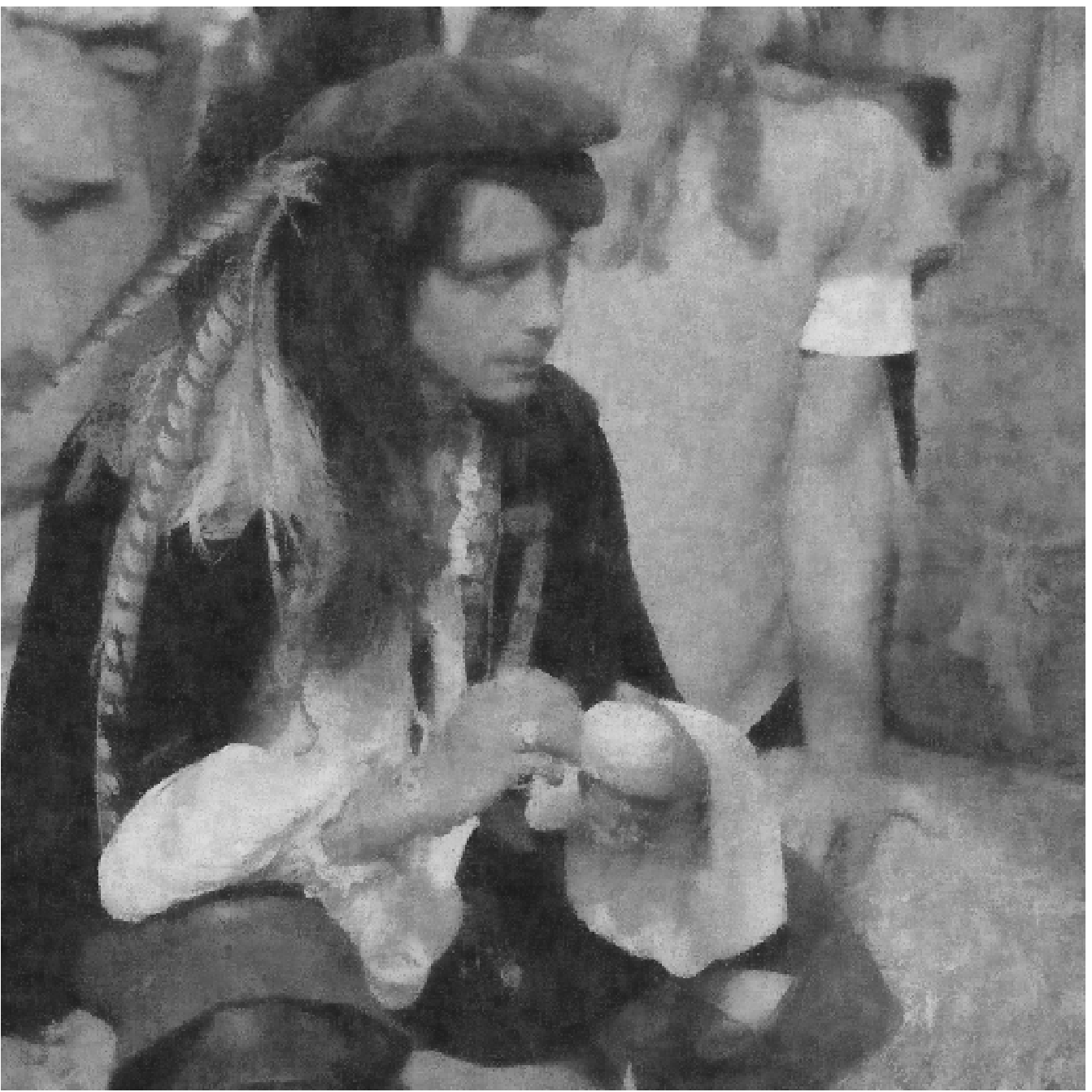}
\caption{AM \cite{Gastal_Oliveira_2012}, 26.5729dB}
\end{subfigure}
\caption{MCNLM on \emph{Man} ($512 \times 512$). Noise level is $\sigma = 30/255$. The search window has a finite size of $21 \times 21$. Patch size is $5 \times 5$. Sampling pattern: spatially approximated sampling pattern.}
\label{fig:Man results}
\end{figure}

In \fref{Man results}, we show a visual comparison between MCNLM, NLM \cite{Buades_Coll_2005_AVSS}, GKD \cite{Adams_Gelfand_2009} and AM \cite{Gastal_Oliveira_2012}. In this experiment, we considered the image \emph{Man} ($512 \times 512$) corrupted with noise of standard deviation $\sigma = 30/255$. To denoise the image, we set search window size as $21 \times 21$, and patch size as $5 \times 5$. The sampling pattern used is the spatially approximated sampling pattern.

\bibliographystyle{IEEEbib}
\bibliography{ref_MCNLM}